\documentclass{article} 
\usepackage{iclr2024_conference,times}
\usepackage{multirow}
\usepackage{multicol}

\usepackage{amsmath,amsfonts,bm}

\def\eqref#1{equation~\ref{#1}}

\def\1{\bm{1}}

\DeclareMathAlphabet{\mathsfit}{\encodingdefault}{\sfdefault}{m}{sl}
\SetMathAlphabet{\mathsfit}{bold}{\encodingdefault}{\sfdefault}{bx}{n}

\usepackage{booktabs}
\usepackage{hyperref}
\usepackage{makecell}
\usepackage{wrapfig}  
\usepackage{mathrsfs}
\usepackage{listings}
\usepackage{pifont}
\usepackage{graphicx}
\usepackage{subfigure}
\usepackage{alltt}  
\usepackage{url}
\usepackage{svg}
\usepackage{algorithm}
\usepackage{algpseudocode}
\usepackage{cleveref}

\crefname{section}{§}{§§}
\Crefname{section}{§}{§§}

\title{Beyond Hard Samples: Robust and Effective Grammatical Error Correction with Cycle Self-Augmenting}

\author{Zecheng Tang \quad Kaiqi Feng \quad Juntao Li\thanks{\scriptsize Corresponding author.} \quad Min Zhang \\
School of Computer Science and Technology, Soochow University \\
{\tt\scriptsize \{zctang, fengkq\}@stu.suda.edu.cn,\ \{ljt,minzhang\}@suda.edu.com}
}

\iclrfinalcopy 
\begin{document}
\maketitle

\begin{abstract}
Recent studies have revealed that grammatical error correction methods in the sequence-to-sequence paradigm are vulnerable to adversarial attack, and simply utilizing adversarial examples in the pre-training or post-training process can significantly enhance the robustness of GEC models to certain types of attack without suffering too much performance loss on clean data.
In this paper, we further conduct a thorough robustness evaluation of cutting-edge GEC methods for four different types of adversarial attacks and propose a simple yet very effective Cycle Self-Augmenting (CSA) method accordingly.
By leveraging the augmenting data from the GEC models themselves in the post-training process and introducing regularization data for cycle training, our proposed method can effectively improve the model robustness of well-trained GEC models with only a few more training epochs as an extra cost.
More concretely, further training on the regularization data can prevent the GEC models from over-fitting on easy-to-learn samples and thus can improve the generalization capability and robustness towards unseen data (adversarial noise/samples).
Meanwhile, the self-augmented data can provide more high-quality pseudo pairs to improve model performance on the original testing data.
Experiments on four benchmark datasets and seven strong models indicate that our proposed training method can significantly enhance the robustness of four types of attacks without using purposely built adversarial examples in training.
Evaluation results on clean data further confirm that our proposed CSA method significantly improves the performance of four baselines and yields nearly comparable results with other state-of-the-art models. Our code is available at~\url{https://github.com/ZetangForward/CSA-GEC}.
\end{abstract}

\section{Introduction}
Grammatical error correction (GEC) is one of the most essential application tasks in the NLP community for its crucial values in many scenarios including, but not limited to, writing assistant~\citep{napoles2019enabling,fitria2021grammarly}, automatic speech recognition~\citep{karat1999patterns,namazifar21_interspeech,zhao21_interspeech,wang21q_interspeech,zhang21d_interspeech}, information retrieval~\citep{gao2010large,duan2011online,hagen2017large,zhuang2021dealing}, which mainly aims to detect and correct various textual errors, such as spelling, punctuation, grammatical, word choice, and other article mistakes~\citep{wang2020comprehensive}.
Existing solutions to tackle this task can be roughly divided into two categories, i.e., sequence-to-sequence generation (\textit{Seq2Seq})~\citep{ji2017nested,chollampatt2018multilayer} and sequence-to-editing (\textit{Seq2Edits})~\citep{stahlberg2020seq2edits,awasthi2019parallel,li-shi-2021-tail}.
The former group performs the translation from ungrammatical sentences to the corresponding error-free sentences, while the latter introduces tagging or sequence labeling to merely edit a small proportion of the input sentences, remaining the rest part unchanged.

With the well-tested encoder-decoder framework~\citep{sutskever2014sequence,vaswani2017attention} as the backbone, GEC methods in the \textit{Seq2Seq} paradigm can achieve promising performance but is sensitive to the quality and scale of training data. 
Thus, many recent works have studied the problem of automatically obtaining high-quality paired data to compensate for the lack of human-labeled data pairs~\citep{zhao2019improving,kiyono2019empirical,yasunaga2021lm}.
As for the Seq2Edits group, it generally achieves a faster inference speed than Seq2Seq methods by decoding the target text in parallel and meanwhile obtaining very competitive performance~\citep{awasthi2019parallel, omelianchuk2020gector}\footnote{According to the leaderboard of CoNLL-14 shared task, three of the top-5 best performed GEC systems belong to the Seq2Edits paradigm. The link is: \url{http://nlpprogress.com/english/grammatical_error_correction.html}}.
Existing literature has also revealed that incorporating large-scale pre-trained language models (PLMs) can enhance the GEC performance of both \textit{Seq2Seq}~\citep{kaneko2020encoder} and \textit{Seq2Edits}~\citep{malmi2019encode,omelianchuk2020gector} methods.
However, recent studies have disclosed that \textit{Seq2Seq} GEC models (even with data augmentation) are vulnerable to adversarial examples, which are purposely constructed to confuse a converged model to generate wrong predictions by perturbing its inputs~\citep{wang2020improving}.
Fooling a model by perturbing its inputs, which is also called an adversarial attack, has become an essential means of exploring the model vulnerabili- ties. 
Studies on other classification tasks and PLMs further hint at the possible vulnerability of PLMs-based GEC methods~\citep{li2021searching}.
In view of the above-mentioned facts, it is imperative to conduct a systematical evaluation of existing GEC methods to adversarial attacks, especially for the under-explored \textit{Seq2Edits} paradigm and PLMs-based models. 

To fill this gap, we propose to evaluate the robustness of cutting-edge GEC models to different adversarial attacks.
More concretely, we introduce three discrete adversarial attack strategies and one continuous attack method to obtain adversarial examples.
These three discrete variations are motivated by an existing GEC work~\citep{wang2020improving} to detect the vulnerable tokens/positions that are most likely to cause the failure of the GEC models once they are substituted with the grammatical errors people may make.
In this paper, we implement three different substitution strategies, i.e., rule-based perturbations that follow the rules of~\citet{wang2020improving} and relatively imperceptible grammatical errors people may make, including synonyms and antonyms based on WordNet\footnote{\url{https://wordnet.princeton.edu/}}. 
As for the continuous adversarial noise, back-translation~\citep{sennrich2016improving} is widely used for neural GEC models before the era of large-size pre-trained language models (PLMs), and we leverage this type of perturbation to test whether GEC models constructed on powerful (PLMs) are robust enough to such simple adversarial noise.
We also propose one evaluation metric \textit{Recovery Rate} with one associated attack set which contains fixed number of attack per sentence.
Resembling the observation on the previous \textit{Seq2Seq} method attacked by mapping \& rules, cutting-edge GEC models are also susceptible to the introduced attacks.
Taking the BART-based method~\citep{lewis2020bart,katsumata2020stronger} for example, its performance ($F_{0.5}$) on CoNLL-2014~\citep{ng2014conll} decreases sharply, from 62.6 to 36.8.
Intuitively, the dramatic performance decline can be mitigated by pre-training or post-training with a great number of adversarial examples for a certain type of attack~\citep{wang2020improving}.
However, such methods require preparing considerable data for each attack type in advance, which is infeasible for real-world scenarios.
Another minor flaw of these methods is that the significant improvement in robustness is possibly accompanied by a performance decrease on the original testing data. 

To avoid these problems, we introduce the concept of regularization data which is a kind of strict hard sample, and propose a very effective cycle self-augmenting (CSA) method.
Concretely, our proposed CSA is only introduced in the post-training process of a converged GEC model and merely needs the original training data.
The self-augmented data can provide more high-quality pseudo pairs to improve model performance on the original testing data, and meanwhile, further training on the regularization data can prevent the GEC models from over-fitting on easy-to-learn samples.
Thus, our proposed CSA method can significantly improve model robustness with only a few more training epochs as the extra cost.  
Since our CSA no longer requires well-crafted adversarial examples for model training, it is more feasible in applications and can generalize well to different GEC frameworks.
Experimental results on \textbf{seven} strong models (e.g., BERT-fuse, BART, RoBERTa, XLNET) and \textbf{four} benchmark datasets (i.e., BEA, CoNLL, FCE, JFLEG) demonstrate the effectiveness of our proposed CSA method.
Our CSA method achieves significant robustness improvement on all settings and, at the same time, yields meaningful performance improvement on the original testing data (four out of seven tested models), with nearly comparable results for the left three SOTA baselines. 
By further analyzing the effects of hard samples and regularization data, we observe the advancement of the regularization data.
Besides, we also find that the trade-off between the robustness to attack and the performance on the original testing data is associated with regularization examples, where more regularization pairs in training lead to better robustness but with performance decline on the original testing data, and vice versa.
Furthermore, we find a deeper relationship between the performance of the GEC model and data characteristics by conducting quantitative experiments on the two fine-grained data components.
As for the anomalies of the slight improvement of the original testing data or attack sets for some models, we take a thorough analysis of this phenomenon and summarize a paradigm of utilizing regularization data for different model architectures in the end.

It is worthy noting that the recent work~\citep{zhang2023robustgec} also explores the robustness of the GEC systems when nuanced modifications irrelevant to errors are introduced by users, but our work is more inclined towards using CSA methods to enhance the robustness of the GEC system.

\section{Preliminary}
In this section, we summarize the key components of cutting-edge GEC methods and present a few representative works correlated with the robustness of GEC models against adversarial attacks. 
We firstly give the definition of the grammatical error correction (GEC) task. 
Then, we review some typical methods for obtaining synthetic data and introduce two most popular GEC model architectures in existing literature, i.e., \textit{Seq2Seq} and \textit{Seq2Edits}, 
In the end, we present the definition of textual adversarial attacks in NLP, following with the pilot studies of adversarial attacks in GEC and a few widely-used attack methods for other NLP tasks.

\begin{table}[!tbp]
\small
\centering
\begin{tabular}{l | l}
    \toprule
    \textbf{Level} & \textbf{Example} \\
    \midrule
    \multirow{2}{*}{Lexical} & The best place for young people in our \textcolor{red}{aree} is without doubt the lake .  \\ 
                            & The best place for young people in our \textcolor{blue}{area} is without doubt the lake . \\
    \midrule
    \multirow{2}{*}{Syntactic} & I don't recommend it to children \textcolor{red}{lower than} thirteen years old . \\ 
                              & I don't recommend it to children \textcolor{blue}{under} thirteen years old . \\
    \midrule
    \multirow{2}{*}{Semantic} & I had a \textcolor{red}{big} conversation yesterday in that house . \\
                              & I had a \textcolor{blue}{long} conversation yesterday in that house . \\
    \midrule
    \multirow{2}{*}{Discourse} & Water is made up of \textcolor{red}{one} elements, hydrogen and oxygen. \\
                               & Water is made up of \textcolor{blue}{two} elements, hydrogen and oxygen. \\
    \midrule
    \multirow{2}{*}{Pragmatic} & I need \textcolor{red}{sunscreen} because it rains so hard. \\
                               & I need \textcolor{blue}{an umbrella} because it rains so hard. \\
    \bottomrule
\end{tabular}
\caption{Different levels of errors existed in the text. Texts colored with \textbf{\textcolor{red}{red}} are wrongly written and corrected into \textbf{\textcolor{blue}{blue}} ones.}
\label{tab:error-type} 
\end{table}

\subsection{Task Definition}
\label{sec:task-definiton}
Grammatical error correction (GEC) is the task of converting an ungrammatical writing text into a grammatical one. Specifically, giving a text $x_{1}, \dots, x_{n} \in \mathcal{X} $, this task is to build a system $\mathcal{F}$ which can detect and correct the ungrammatical context without changing the meaning of the original text, and finally return the grammatical text $y_{1}, \dots, y_{m}  \in \mathcal{Y}$ ($m$ maybe not equal with $n$).  From a linguistic point of view, errors existing in the text can be classified at five levels \citep{kukich1992techniques} which are listed with their explanations below, and Table \ref{tab:error-type} provides some cases of these five error types. 
\begin{itemize}
    \item \textbf{Lexical errors} refer to misspelling a word into a non-existent one. 
    \item \textbf{Syntactic errors} violate some syntactic rules, e.g., subject-verb agreement, which causes contextual or rules mismatching.
    \item \textbf{Semantic errors} are caused by contextual spelling mistakes. Although there is no syntactic error existing, the sentence's meaning may be changed, or there are collocation/co-occurrence errors.
    \item \textbf{Discourse errors} break inherent coherence relations in a text, which may cause temporal or other conflicts in or among the sentences.
    \item \textbf{Pragmatic errors} correlate to the disobedience of common sense in the text. 
\end{itemize}

In application, GEC systems mainly focus on the first three types of errors because correcting errors from the last two types requires auxiliary tasks, e.g., discourse analysis. 

\subsection{Data Expansion}
The recent success of GEC models highly relies on the availability of massive training data pairs~\citep{awasthi2019parallel,kiyono2019empirical,omelianchuk2020gector,rothe2021simple}.
Considering that human-labeled pairs are expensive to obtain, many efforts have been devoted to exploring the generation of pseudo data pairs for GEC~\citep{lichtarge2019corpora,grundkiewicz2019neural,naplava2019grammatical}, and the combination of synthetically generated data has almost been indispensable for recently proposed GEC models~\citep{kiyono2019empirical}. 
Furthermore, in addition to generating and utilizing a large amount of synthetic data for improving the model performance under the supervised setting, how to efficiently use different data variations for probing model capability or improving model performance under an unsupervised setting during the fine-tuning stage also catches much research interests.  
We classify the prevalent operations of synthesizing data in GEC tasks into two categories, i.e., vocabulary-based perturbation and generation-based perturbation, which are detailed below.

\paragraph{Vocabulary-Based Perturbation}
Representative vocabulary-based perturbation methods are based on three basic noising operations: deletion, insertion, and replacement. 
The first two operations can be applied directly to the original grammatical sentences, while the replacement operation requires a confusion set that contains words that people often make mistakes with \citep{bryant2018language,rozovskaya2014correcting}.
We introduce a few of them as follows:
\begin{itemize}
\item With the convenient access to large scale general-domain or out-of-domain corpora, e.g., Wikipedia, some works \citep{lichtarge2018weakly, awasthi2019parallel, katsumata2020stronger} utilize vocabulary-based perturbation for these datasets to acquire massive data with relatively low-quality for pre-training a GEC model, in which the pre-training process can be further enhanced by strategies from language model pre-training, e.g., masking operation~\citep{kiyono2019empirical}.
\item Many other works explore constructing high-quality data samples. For instance, \citet{yasunaga2021lm, naplava2022czech, mita2021grammatical, zhao2019improving} focus on how to build better obfuscation sets to improve the performance further or prob the characteristics of the model by manually annotating or utilizing linguistic rules such as POS tagging. 
\end{itemize}

\paragraph{Generation-Based Perturbation}
Although the vocabulary-based perturbation is convenient and straightforward, some unrealistic error patterns that do not resemble those observed in the actual data may occur in the results~\citep{koyama2021comparison}.
Instead, some researchers utilize the neural network to learn error distributions and automatically generate errors:
\begin{itemize}
\item \citet{ge2018fluency} propose a fluency metric to evaluate the quality and correctness of sentences.
For each ungrammatical-corrected sentence pair, they utilize a sequence-to-sequence error generation model to create n-best pseudo pairs during inference, which are further sorted in ascending order of fluency. 
Then, different fluency boost learning strategies are introduced to enhance the training process.
\item \citet{wan2020improving} perform data augmentation in a controllable manner. 
Specifically, they train a model which can reconstruct a grammatical sentence into an error-specific one based on the given error type.
\item Back-translation \citep{sennrich2016improving,lample2018phrase} is an effective method to improve neural machine translation with monolingual data by augmenting the parallel training corpus with target language sentences. 
It can also be utilized for the GEC task as we can expand the training corpus with the back-translations of grammatical texts \citep{lichtarge2019corpora} or texts in other languages \citep{zhou2020improving}.
Different from the above two methods, the back-translation method is more direct and typically introduced in the post-processing procedure, i.e., modifying the decoding strategy \citep{xie2018noising,kasewa2018wronging}.
    
\item To improve the performance of GEC systems under the unsupervised setting, \citet{yasunaga2021lm} apply the BIFI framework~\citep{yasunaga2021break} to the GEC task, which contains a critic module to evaluate the outputs from the back-translation model to acquire more realistic error distributions.
\end{itemize}

Some works also pursue improving the data quality in addition to expanding the data quantity. \citet{rothe2021simple} release a CLANG-8 dataset by using a multilingual model (mT5~\citep{xue2021mt5}) to automatically clean and relabel the original LANG-8 corpus~\citep{mizumoto2011mining, tajiri2012tense}. \citet{wan2021syntax} propose a syntax-guided model to make use of the syntactic knowledge from raw data, which can achieve comparable performance without the enhancement of pre-trained models.

\begin{figure*}[!tbp]
    \centering
	\subfigure[\textit{Seq2Seq} model architecture]{
		\includegraphics[width=0.45\linewidth]{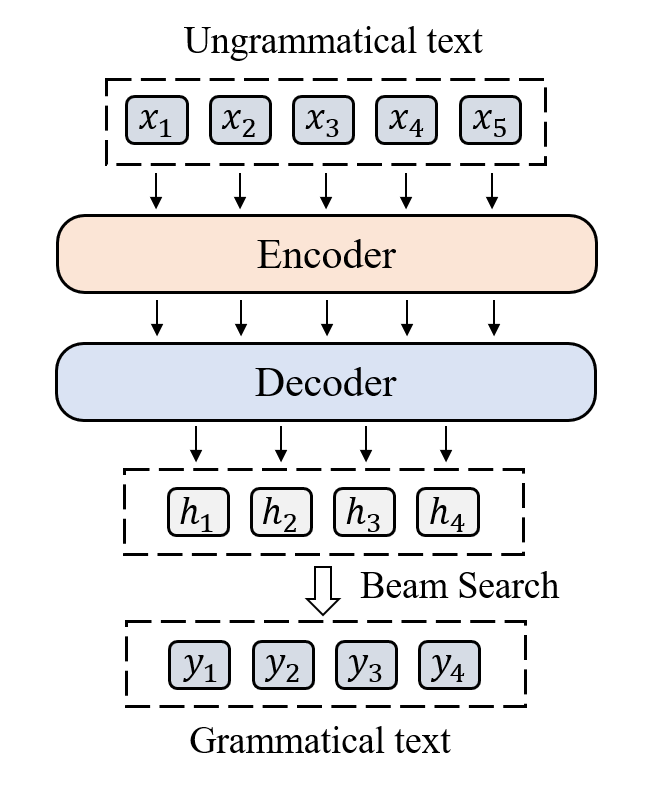}}
	\quad
	\subfigure[\textit{Seq2Edits} model architecture]{
		\includegraphics[width=0.45\linewidth]{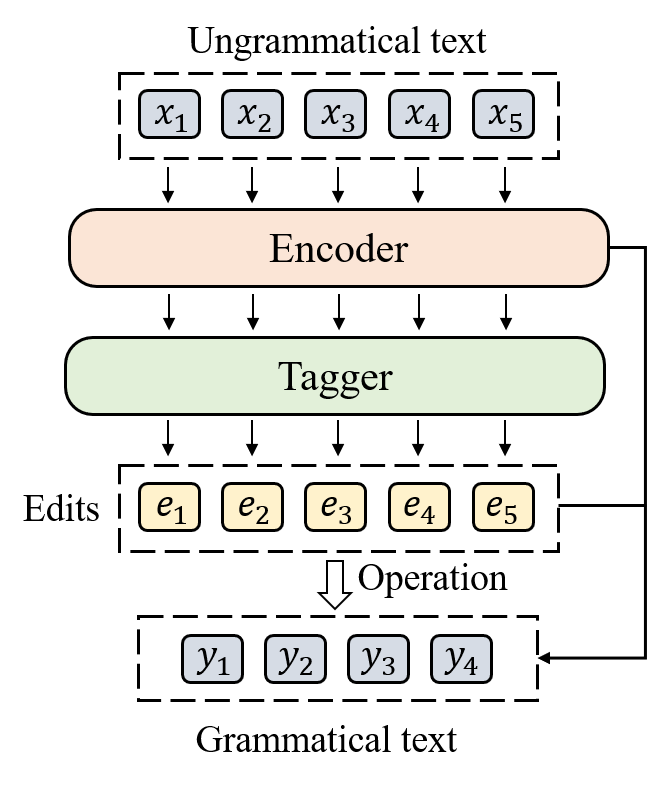}}
\caption{Two prevalent GEC model architectures. Figure (a) illustrates the \textit{Seq2Seq} framework, where $x_{i}$, $h_{i}$ and $y_{i}$ represents the input token, the correlated probability distribution outputs by the decoder and the output token, respectively. The output tokens are generated by beam search~\citep{tillmann2003word} or other decoding strategies~\citep{holtzman2019curious}. 
Figure (b) illustrates the \textit{Seq2Edits} framework, where $e_{i}$ represents an editing operation for the correlated input token $x_{i}$.
The output token $y_{i}$ is then generated by editing the corresponding input token.}
	\label{fig:model-arch}
\end{figure*}

\subsection{Adversarial Attack}
\subsubsection{Textual Adversarial Attacks Definition}
\label{sec:adversarial-definition}
The core of building textual adversarial examples is to confuse the NLP models. 
As mentioned in Section \ref{sec:task-definiton}, a GEC system $\mathcal{F}$ can map a ungrammatical sentence $x \in \mathcal{X}$ to a correct sentence $y \in \mathcal{Y}$, while an adversarial example $x^{\prime} = x^{\prime}_{1}, \dots, x^{\prime}_{n} \notin \mathcal{X} $ is built to satisfy the following paradigm:
\begin{equation}
    \label{eq:dataaug}
    \mathcal{F}(x^{\prime}) \neq y, \quad sim(x, x^{\prime}) \geq \delta,
\end{equation}
where $sim$ represents a metric function that calculates the similarity of the original $x$ and the associated adversarial example $x^{\prime}$, and $\delta$ is the minimum similarity.

\subsubsection{Application of Adversarial Attack}
In actual application scenarios, there exist different types of noise in text, and recent studies on the GEC task have revealed that existing \textit{Seq2Seq} methods are quite vulnerable to adversarial examples under the white-box setting~\citep{wang2020improving}. 
In other words, a well-trained off-the-shelf model may collapse when facing texts which happen to be adversarial examples. 
As a result, building a GEC system with robustness and generalization capability over noisy examples is critical. 
Inspired by adversarial attack and defense in NLP~\citep{jia2017adversarial, zhao2018generating}, we explore the model performance on both adversarial attack sets and regular benchmarking datasets of the GEC task.

To obtain adversarial examples, \citet{wan2020improving} propose to first identify the weak spots of a model and then replace the vulnerable tokens with two different strategies.
One is to create a correct-to-error mapping from the GEC training set.
Another is to present a series of substitution rules if there is no candidate in the mapping.
Hereafter, we denote this method as Mapping \& Rules for short.
There are also other popular adversarial example construction methods for PLMs and other tasks but are less explored in GEC, e.g., word substitutions~\citep{ma2019nlpaug,dong2021towards,li2021searching}.

\subsection{Model Architecture}
As mentioned above, the goal of GEC task is to map an ungrammatical piece ($x_{1}, \dots, x_{n}$) into a grammatical one ($y_{1}, \dots, y_{m}$) ($n$ may not be equal with $m$) from the dataset $\mathcal{D}$ with the use of GEC system $\mathcal{F}$.
As illustrated in Figure \ref{fig:model-arch}, there are two main model architectures, i.e., the \textit{Seq2Seq} framework~\citep{sutskever2014sequence,vaswani2017attention,lewis2020bart} and the \textit{Seq2Edits} framework~\citep{awasthi2019parallel,devlin2019bert}. 

\subsubsection{\textit{Seq2Seq} framework}
Many works~\citep{xie2016neural,yuan2016grammatical,xie2018noising,junczys2018approaching,zhao2019improving,sun2021instantaneous,kaneko2020encoder} formulate GEC as a natural language generation (NLG) task and utilize an encoder-decoder structure to complete the sequence-to-sequence (\textit{Seq2Seq}) generation task.
  
Given an input sentence $\bm x$ of $n$ tokens, the encoder first encodes it into the hidden representation $h^{s}_{1: n}$, and then the decoder outputs each token in an auto-regressive fashion.
The output distribution over the vocabulary at the $k$-th decoding step is conditioned on $h^{s}_{1: n}$ from the encoder and the summarized representation of previously generated $k$-1 tokens $h^{t}_{1:k-1}$ from the decoder, formulated as Pr$(y_{k} |\bm y_{< k},\bm x)$ = Pr$(y_{k} | h^{t}_{1:k-1},h^{s}_{1:n})$. 
The training objective of \textit{Seq2Seq} architecture is the negative log-likelihood, written by
\begin{equation}
    \mathcal{L}(\theta)=-\frac{1}{|\mathcal{D}|}\sum_{x,y \in \mathcal{D}}log(p(y|x))
\end{equation}
where $\theta$ refers to trainable model parameters.
To get a optimal output, beam search decoding~\citep{yuan2016grammatical, chollampatt2018multilayer} and its variations are also utilized~\citep{sun2021instantaneous}.
This architecture can achieve promising performance with massive data but will sacrifice inference efficiency due to the iterative decoding. 

\subsubsection{\textit{Seq2Edits} framework}
\label{sec:seq2edits}
To alleviate the embarrassed situation of inference speed and large decoding space problems in \textit{Seq2Seq} model architecture, \textit{Seq2Edits} provides another alternative that casts GEC into a tagging problem~\citep{awasthi2019parallel, omelianchuk2020gector, malmi2019encode} along with the non-autoregressive sequence prediction~\citep{li-shi-2021-tail}. 
Instead of directly predicting the token, \textit{Seq2Edits} architecture first predicts the edit operation type $e_{i}\in E$ for each input token $x_{i}$ and then performs a series of transformation operations based on the predicted edit to realize the grammatical output.  
The training objective of tagging is formulated as:
\begin{equation}
    \mathcal{C}(\phi)=-\frac{1}{|\mathcal{D}|}\sum_{x \in D, e \in E}log(p(e|x))
\end{equation}
where $\phi$ corresponds to model parameters to be trained.
This architecture can achieve competitive performance and faster inference speed with limited data but requires heuristic prior and human efforts to obtain labeled data for the tagging task.





\section{Cycle Self-Augmenting Method}
In this section, we introduce our Cycle Self-Augmenting method (\textbf{CSA}).
Specifically, we elaborate on how Self-Augmenting and Cycle Training work under our settings in Section~\ref{sec:self-augment} and Section~\ref{sec:cycle-train} respectively, following with a comparison between our method and conventional data augmentation methods in Section \ref{sec:con_imp}. 
In the cycle training process, we will present the concept of regularization data for GEC, which is the key to the robustness against adversarial attacks.

\begin{figure*}
    \centering
    \includegraphics[scale=0.4]{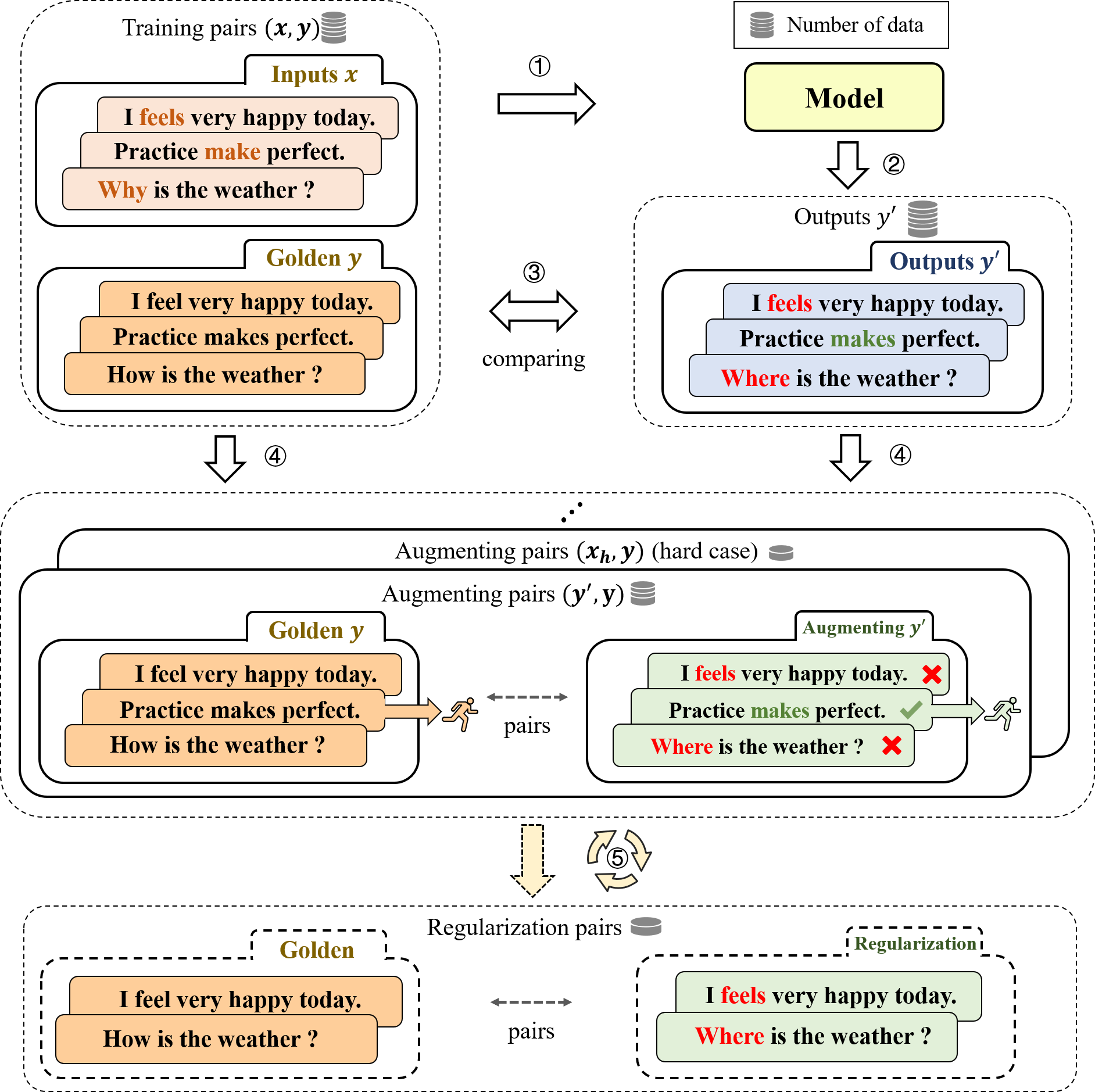}
    \caption{The overall framework of our proposed Cycle Self-Augmenting (CSA). The Self-Augmenting mechanism corresponds with step \ding{172}-\ding{175}. We launch cycle training in step \ding{176}, along with the utilization of regularization data. It is worth noting that the types of augmenting pairs are numerous, and we utilize hard sample $(x_h, y)$ and $(y^{'}, y)$ in our experiment.}
    \label{fig:overall}
\end{figure*}

\subsection{Self-Augmenting}
\label{sec:self-augment}
To enhance model robustness, existing works mainly utilize numerous well-crafted adversarial examples for certain types of adversarial attacks in the pre-training or/and post-training stages~\citep{wang2020improving,li2021searching}.
Instead of elaborately designing adversarial example generation strategies for each type of attack, we make the GEC model itself perform self-augmenting to defend against various types of attack, which is more efficient and adaptive to varied GEC models.
To better exploit the capability of GEC models, we introduce the self-augmenting mechanism  for post-training.

Concretely, the crux of Self-Augmenting is to obtain augmenting data pairs $\mathcal{D}_{Aug}$ which consists of extra pairs $\mathcal{D}_{self}$ constructed by the model itself.
The detailed process is illustrated in Figure~\ref{fig:overall}.
Given a well-trained GEC model $f(\cdot)$ and the original training dataset $\mathcal{D}$=$\{(x^{(i)}, y^{(i)})\}_{i=1\cdots |\mathcal{D}|}$, we feed each input $x$ into $f(\cdot)$ to obtain the corresponded output $y^{'}$ (step \ding{172}-\ding{173} in Figure~\ref{fig:overall}), and compare $y^{'}$ with the golden $y$ word for word (step \ding{174}).
If there is any difference between $y^{'}$ and $y$, we will add $(y^{'},y)$ into $\mathcal{D}_{self}$, and utilize these extra data for post-training (step \ding{175}).

One merit of this is enabling the model to iteratively refine the outputs, which is a popular paradigm in many other situations, e.g., non-auto regressive machine translation task~\citep{lee2018deterministic,geng2021learning,huang2021improving}, and also consistent with the GEC task.
However, it may cause a dramatic forgetting problem.
Thus, we also collect another augmenting data variant $\mathcal{D}_{hard}$ as $\mathcal{D}_{Aug}$ which consists of hard samples directly extracted from the original training set.
The construction of $\mathcal{D}_{hard}$ is similar to the $\mathcal{D}_{self}$, except that we insert each original training pair $(x, y)$ into $\mathcal{D}_{hard}$.
To better distinguish the hard samples from the original training set, we denote pairs in $\mathcal{D}_{hard}$ as $(x_{h}, y)$ in this paper.
We will compare these two strategies in our experiment (Section~\ref{sec:exp}).

Besides, one can collect $(x,y^{'})$ for self-distillation proposed by~ \citeauthor{kim2020self}, which is a regularization method to mitigate the over-fitting problem for the situation of mismatch between the model capacity and the number of data.
However, such method is unsuitable for the GEC task since the the quantity of training data is sufficient and the the target of the GEC model should be grammatical texts.

Instead, utilizing $D_{hard}$ or $D_{self}$ for post-training can provide more feasible training pairs and is more tally with the GEC task, i.e., only part of the input sentence is edited.
We will show the superiority of our self-augmenting in Section~\ref{ana:c2b}.

\subsection{Cycle Training}
\label{sec:cycle-train}
To effectively utilize augmenting pairs from the above introduced self-augmenting process, we further present a cycle training strategy, which is sketched out in step \ding{176} of Figure~\ref{fig:overall}.
Specifically, we use the aforementioned self-augmenting strategy to construct a new dataset $\mathcal{D}_{Aug}^{k}$ in each cycle $k$, where $0\le $$k$$\leq\epsilon$.
Thus, we can leverage $\epsilon$ augmenting datasets ($\mathcal{D}_{Aug}^{1}$,$\dots$,$\mathcal{D}_{Aug}^{k}$,$\dots$,$\mathcal{D}_{Aug}^{\epsilon}$) in cycle training, where these datasets are utilized differently in two training stages.

In \textbf{Stage \uppercase\expandafter{\romannumeral1}}, the obtained augmenting datasets contain many unseen data pairs in the original training dataset, which can be simply used by conducting further training to improve both model performance and robustness. 
Accordingly, we adopt the following training process for each cycle at the early stage, i.e., when $0\le $$k$$\leq\mathcal{P}$:
\begin{itemize}
    \item Perform training on $\mathcal{D}_{Aug}^{k}$ until convergence.
    \item Conduct further tuning on a small high-quality GEC dataset $\mathcal{D}_{tune}$ to prevent over-fitting on the augmenting dataset. 
\end{itemize}
Note that the improvement of performance and robustness is not caused by merely using the small dataset, which is discussed in Section~\ref{ana:effect_CSA}.

Along with the model training, there are fewer and fewer unseen data pairs in the augmenting datasets.
Simply utilizing the augmenting dataset in each cycle for model training might yield over-fitting on these datasets. 
Thus, we turn to focus on these hard-to-learn data, i.e., these data pairs that have not been learned after $\mathcal{P}$ cycles.
Inspired by previous work \citep{zhou2021rethinking} that names some specific samples that are negatively associated with the performance of knowledge distillation as regularization examples, we treat these hard-to-learn data as \textbf{Regularization Data} for the GEC task.
When $\mathcal{P}\leq $$k$$\le\mathcal{\epsilon}$, the regularization data of the $k$-th cycle is obtained as $\mathcal{D}_{Reg}^{k}$=$\mathcal{D}_{Aug}^{k-p+1}\cap\dots\cap\mathcal{D}_{Aug}^{k}$.
In this stage (\textbf{Stage \uppercase\expandafter{\romannumeral2}}), the trained GEC model from \textbf{Stage \uppercase\expandafter{\romannumeral1}} is further trained as below:
\begin{itemize}
    \item Perform training on $\mathcal{D}_{Reg}^{k}$ until convergence.
    \item Conduct further tuning on a small high-quality GEC dataset $\mathcal{D}_{tune}$.
\end{itemize}
We summarize the whole procedure of \textbf{CSA} method into algorithm \ref{alg:csa}.
  
\begin{algorithm}[t]
    \caption{Procedure of \textbf{CSA} method}  
    \label{alg:csa}  
    \begin{algorithmic}[1]   
        \Require max cycle times of \textbf{Stage \uppercase\expandafter{\romannumeral1}} : $\mathcal{P}$;  
        \quad max cycle times of \textbf{Stage \uppercase\expandafter{\romannumeral2}} : $\epsilon$;   
        \quad original training dataset : $(x_{i}, x_{i})$;   
        \quad a small high-quality GEC dataset : $\mathcal{D}_{tune}$;  
        \quad a well-trained GEC model : $g(\theta)$;   
        \Ensure a converged GEC model $g(\theta)$ after cycle self-augmenting training  
        \Function{CSA}{$\mathcal{P}, x_{i}, y_{i}, g(\theta), \epsilon$}  
            \State $k \gets 0$  
            \State $converged \gets False$  
            \State $p \gets True$  
            \While{$k < \mathcal{P}$ \textbf{and} $converged == False$}       
                \If{$k == 0$}  
                    \State $\mathcal{D}^{k}_{Aug} \gets $ \text{Compare}$(y_{i}, g(x_{i};\theta))$  
                \Else  
                    \State utilize $\mathcal{D}^{k}_{Aug}$ \textbf{and} $\mathcal{D}_{tune}$to train model $g(\theta)$   
                    \If{performance of $g(\theta)$ does not improve}  
                        \State $converged \gets True$  
                    \EndIf  
                \EndIf  
            \EndWhile  
            \State $converged \gets False$  
            \While{$k < \epsilon$ \textbf{and} $converged == False$}  
                \State $\mathcal{D}^{k}_{Reg} \gets \mathcal{D}_{Aug}^{k-p+1}\cap\dots\cap\mathcal{D}_{Aug}^{k}$  
                \State utilize $\mathcal{D}^{k}_{Reg}$ \textbf{and} $\mathcal{D}_{tune}$ to train model $g(\theta)$   
                \If{performance of $g(\theta)$ does not improve}  
                    \State $converged \gets True$  
                \EndIf  
                \State $k \gets k + 1$  
            \EndWhile  
            \State   
            \Return{$g(\theta)$}  
        \EndFunction  
    \end{algorithmic}  
\end{algorithm}

The benefits of launching further training on regularization data are four-folds: 1) it prevents over-fitting on the easy-to-learn data pairs in the augmenting datasets; 2) it can reduce model capacity to improve its generalization ability and robustness; 3) it gives more opportunities for the model to address hard-to-learn pairs; 4) it can accelerate each training cycle by using fewer data pairs.
More analysis of regularization data is given in Section~\ref{ana:regular}. 

\subsection{Comparison with Previous Works}
Our CSA method contains a certain degree of correlation in \textbf{Fluency Boost Learning}\citep{ge2018fluency} methods and one may view our method is a traditional data augmentation method. We will elaborate on the advancement of our method from two aspects: (1) continual improvement on a well-trained model (2) a efficient way to filter regularization Data.

\begin{figure}[!tbp]
  \centering
    \subfigure[Dual-boost method]{\includegraphics[width=0.45\textwidth]{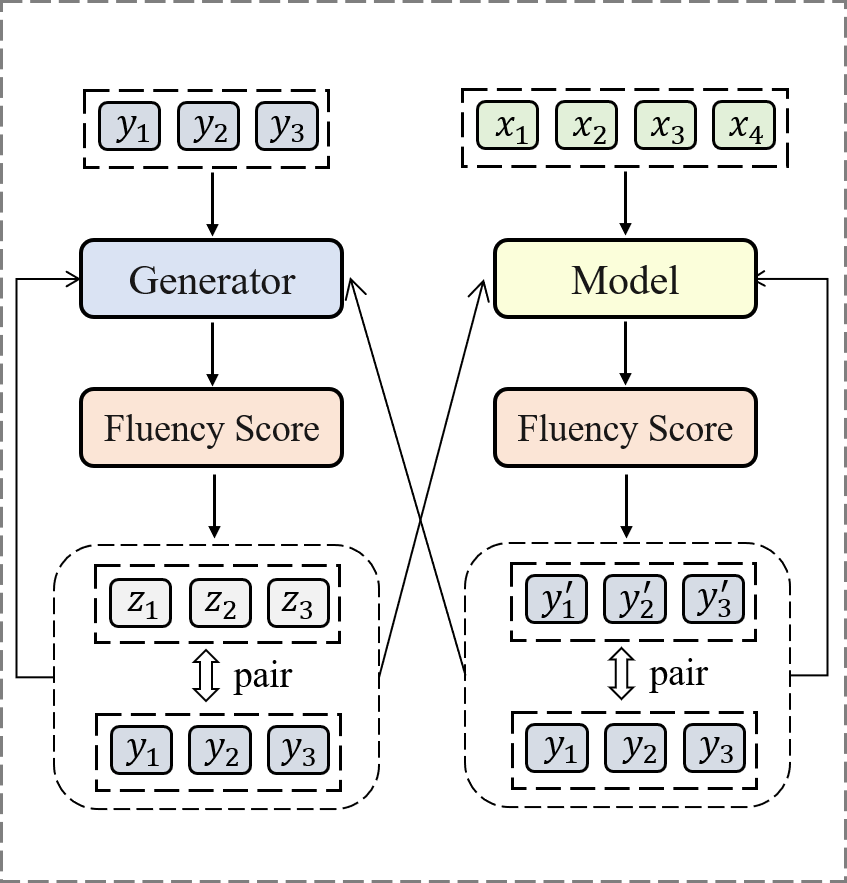}}
    \hspace{0.05\linewidth}
    \subfigure[CSA method]{\includegraphics[width=0.45\textwidth]{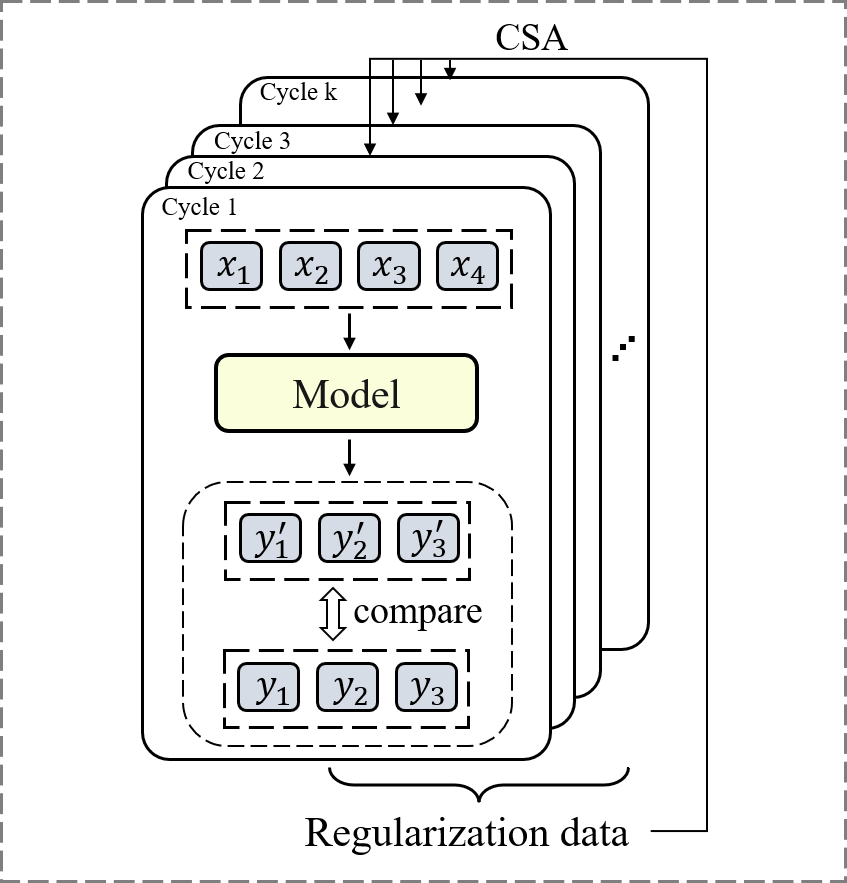}} 
    \caption{Comparison between our \textbf{CSA} with previous data augmentation method. Picture (a) shows the framework of the \textbf{fluency boost learning} method proposed by \citeauthor{ge2018fluency}, which contains a generator that produces error data and a GEC model trained with both real data and the data produced by the generator. The fluency score is introduced to sample the fluency boost pairs output by models. Picture (b) illustrates our \textbf{CSA} method, which contains $k$ cycles, and in each cycle, the GEC model is trained with both read data and the regularization data. The regularization data is selected among the cycles.}
    \label{fig:compara2}
\end{figure}

\paragraph{Continual Improvement}
\label{sec:con_imp}
As shown in Figure \ref{fig:compara2}, unlike the previous work \citep{ge2018fluency} which utilize dual learning method\footnote{There are actually three methods proposed in their work: self-boost, back-boost and dual-boost. 
However, dual-boost method is finally put into implementation.} to improve the GEC model, our method mainly focus on self-iterative refinement on one well trained model. 
From the implementation perspective, we do not modify the original model structure (including decoding strategy) and do not add an auxiliary module.
This plug-and-play method is significant, especially in some environments where the model architecture remains unchanged or has a strict requirement for the number of model parameters.

\paragraph{Regularization Data Extraction} 
\label{sec:effi_re}
A critical merit of our method is to select regularization data that simultaneously improves model performance and robustness among the cyclic procedure. 
Following the aforementioned data augmentation paradigm shown in equation \ref{eq:dataaug} for improving robustness, the regularization data can improve both the model robustness and the performance on the original testing data, as it contains two main components:
\begin{itemize}
    \item $X_{unl}$: A set of data where the model keeps generating the same errors or is unable to edit correctly mainly contributes to improving the model's robustness on attack sets.
    \item $X_{unc}$: A set of data where the model keeps generating different errors or is unable to edit correctly mainly contributes to improving the model's performance on the original testing data.
\end{itemize}
We will analyse the influence brought by regularization data in Section \ref{ana:regular}, and illustrate the procedure of generating attack sets in Section \ref{sec:adversarial_data}.
Moreover, we select different variants of augmenting pairs for post-tuning, i.e., $(x^{'}, y)$ and $(y^{'}, y)$, other than simply filtering one type of data like the fluency boost learning method.

\section{Adversarial Data}
\label{sec:adversarial_data}
Since adversarial data set is the cornerstone of testing the defense ability of the GEC model, we introduce four textual adversarial attack methods to construct different variants for each original test sets, including back-translation~\citep{xie2018noising}, mapping \& rules \citep{wang2020improving}, antonym substitution \citep{ma2019nlpaug} and synonyms substitution~\citep{li2021searching}. 
We divide these four ways of constructing adversarial data into two categories: discrete attack and continuous attack, which are described in Section \ref{sec:dis} and \ref{sec:con}, respectively. 
In Section \ref{sec:rec}, we introduce \textit{Recovery Rate}, one evaluation metric that tests the model's robustness.

\begin{figure}[! tbp]
  \centering
    \subfigure[Synonym]{\includegraphics[width=0.4\textwidth]{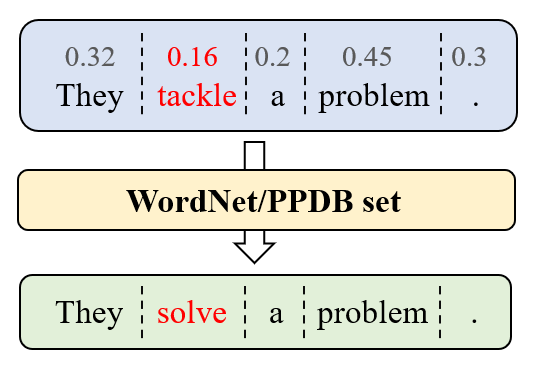}} 
    \hspace{0.1\linewidth}
 \subfigure[Antonym]{\includegraphics[width=0.4\textwidth]{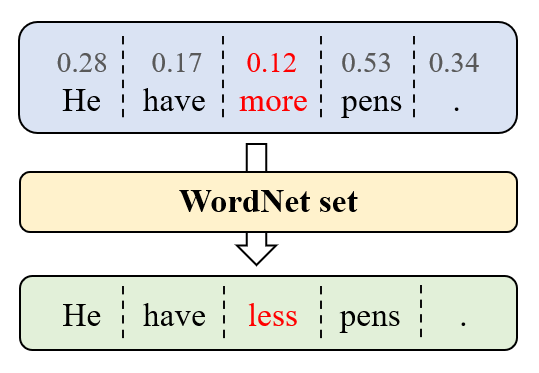}} 
 \vfill
    \subfigure[Mapping \& Rules]{\includegraphics[width=0.4\textwidth]{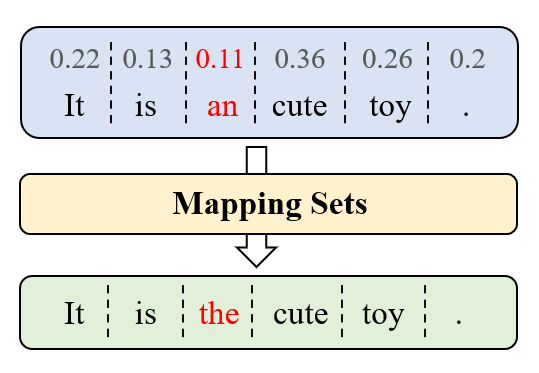}} 
    \hspace{0.1\linewidth}
 \subfigure[Back-Translation]{\includegraphics[width=0.4\textwidth]{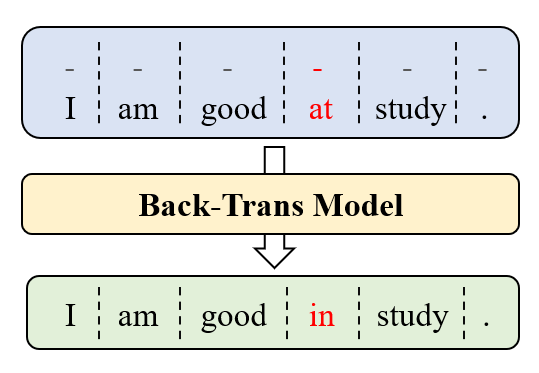}} 
    \caption{Comparison between our \textbf{CSA} and the adversarial training method on four different attack sets.}
 \label{fig:compara1}
\end{figure}

\subsection{Discrete Attack}
\label{sec:dis}
Inspired by the previous work~\citep{wan2020improving} which identifies the vulnerable tokens according to the positional scores, we locate the weak spots with a probability distribution. 
We formally describe the implementation details of identification operation for \textit{Seq2Seq} model and \textit{Seq2Edits} model in Section \ref{sec:con_seq2seq} and \ref{sec:con_seq2edit} respectively.
Then, we introduce three discrete types of substitution operations in Section \ref{sec:sub_op}.

\subsubsection{Constructing data for \textit{Seq2Seq} model architecture}
\label{sec:con_seq2seq}
For \textit{Seq2Seq} model architecture, we use beam search \citep{tillmann2003word} decoding strategy and set beam size as 5.
After the original ungrammatical sentence $X = \{x_{1}, \dots, x_{n} \}$ passes through the well-trained cutting-edge GEC system $\mathcal{F}$, we can obtain 5 candidate sentences, as well as token-level prediction probability for $i$-th token $p^{t}_{i}$.
Then, we select the sentence with the highest generation probability, and
use the alignment function implemented in fairseq toolkit~\citep{ott2019fairseq}
to align the source and target sentences.
We average all the token-level prediction probability to get a threshold $\epsilon$, we take $x_{k}$ as a candidate to be perturbed if its prediction probability $p^{t}_{k}$ is less than the threshold $\epsilon$.

\subsubsection{Constructing data for \textit{Seq2Edits} model architecture}
\label{sec:con_seq2edit}
For \textit{Seq2Edits} model architecture, there is token-level predicting probability for editing token. 
Since the \textbf{Tagger} module (as shown in Figure \ref{fig:model-arch}(b)) often consists of several linear layers with \texttt{softmax} layers on the top \citep{omelianchuk2020gector, awasthi2019parallel, malmi2019encode}, each token-level prediction probability of tag $p^{e}_{i}$ is obtained by applying an \texttt{argmax} over the encoder logits. 
Owing to the natural one-to-one correspondence between the original ungrammatical sentence and the tags, we can calculate the threshold $\epsilon$ and obtain the vulnerable candidates directly without aligning operation.

\subsubsection{Substitution Operation}
\label{sec:sub_op}
After detecting the vulnerable spots for each sentence, a substitution operation is followed.
Details of three common attack strategies, which we call mapping \& rules, antonym substitution, and synonyms substitution, are listed below.
\paragraph{\textit{Mapping \& Rules}} 
We firstly build a $Good \mapsto Poor$ mapping $\mathcal{M}$ from training datasets. 
Specifically, given one ungrammatical sentence and the corresponding correct sentence, we can get the alignment between two sentences by utilizing \texttt{ERRor ANnotation Toolkit}\footnote{\url{https://github.com/chrisjbryant/errant}} \citep{bryant2017automatic, felice2016automatic}. 

Then, we compare the aligned pieces from the source sentence and the golden sentence, respectively, and add the inconsistent piece pairs to the mapping set $\mathcal{M}$.
We extend such operation to the scope of all public annotated corpus (as shown in Table \ref{tab:datasets}).
After acquiring the mapping $\mathcal{M}$, we apply a weighted random sampling \citep{efraimidis2006weighted} method (Equation \ref{eq:sampling}) to select the perturbations for the vulnerable spots in the original sentence step by step:
\begin{equation}
\label{eq:sampling}
    e \sim(\mathbb{U}_{i}) ^ {1/c_{i}}
\end{equation}
where $\mathbb{U}_{i}$ is the random numbers between 0 and 1 for $i$-th token, and $c_{i}$ is calculated by Equation \ref{eq:sim}:

\begin{equation}
\label{eq:sim}
    c_{i}=\left\{
        \begin{aligned}
            & sim(s_{i}, s_{i}^{\prime}) \quad & sim(s_{i}, s_{i}^{\prime}) & \textgreater \zeta  \\
            &  \quad\quad \zeta  \quad & sim(s_{i}, s_{i}^{\prime}) & \leq \zeta
        \end{aligned}
    \right.,
\end{equation}
where $s_{i}$ is the original ungrammatical sentence and its corresponding perturbation is $s_{i}^{\prime}$.
Following the previous work \citep{wang2020improving}, we select edit distance \citep{ristad1998learning} as $sim$ function here, and $\zeta$ is a threshold which controls similarity between two sentences. 
In order to keep the semantic consistency between the original text and adversarial text as much as possible, we must denote that if the edit distance $sim(s_{i}, s_{i}^{\prime})$ continually greater than $\zeta$ for three perturbation steps\citep{wang2020improving}, this strategy ends.
We set $\zeta$ as 0.1 by default.

\paragraph{\textit{Antonym and Synonyms Substitutions.}} 
We detect the vulnerable tokens by using the method proposed above and simply use the open-source tools NLPAug \citep{ma2019nlpaug}\footnote{\url{https://github.com/makcedward/nlpaug}} to substitute opposite meaning word according to WordNet antonym~\citep{miller1995wordnet} or substitute similar words according to WordNet/PPDB~\citep{university2007ppdb} synonym.
The sampling method and default settings of these two perturbation strategies keep up with the \textit{Mapping \& Rules} strategy.

\begin{algorithm}[t]
    \caption{Procedure of \textit{Recovery Rate} metric (\textit{SR})} 
    \label{alg:rec}
    \begin{algorithmic}[1]
        \Require source sentences set: $S$; \quad attacked position set: $P$; \quad golden sentences set: $Y$; \quad model outputs set: $Y^{\prime}$
        \Ensure error\_recovery $er$
        \Function{ErrorRecovery}{$S, P, Y, Y^{\prime} $}
            \State $r \gets 0$
            \State $e \gets 0$
            \State $LY \gets \textbf{Length}(Y)$
            \State $i \gets 0$
            \While{$i < LY$}
                \State $s \gets S[i]$
                \State $pl \gets P[i]$
                \State $y \gets Y[i]$
                \State $y^{\prime} \gets Y^{\prime}[i]$
                \State $editSeq_{1} \gets $ \textbf{Align} $(y, s)$
                \State $editSeq_{2} \gets $ \textbf{Align} $(y^{\prime}, y)$
                \State $LP \gets$ \textbf{Length}$(pl)$
                \State $j \gets 0$
                \State $Flag \gets $ TRUE
                \While{$j < LP$}
                    \State $s\_pos \gets pl[j]$
                    \State $pos_{1} \gets$ \textbf{getPos}$(editSeq_{1}, s\_pos)$
                    \State $pos_{2} \gets$ \textbf{getPos}$(editSeq_{2}, pos_{1})$
                    \If{$y[pos_{1}] \neq y^{\prime}[pos_{2}]$}
                        \State $e \gets e + 1$
                        \State $Flag \gets$ FALSE
                        \State Break
                    \EndIf
                    \State $j \gets j + 1$
                \EndWhile
                \If{$Flag == $ TRUE}
                        \State $r \gets r + 1$
                \EndIf
                \State $i \gets i + 1$
            \EndWhile
            \State $er \gets r / (e+r)$
            \State
            \Return{$er$}
        \EndFunction
    \end{algorithmic}
\end{algorithm}

\subsection{Continuous Attack}
\label{sec:con}
Besides, we also want to explore the robustness of the GEC models by perturbing the inputs in the representation space. 
We utilize the back-translation method~\citep{sennrich2016improving} and the implementation details are illustrated below.
\paragraph{\textit{Back-Translation}}
Inspired by the previous works \citep{sennrich2016improving,edunov2018understanding}, back-translation plays an important role in enlarging the monolingual data and boosting fluency for phrase-based statistical machine translation, as well as GEC tasks. 
We reverse the examples from pre-training datasets to train a back-translation model which can generate errorful examples from a clean corpus.
However, this method generates too few errors when using a traditional decoding strategy, and the outputs cannot confuse the model easily as it samples the most likely token in each decoding step.
Then we implement a variant back-translation method \citep{xie2018noising} which adds $r\beta_{random}$ to penalize every hypothesis during the beam search step, where $r$ is drawn uniformly from the interval [0, 1] and $\beta_{random}$ is a hyper-parameter sampling from $\mathbb{R}$. 
We follow the previous work~\citep{kiyono2019empirical} to set $\beta_{random} = 6$.



\subsection{Recovery Rate to Statistics}
\label{sec:rec}
As long as the current prevalent evaluation metric of the GEC task mainly focuses on the global performance of the post-editing sentences, a specific metric for evaluating partial changes is missing. 
Following the previous work~\citep{wang2020improving}, we propose an evaluation metric Recovery Rate to measure the correction rate of the attacked position.
The entire procedure is described in the algorithm \ref{alg:rec} and the following notations are defined for clearer illustration:

\begin{itemize}
    \item $P$ represents attacked positions of each sentence, which can be automatically generated or manually defined. As for discrete attacks, we regard vulnerable spots of the original text as $P$, while for continuous attacks, $P$ is automatically generated by aligning the attack texts and the original texts. 
    \item $r$ represents the number of recovery sentences. It is worth noting that if all the attack positions are recovered in one sentence, this sentence can be judged as recovery.
    \item $e$ represents the number of unrecovered sentences. If some attack positions are missed to be recovered, this sentence is judged to be unrecovered.
    \item \textbf{Align($X$,$Y$)} is an aligning function implemented by \texttt{ERRor ANnotation Toolkit} which can produce the editing operation sequence between two sentences, i.e., sentence $X$ can transfer to sentence $Y$ by following the editing operation sequence. 
    \item \textbf{getPos($editSeq$,$pos$)} can locate the recovery position by simulating the operations in the editing operation sequence $editSeq$ until reaching the attack position $pos$ and calculating the simulation result's length. 
\end{itemize}

We can see that the \textit{Recovery Rate} metric dictates that all the attack positions per sentence need to be recovered while ignoring the number of attacks which indicates the global modification information and can be abbreviated into \textit{SR} (\textbf{S}entence level \textbf{R}ecovery) rate.
As we are concerned about the local modification, we take each recovery position into consideration, i.e, calculating the \textit{Recovery Rate} of each attacked token rather than totally recovery sentence, and we abbreviate this metric into \textit{TR} (\textbf{T}oken level \textbf{R}ecovery) rate.

To unify the evaluation settings, we also build corresponding datasets which contain a fixed number of errors per sentence.
Specifically, based on the \textit{Mapping \& Rules} substitution operation mentioned in Section \ref{sec:sub_op}, we perturb the fixed number of positions for each sentence by ranking the generation probabilities, selecting the tokens with lower scores and recording the positions $P$.

\section{Experiments}
\label{sec:exp}
We conduct experiments on both original testing sets and attack sets.
We first present necessary details about train sets and evaluation metrics, followed with the description of baselines and concrete implementation settings of all models.
We then give the evaluation results on original testing data and attack sets.

\subsection{Datasets and Evaluations}
In this section, we summarize the training data in Section \ref{sec:traindata} as well as introduce the construction details for attack data in Section \ref{sec:attackdata}, and describe our evaluation metric in Section \ref{sec:evaliuation}.

\subsubsection{Training Datasets}
\label{sec:traindata}
\begin{table}[t]
\centering
\small
\begin{tabular}{l c c c}
    \toprule
    \textbf{Dataset} & \textbf{\#Sentences} & \textbf{Errors (\%)} & \textbf{Usage} \\
    \midrule
    PIE-synthetic & 9,000,000 & 100.0 \% & Pre-training \\ 
    Pseudo-generate & 15,000,000 & 100.0 \% & Pre-training \\ 
    \midrule
    Lang-8 & 1,102,868 & 51.1 \% & Fine-tuning $^\dagger$ \\
    FCE  & 34,490   & 62.6 \% & Fine-tuning $^\dagger$\\
    NUCLE  & 57,151  & 38.2 \% & Fine-tuning $^\dagger$ \\
    W\&I+LOCNESS & 34,308 & 66.3 \% & Fine-tuning $^\ddagger$ \\
    \midrule
    W\&I-Dev (BEA-test) & 4,384 & - & Evaluation \\
    CoNLL-2014 (test) & 1,312 & - & Evaluation \\
    FCE-test  & 2,695 & - & Evaluation \\
    JFLEG & 1,951 & - & Evaluation \\
    \bottomrule
\end{tabular}
\caption{Statistics of datasets used in our experiments. The top group shows the data for pre-training. The second group shows the data for fine-tuning. The third group shows the data for evaluation. ``$\dagger$'' denotes data used in fine-tuning stage \uppercase\expandafter{\romannumeral1}, while ``$\ddagger$'' refers to data used in both fine-tuning stage \uppercase\expandafter{\romannumeral1} and \uppercase\expandafter{\romannumeral2}.}
\label{tab:datasets} 
\end{table}

The first and second group of Table \ref{tab:datasets} describes all the datasets that are utilized in model training.
Following the previous study~\citep{omelianchuk2020gector}, we leverage these datasets in two different training phases:

\paragraph{\textit{Pre-Training}} In this phase, we use 9M pseudo parallel sentences with synthetic errors \citep{awasthi2019parallel}\footnote{\url{https://drive.google.com/open?id=1bl5reJ-XhPEfEaPjvO45M7w0yN-0XGOA}} and pseudo-generated data \citep{kiyono2019empirical} for pre-training. 
Specifically, as for pseudo-generated data, we follow the operation mentioned in \citep{kiyono2019empirical} and use unlabeled corpus from Gigaword\footnote{English Gigaword Fourth Edition (LDC Catalog No.LDC2009T13)}\citep{graff2003english} as seed corpus.
The setting of ($\mu_{mask}$, $\mu_{deletion}$, $\mu_{insertion}$, $\mu_{keep}$) is (0.5, 0.15, 0.15, 0.2). 
We randomly sample 10000 sentences from the training data as the development set and pre-train the model with 15 epochs with the batch size of 9012 tokens.
Based on the results on the development set, we select the best checkpoint for the following stages.

\paragraph{\textit{Fine-Tuning}} During the fine-tuning phase, we use the official corpora from BEA-2019 shared task~\citep{bryant-etal-2019-bea}\footnote{\url{https://www.cl.cam.ac.uk/research/nl/bea2019st/}} for fine-tuning, which comprises four datasets, i.e., Lang-8 Corpus of Learner English (Lang-8) \citep{mizumoto2011mining, tajiri2012tense}, National University of Singapore Corpus of Learner English (NUCLE) \citep{dahlmeier2013building}, the First Certificate in English (FCE) \citep{yannakoudakis2011new}, and Cambridge English Write \& Improve + LOCNESS Corpus (W\&I+LOCNESS) \citep{granger1998prefabricated,yannakoudakis2018developing}.
We split out validation data by random sampling from the official training corpora with a ratio of 2/98 and decompose the fine-tuning phase into two stages. 
In stage \uppercase\expandafter{\romannumeral1}, the model is fine-tuned on errorful-only sentences.
In stage \uppercase\expandafter{\romannumeral2}, the model is tuned on a high-quality and more realistic dataset as in~ \citep{kiyono2019empirical,omelianchuk2020gector}.
In each stage, we set the max training epoch as 20 and select the best checkpoint according to the result on the validation set.

\subsubsection{Attack Datasets}
\label{sec:attackdata}
To evaluate the defense capability of different models, we utilize four perturbation methods mentioned in Section \ref{sec:sub_op} for the official evaluation data (the third group in Table \ref{tab:datasets}) to generate corresponding attack sets and exploit the obtained attack sets for testing.
Specifically, for the back-translation strategy, we utilize the PIE-synthetic dataset for training the model and the training details are listed in Table \ref{tab:back-parameters}.
\begin{table}[!tbp]
\centering
\begin{tabular}{l l}
    \toprule
    Model architecture & transformer\_vaswani\_wmt\_en\_de\_big \cite{vaswani2017attention} \\
    Number of epochs  &  25 \\
    Batch Size (Tokens) & 9012 \\
    Max Sentence Length & 512 \\
    Optimizer         &  Adam \cite{kingma2014adam} \\
    Adam setting      & ($\beta_{1}$=0.9, $\beta_{2}$=0.98) \\
    Learning rate     & 3e-4 \\
    Lr-Scheduler      & inverse\_sqrt \\
    Warm-up           & 4000 \\
    Dropout           & 0.1 \\
    Loss function     & label smoothed cross-entropy \cite{szegedy2016rethinking} \\
    Label Smoothing   & 0.1 \\
    \bottomrule
\end{tabular}
\caption{Hyper-parameters of back-translation model. The whole training strategy follows the pre-training and fine-tuning paradigm, and we utilize the PIE-synthetic datasets in pre-training stage and Lang-8 datasets in fine-tuning stage. We split the validation data with a ratio of 2/98 for model selection. The hyper-parameters of the back-translation model are shown in this table.}
\label{tab:back-parameters}
\end{table}
Moreover, to report the \textit{Recovery Rate}, i.e., \textit{TR} and \textit{SR}, we stipulate the number of errors in each sentence from one to three and build three corresponding variants of CoNLL-2014 evaluation set \footnote{We ignore the sentences in the test set which are shorter than two, e.g., one single quotation mark, both in the perturbation and the calculation procedure.} by applying the method mentioned in Section \ref{sec:rec}.
The datasets created for calculating the \textit{Recovery Rate} are marked as ${\rm ATK}_{i}$, where $i \in \{1,2,3\}$ represents for the number of attack positions per sentence. 

\subsubsection{Evaluations}
\label{sec:evaliuation}
For official evaluation data (original testing sets) and attack datasets build by ourselves (attack datasets), we utilize different evaluation metrics shown below:
\paragraph{Original Testing Sets}
We report results on the official evaluation datasets of BEA, CoNLL-2014~\citep{ng2014conll}, FCE, and JFLEG~\citep{napoles2017jfleg}. 
We measure the results of CoNLL-2014 and FCE by $M^{2}$ scorer~\citep{dahlmeier2012better}. 
For JFLEG results, we use the GLEU metric~\citep{napoles2015ground,napoles2016gleu}. 
We report the scores measured by ERRANT~\citep{bryant2017automatic, felice2016automatic} for BEA-test. 
As the reference of the BEA-test is unavailable, we report results from CodaLab\footnote{\url{https://competitions.codalab.org/competitions/20228}}.

\paragraph{Attack Sets}
As each variant of the attack set is constructed from the original test set, we leverage the same metrics to calibrate model robustness, i.e., $M^{2}$ scorer, GLEU metric, and ERRANT.
We also report \textit{TR} and \textit{SR} score for ${\rm ATK}_{i}$ $(i \in \{1,2,3\})$ set.

\subsection{Baselines and Settings}
Note that our proposed CSA method is a post-training strategy, which can be utilized upon any neural GEC model.
We leverage seven cutting-edge models as our baselines to conduct experiments under the supervised setting. 
To make sure all the baselines are well fine-tuned, if there exists a publicly available checkpoint for each baseline model, we will use it directly.
Otherwise, we will follow the original settings to train a model by ourselves.\footnote{To ensure that the baseline models are fully fine-tuned, we conduct experiments on the original testing datasets and compare the results reported in the original paper. Detailed results are presented in Section \ref{sec:gec_res}}
As for the gradient-based adversarial attack, we set $\tau$ as 1 and $\gamma$ as 0.5.
Specifically, we carry out experiments on four \textit{Seq2Seq} Models, including \textbf{Transformer}~\citep{kiyono2019empirical}\footnote{\url{https://github.com/butsugiri/gec-pseudodata}}, \textbf{BERT-fuse}~\citep{kaneko2020encoder}\footnote{\url{https://github.com/bert-nmt/bert-nmt}}, \textbf{BART} large~\citep{katsumata2020stronger}\footnote{\url{https://github.com/Katsumata420/generic-pretrained-GEC}},\textbf{LM-Critic} method~\citep{yasunaga2021lm}\footnote{We apply LM-Critic under the unsupervised setting and the basic model architecture is Transformer. The implementation is in the  \url{https://github.com/michiyasunaga/LM-Critic}} , and three \textit{Seq2Edit} model variants (\textbf{RoBERTa}, \textbf{BERT}, \textbf{XLNet}) based on large-scale pre-trained language models in GECToR~\citep{omelianchuk2020gector}\footnote{\url{https://github.com/grammarly/gector}}.

Besides, to show the superiority of our CSA method, we utilize one gradient-based defence method~\citep{yasunaga2017robust} for comparison.
Followed by the previous work, we inject the noise into the embedding of each model during the training stage.
More details are shown in Appendix D.

As for the CSA method, we set max cycle times $\epsilon$ = 5 and patience $\mathcal{P}$ = 2. If the model performance does not improve over two consecutive cycles, the training process is stopped.
Note that we use the same official development set throughout the fine-tuning process of baselines and the cycle training process of our CSA method\footnote{\url{https://www.cl.cam.ac.uk/research/nl/bea2019st/}} for checking training convergence.
During the post-training stage, all hyper-parameter settings are the same with baselines.

\subsection{Main Results}

We report three results, including the performance on the original testing sets (Section \ref{sec:gec_res}), the performance on attack sets (Section \ref{sec:att_res}), and the value of \textit{Recovery Rate} on ${\rm ATK}_{i}$ $(i \in \{1,2,3\})$ sets (Section \ref{sec:rec_res}). 
For each baseline, we compare the performance with two CSA variants, i.e., $y^{'} \mapsto y$ and $x_{h} \mapsto y$, and one gradient-based method~($\nabla_{ATK}$).
Moreover, to explore whether our proposed method can adapted to non-English GEC task, we also conduct some experiments on the Chinese GEC task and report the results in Appendix D.

\begin{table*}[t]
\centering
\resizebox{\textwidth}{!}{
    \begin{tabular}{l |c c c| c c c| c c c| c}
    \toprule[1pt]
    \multirow{2}{*}{\textbf{Model}} & \multicolumn{3}{c|}{\textbf{BEA} (ERRANT)} & \multicolumn{3}{c|}{\textbf{CoNLL-2014} ($M^{2}$)} & \multicolumn{3}{c|}{\textbf{FCE} ($M^{2}$)} & \textbf{JELEG~~}\\
    \cmidrule(r){2-4} \cmidrule(r){5-7} \cmidrule(r){8-10} \cmidrule(r){11-11}
      & \textbf{Prec.} & \textbf{Rec.} & \textbf{F\_{0.5}~~} & \textbf{Prec.} & \textbf{Rec.} & \textbf{F\_{0.5}~~} & \textbf{Prec.} & \textbf{Rec.} & \textbf{F\_{0.5}~~} & \textbf{GLEU~~} \\
    \midrule
    
    Transformer~ & 65.5 & 59.4 & 64.2 & 68.9 & 43.9 & 61.8 & 59.4$^*$ & 39.5$^*$ & 54.0$^*$ & 59.7 \\ 
    \textbf{$y^{'} \mapsto y$} (4 Cycles)& 69.6 & 64.7 & \bf68.6 & 69.5 & 49.5 & \bf64.3 & 63.2 & 43.3 & 57.9 & \bf62.7 \\
    \textbf{$x_{h} \mapsto y$} (3 Cycles)& 67.9	& 64.6 & 67.2 & 69.0 &50.1 & 64.2 & 63.0 & 43.9 &	\bf58.0 & 61.8 \\
    \textbf{\bf $\nabla_{ATK}$} (2 Cycles) &66.4 &62.4 &65.6 &69.0 &48.7 &63.7 &61.1 &42.5 &56.2 &61.2   \\
    \midrule

    BERT-fuse~& 67.1 & 60.1 & 65.6 & 69.2 & 45.6 & 62.6 & 59.8 & 46.9 & 56.7 & 61.3 \\ 
    \textbf{$y^{'} \mapsto y$} (3 Cycles) & 68.9 & 64.5 & \bf68.0 & 69.4 & 49.8 & \bf64.4 & 64.4 & 46.6 & \bf59.9 & \bf 62.5 \\
    \textbf{$x_{h} \mapsto y$} (2 Cycles) & 68.1 & 64.0 & 67.2 & 67.5 & 49.4 & 62.9 & 63.7 & 46.9 & 59.4 & 62.0 \\
    \textbf{\bf $\nabla_{ATK}$} (1 Cycle)  &68.1 &62.6 &66.9 &68.1 &50.4 &63.6 &61.4 &43.1 &56.6 &61.4   \\
    \midrule 
    
    BART~ & 68.3 & 57.1 & 65.6 & 69.3 & 45.0 & 62.6 & 59.6$^*$ & 40.3$^*$ & 54.4$^*$ & 57.3 \\ 
    \textbf{$y^{'} \mapsto y$} (2 Cycles) & 70.9 & 61.9 & \bf68.9 & 70.4 & 46.7 & \bf63.9 & 65.2 & 34.4 & \bf55.3 & \bf59.4 \\ 
    \textbf{$x_{h} \mapsto y$} (1 Cycle)&66.8  &61.2  &65.6 &66.3  &45.7  &60.8  &60.1  &39.8  &54.5  & 58.6 \\
    \textbf{\bf $\nabla_{ATK}$} (2 Cycles)  &73.4 &56.5 &69.2 &71.6 &42.6 &63.0 &62.9 &34.7 &54.1 &57.4   \\
    \midrule
	
	LM-Critic~ & 51.6 & 24.7 & 42.4 & 64.4 & 35.6 & 55.5 & 49.6$^*$ & 24.6$^*$ & 41.2$^*$ & 51.4$^*$ \\ 
    \textbf{$y^{'} \mapsto y$} (2 Cycles) & 68.4 & 61.6 & \bf 66.9 & 65.7 & 47.4 & \bf 61.0 & 58.0 & 39.6 & \bf 53.1 & \bf 59.1 \\
    \textbf{$x_{h} \mapsto y$} (1 Cycle)&69.2 &49.6 &64.2 &64.8 &36.4 &56.1 &59.9 &30.7 &50.3 &55.1 \\
    \textbf{\bf $\nabla_{ATK}$} (1 Cycle) &63.2 &56.0 &61.6 &61.9 &42.4 &56.7 &57.0 &35.3 &50.7 &56.6   \\
    \midrule
    
    BERT& 71.5 & 55.7 & \bf 67.6 & 72.1 & 42.0 & \bf 63.0 & 66.2$^*$ & 42.0$^*$ & \bf 59.4$^*$ & 57.5$^*$ \\ 
    \textbf{$y^{'} \mapsto y$} (1 Cycle) & 67.7 & 57.2 & 65.3 & 70.0 & 44.3 & 62.3 & 64.0 & 43.1 & 58.3 & 57.8 \\
    \textbf{$x_{h} \mapsto y$} (1 Cycle) & 66.4 & 56.0 & 64.0 & 68.7 & 42.8 & 61.3 & 63.8 & 42.8 &	58.1 & \bf 58.4 \\
    \textbf{\bf $\nabla_{ATK}$} (1 Cycle) &66.1 &57.8 &64.3 &66.3 &44.5 &60.4 &57.9 &36.5 &51.8 &57.5   \\
    \midrule
    
    RoBERTa &68.4 &60.8 &66.8 &68.7 &47.2 &\bf62.9 &61.6$^*$ &45.3$^*$ &57.5$^*$ &59.1$^*$ \\ 
    \textbf{$y^{'} \mapsto y$} (1 Cycle) & 68.8 & 60.3 &\bf66.9 &68.0 &46.9  &62.4  &62.7 &44.8  &\bf58.0  & 58.6 \\
    \textbf{$x_{h} \mapsto y$} (1 Cycle) & 66.2 & 60.4 & 64.9 & 66.3 & 47.7 & 61.5 & 61.4 & 44.8 & 57.2 & \bf59.2	\\
    \textbf{\bf $\nabla_{ATK}$} (1 Cycle) &64.2 &61.7 &63.7 &64.4 &49.1 &60.6 &56.2 &40.1 &52.0 &59.0   \\
    \midrule
    
    XLNet~& 79.2 & 53.9 & \bf 72.4 & 77.5 & 40.1 & \bf 65.3 & 71.9$^*$ & 41.3$^*$ & 62.7$^*$ & 56.0$^*$ \\ 
    \textbf{$y^{'} \mapsto y$} (1 Cycle) & 77.8 & 55.0 & 71.8 & 75.3 & 41.6 & 64.8 & 71.5 & 42.7 & \bf 63.1 & 56.5 \\
    \textbf{$x_{h} \mapsto y$} (1 Cycle) & 65.9 & 62.7 & 65.3 & 66.7 & 48.6 & 62.1 & 64.5 & 51.1 &	61.3 & \bf 60.1	\\
    \textbf{\bf $\nabla_{ATK}$} (1 Cycle) &67.9 &60.0 &66.1 &67.2 &47.5 &62.0 &59.4 &38.5 &53.6 &59.0   \\
    \bottomrule
\end{tabular}}
\caption{Evaluation results on the original testing data.
The numbers labeled with ``\textbf{*}'' refer to the results tested by ourselves with the released checkpoints from the original papers, while all the left numbers are copied from the original papers.
We report the performance of two variants of regularization data for each model and the corresponding best cycle times, where $\mapsto$ represents the direction of data flow.
The bold fonts are the best performance in each comparison. 
Note that all the non-CSA baselines here are fine-tuned on the high-quality fine-tuning set (except for LM-Critic, which is an unsupervised method). The CSA and its variants are trained from the same checkpoints as baselines.}
\label{tab:main_gec_res} 
\end{table*}

\begin{table}[t]
\resizebox{\textwidth}{!}{
    \begin{tabular}{l |c c c| c  c c| c c c| c}
    \toprule[1pt]
    \multirow{2}{*}{\textbf{Model}} & \multicolumn{3}{c|}{\textbf{BEA~} (ERRANT)} & \multicolumn{3}{c|}{\textbf{CoNLL-2014~} ($M^{2}$)} & \multicolumn{3}{c|}{\textbf{FCE~} ($M^{2}$)} & \textbf{JFLEG~~}\\
    \cmidrule(r){2-4} \cmidrule(r){5-7} \cmidrule(r){8-10} \cmidrule(r){11-11}
      & \textbf{Prec.} & \textbf{Rec.} & \textbf{F\_{0.5}~~} & \textbf{Prec.} & \textbf{Rec.} & \textbf{F\_{0.5}~~} & \textbf{Prec.} & \textbf{Rec.} & \textbf{F\_{0.5}~~} & \textbf{GLEU~~} \\
    \midrule
    
    Transformer~ &21.0 &48.0 &23.4 &34.1 &39.7 &34.9 &29.7 &34.2 &30.3 &45.4 \\
    \textbf{$y^{'} \mapsto y$} (4 Cycles) &23.9	&53.3	&26.6	&37.6	&45.5	&38.8	&32.5	&38.8	&33.4	&46.4 \\
    \textbf{$x_{h} \mapsto y$} (3 Cycles) & 23.8 & 53.6 & \bf26.5 & 38.2 & 45.9 &	\bf 39.4 & 32.5 &	38.8 & \bf 33.8 & \bf47.2 \\
    \textbf{\bf $\nabla_{ATK}$} (2 Cycles)&22.9	&51.8	&25.6	&34.9	&39.7	&35.7	&30.1	&35.2	&30.6	&44.8 \\
    \midrule
    
    BERT-fuse~ &20.4 &46.1 &22.6 &33.5 &38.2 &34.1 &31.0 &34.5 &31.4 &45.4 \\
    \textbf{$y^{'} \mapsto y$} (3 Cycles) &23.9	&53.7	&26.6	&37.9	&45.4	&39.0	&33.7	&39.9	&34.6	&47.0 \\
    \textbf{$x_{h} \mapsto y$} (2 Cycles) & 23.8 & 53.6 & 26.5 & 37.4 & 45.5 &  38.7 & 33.9 & 40.6 &	\bf34.9 & \bf47.2 \\
    \textbf{\bf $\nabla_{ATK}$} (1 Cycles)&24.5	&54.8	&27.3	&38.6	&46.1	&39.7	&33.1	&39.5	&33.9	&47.3 \\
    \midrule
    
    BART~ &20.9 &44.7 &23.0 &34.5 &38.8 &35.0 &30.1 &31.5 &30.0 &43.8 \\
    \textbf{$y^{'} \mapsto y$} (2 Cycles) &25.0	&53.9	&27.7	&39.1	&46.0	&40.1	&32.1	&37.5	&32.9	&45.8 \\
    \textbf{$x_{h} \mapsto y$} (1 Cycle)&23.4 &51.4 &25.9 &36.7 &43.4 &37.7 &31.8 &36.3 &32.3 &45.1 \\
    \textbf{\bf $\nabla_{ATK}$} (2 Cycles)&23.8	&50.8	&26.2	&36.5	&43.3	&37.6	&31.0	&32.6	&30.8	&44.6 \\
    \midrule
    
    LM-Critic~ &18.6 &39.0 &20.5 &34.5 &35.9 &34.5 &23.6 &24.7 &23.5 &41.1 \\
    \textbf{$y^{'} \mapsto y$} (2 Cycles) &24.5	&52.1	&27.1	&41.1	&46.0	&41.8	&31.9	&36.3	&32.5	&46.1 \\
    \textbf{$x_{h} \mapsto y$} (1 Cycle) &20.6  &43.3  &22.6 &31.7  &35.3  &32.1  &28.2  &29.4  &28.1  &42.9 \\
    \textbf{\bf $\nabla_{ATK}$} (1 Cycles) &22.8	&41.3	&24.5	&36.3	&37.7	&36.3	&30.0	&32.7	&30.2	&44.1 \\
    \midrule
    
    BERT~ &23.1 &50.2 &25.7 &35.6 &42.9 &37.4 &33.2 &39.4 &34.2 &45.7 \\
    \textbf{$y^{'} \mapsto y$} (1 Cycle) &23.4	&51.3	&26.1	&37.4	&41.9	&38.0	&33.7	&40.1	&34.6	&45.8 \\
    \textbf{$x_{h} \mapsto y$} (1 Cycle) & 22.0 &47.8 &	24.4 &35.2 & 40.4 &	35.9 &32.5 &37.5 &33.2 &45.3 	\\
    \textbf{\bf $\nabla_{ATK}$} (1 Cycles) &23.8	&50.8	&26.3	&36.8	&43.0	&37.7	&30.2	&35.2	&31.0	&45.6 \\
    \midrule
    
    RoBERTa~ &24.8 &52.4 &27.4 &38.2 &44.1 &39.0 &33.9 &39.9 &34.8 &46.3 \\
    \textbf{$y^{'} \mapsto y$} (1 Cycle) &25.0	&52.7	&27.6	&38.7	&44.6	&39.5	&34.3	&40.3	&35.1	&46.5 \\
    \textbf{$x_{h} \mapsto y$} (1 Cycle) & 24.7 & 52.6 & 27.3 & 38.5 & 45.0 & 39.4 & 34.0 &	40.2 & 34.9 & 46.4 \\ 
    \textbf{\bf $\nabla_{ATK}$} (3 Cycles)&25.0	&52.5	&27.6	&38.2	&44.6	&39.6	&32.2	&37.0	&32.9	&46.3 \\
    \midrule
    
    XLNet~ &25.7 &54.6 &28.4 &38.9 &46.6 &40.1 &36.5 &44.9 &37.7 &47.3 \\
    \textbf{$y^{'} \mapsto y$} (2 Cycles) &25.8	&54.8	&28.5	&39	&46.6	&40.1	&36.6	&44.9	&37.9	&47.5 \\
    \textbf{$x_{h} \mapsto y$} (1 Cycle) & 25.0 & 53.0  & 27.6 & 38.1 &	45.0 & 39.1 & 36.3 & 44.0 & 37.4 & 47.3 \\
    \textbf{\bf $\nabla_{ATK}$} (1 Cycles) &27.2	&50.0	&28.3	&38.9	&45.8	&39.8	&32.5	&36.5	&33.1	&46.5 \\
    \bottomrule
    \end{tabular}
}
\caption{The average of evaluation results on four attack sets and each test set corresponds to four variants for attack, including \textit{Mapping \& Rules}, \textit{Synonyms}, \textit{Back-Translation} and \textit{Antonym}. 
We report the performance of two variants of regularization data for each model and the corresponding best cycle times, where $\mapsto$ represents the direction of data flow.
The bold fonts indicate the optimal performance of each comparison.}
\label{tab:attack-avg} 
\end{table}

\begin{table}[hbt!]
\centering
\small
\begin{tabular}{l | c c| c c| c c}
\toprule[1pt]
\multirow{2}{*}{\textbf{Model}} & \multicolumn{2}{c|}{\textbf{ATK 1} ($\uparrow$)} & \multicolumn{2}{c|}{\textbf{ATK 2} ($\uparrow$)} & \multicolumn{2}{c}{\textbf{ATK 3} ($\uparrow$)} \\
\cmidrule(r){2-3} \cmidrule(r){4-5} \cmidrule(r){6-7} 
  &  \textbf{\textit{TR}(\%)} & \textbf{\textit{SR}(\%)} & \textbf{\textit{TR}(\%)} & \textbf{\textit{SR}(\%)} & \textbf{\textit{TR}(\%)} & \textbf{\textit{SR}(\%)} \\
\midrule

Transformer &32.6 &32.6 &23.5 &13.3 &16.1 &4.1 \\
\textbf{$y^{'} \mapsto y$} (4 Cycles) & \bf37.7 & \bf37.7 & 26.7 & 15.9 & \bf18.7 & \bf5.4 \\
\textbf{$x_{h} \mapsto y$} (4 Cycles) & 35.6 & 35.6 & \bf28.2 & \bf17.6 & 18.2 & 5.4  \\
\midrule

BERT-fuse &33.6 &33.6 &25.0 &15.0 &15.1 &3.7 \\
\textbf{$y^{'} \mapsto y$} (3 Cycles) &\bf36.0 &\bf36.0 &\bf26.6 &\bf15.4 &\bf17.9 &\bf4.9 \\
\textbf{$x_{h} \mapsto y$} (2 Cycles)& 35.1 & 35.1 & 26.5 & 15.4 & 17.8 & 4.9 \\
\midrule

BART &38.2 &38.2 &27.2  &16.6  &18.1 &5.3 \\
\textbf{$y^{'} \mapsto y$} (2 Cycles) &\bf38.5 &\bf38.5 &\bf28.6  &\bf18.1 &\bf18.6 &\bf5.3 \\
\textbf{$x_{h} \mapsto y$} (2 Cycles) & 38.2 & 38.2  & 27.0 & 16.9 & 17.8 & 5.2  \\
\midrule

LM-critic & 34.5 & 34.5& 23.5 &12.9 &15.8 &4.6 \\
\textbf{$y^{'} \mapsto y$} (2 Cycles) &\bf34.9 &\bf34.9 &24.4 &14.4 &\bf16.4 &\bf5.0 \\
\textbf{$x_{h} \mapsto y$} (1 Cycle)&34.5  &34.5  &\bf26.0 &\bf15.8  &16.1  &4.2  \\
\midrule

RoBERTa &36.9 &36.9 & 26.5 & 16.1 &16.9 &4.9 \\
\textbf{$y^{'} \mapsto y$} (1 Cycle) & 36.9 & 36.9 & 25.3 & 15.1 & 16.6 & 4.9 \\
\textbf{$x_{h} \mapsto y$} (1 Cycle) & \bf37.1 & \bf37.1 & \bf26.9 & \bf16.3 & \bf17.9 & \bf6.0 \\
\midrule

BERT &\bf33.4 &\bf33.4 &23.2 &13.4 &15.9 &4.1 \\
\textbf{$y^{'} \mapsto y$} (1 Cycle) & 33.3 & 33.3 &\bf23.3 &\bf13.7 &\bf16.7 &\bf4.5 \\
\textbf{$x_{h} \mapsto y$} (1 Cycle) & 31.7	& 31.7 & 22.8 &	 13.3 & 15.8 & 4.2  \\
\midrule

XLNet & 35.5 &35.5 &25.8 &15.4 & \bf17.7 & \bf5.1 \\
\textbf{$y^{'} \mapsto y$} (2 Cycles) & \bf36.3 & \bf36.3 & \bf26.3 & \bf15.7 & 8.1 & 5.0 \\
\textbf{$x_{h} \mapsto y$} (2 Cycles) & 35.7 & 35.7 & 25.4 & 15.4 & 17.7 & 4.9 \\

\bottomrule

\end{tabular}
\caption{Recovery Rate for ${\rm ATK}_{i}$ $(i \in \{1,2,3\})$ sets. 
We report both the token level recovery rate (\textit{TR}) and the sentence level recovery rate (\textit{SR}), and the cycle times are correlated with the results on clean data (Table \ref{tab:main_gec_res}).
We report the performance of two variants of regularization data for each model and the corresponding best cycle times, where $\mapsto$ represents the direction of data flow.
The bold fonts indicate the optimal performance of each comparison.}

\label{tab:recov_rate} 
\end{table}

\subsubsection{GEC Results}
\label{sec:gec_res}
We first present the experimental results on four original testing sets to calibrate the influence of our proposed CSA to baseline models, where the detailed numbers are shown in Table~\ref{tab:main_gec_res}. 
It can be seen that with several additional cycles, our proposed CSA method yields impressive performance improvement on four baselines, i.e., Transformer, BERT-fuse, BART, and LM-Critic.
Specifically, the improvement is about two points for the supervised setting (Transformer, BERT-fuse, BART). For the unsupervised setting (LM-Critic), the improvement is dramatic, e.g., 19.2 for the BEA test set.
For the rest three strong baselines, our proposed method can achieve nearly comparable results\footnote{Our method achieves better performance on six out of twelve settings, where each setting refers to the combination of a specific baseline and a dataset, e.g., CSA achieves higher F\_0.5 score than the BERT model on the JELEG dataset.}.

\subsubsection{Attack Results}
\label{sec:att_res}
The results of evaluation results on four different attack sets are put in Appendix A.
In Table~\ref{tab:attack-avg}, we report the average evaluation score on attack sets. 
Recall that we construct four variants of attack sets with different methods for each original test set.
To better show the effectiveness of our proposed CSA method, we utilize the averaged results of four attack sets for each original test set.
It can be observed that our proposed CSA method yields robustness improvement on all baseline methods.
In particular, our CSA methods leads to improvements on all attack sets, and there existing dramatical increase of the average score under the $y^{'} \mapsto y$ setting, i.e., the improvements of 4.9 ($F_{0.5}$) points over BERT-fuse and 5.1 ($F_{0.5}$) points over the BART model on the CoNLL-2014 attack sets.
However, it is worth noting that the original \textit{Seq2Edit} models all have a better performance on attack sets than the \textit{Seq2Seq} models and even can compete with the performance of some \textit{Seq2Seq} models, which are reinforced by CSA method.

\subsubsection{Recovery Rate Results}
\label{sec:rec_res}
The results of \textit{Recovery Rate} are shown in Table \ref{tab:recov_rate}.
We report the \textit{TR} (token level recovery rate) and \textit{SR} (sentence level recovery rate) for each model.
It can be observed that the performance of the \textit{Seq2Seq} model (Transformer, BERT-fuse, BART and LM-Critic) is obviously better than that of the \textit{Seq2Edit} model (i.e., including RoBERTa, BERT, and XLNet~).
We can also see that, with the increasing number of attack positions, the results of both \textit{TR} and \textit{SR} decrease a lot, and the improvement of CSA method is diminishing.
Although the performance degradation in this \textit{Seq2Edit} model is not extensive in magnitude, the slight improvement and unstable performance of \textit{Seq2Edit} models attract our interests, and we will dig for the reasons behind it subsequently.

\begin{figure}[t]
  \centering
    \subfigure[Antonym]{\includegraphics[width=0.45\textwidth]{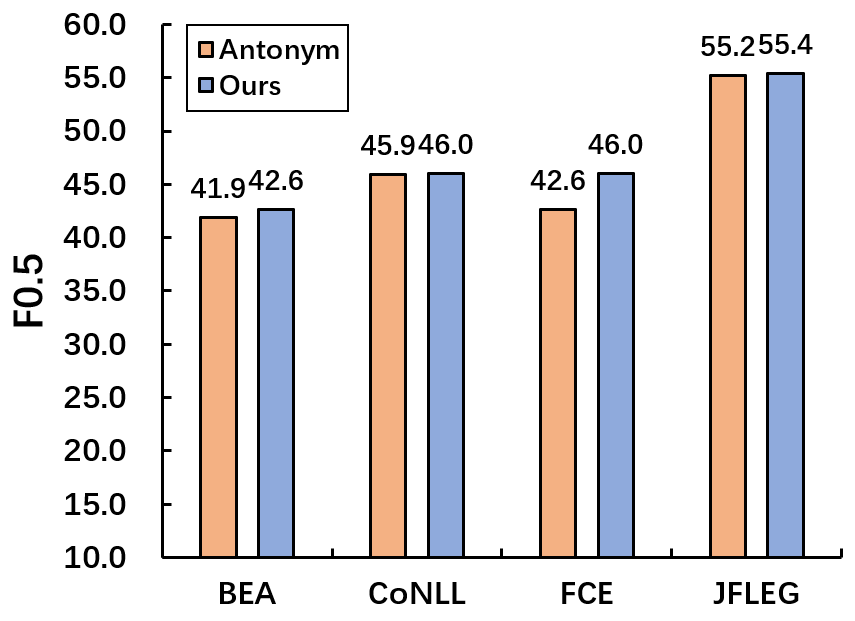}} 
    \hspace{0.01\linewidth}
	\subfigure[Synonym]{\includegraphics[width=0.45\textwidth]{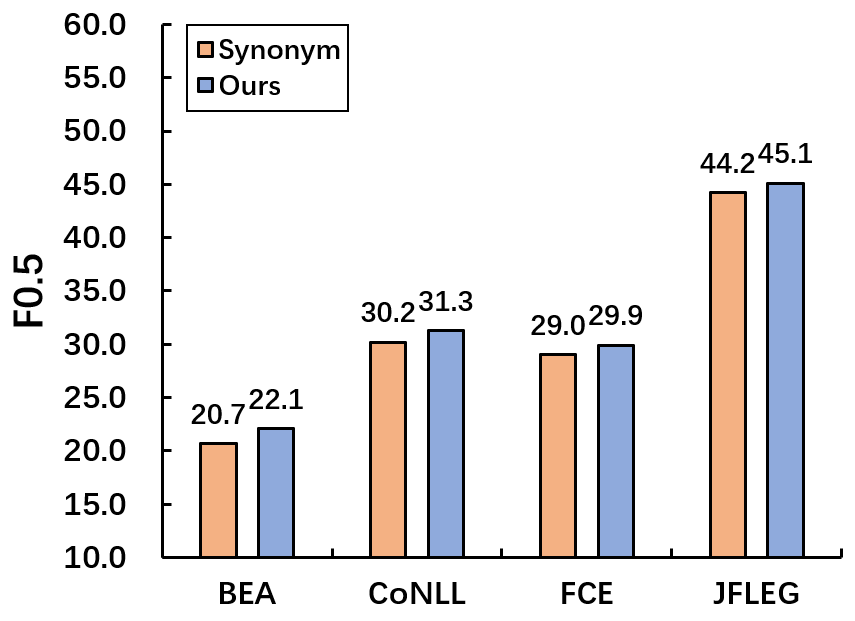}} 
	\vfill
    \subfigure[Mapping \& Rules]{\includegraphics[width=0.45\textwidth]{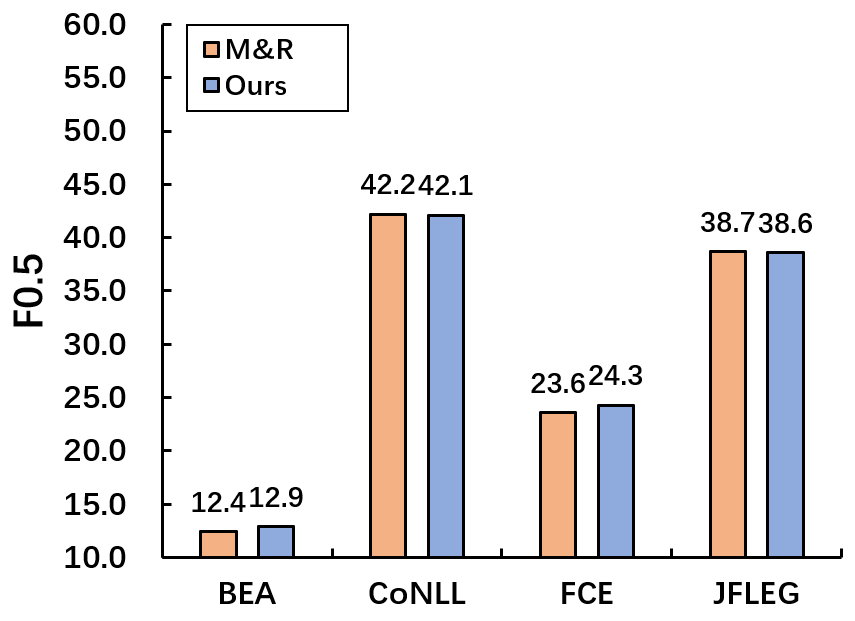}} 
    \hspace{0.01\linewidth}
	\subfigure[Back-Translation]{\includegraphics[width=0.45\textwidth]{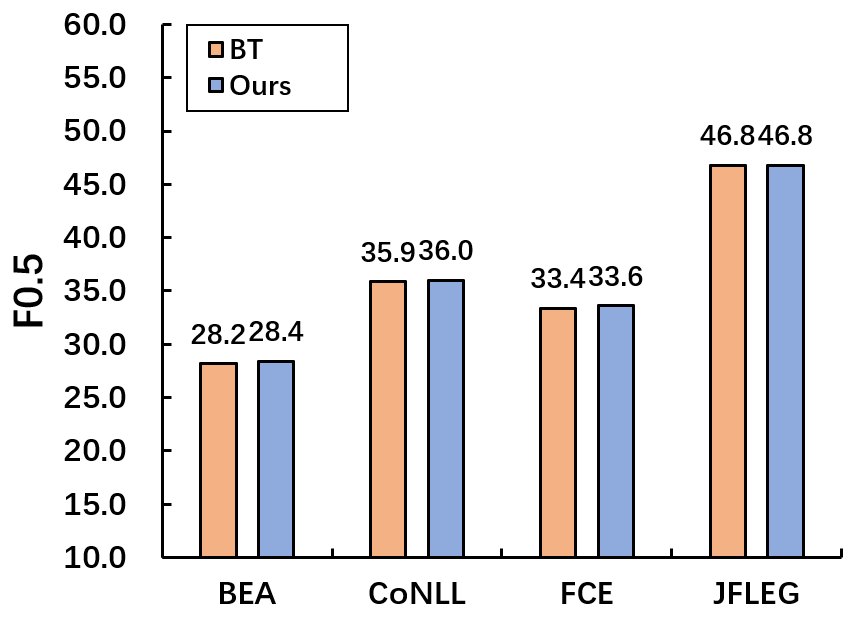}} 
    \caption{Comparison between our \textbf{CSA} and the adversarial training method on four different attack sets. In each sub-figure, the orange straight square represents for the traditional defence method, while blue one represents for our \textbf{CSA} method. }
	\label{fig:compara1}
\end{figure}

\section{Analysis and Discussion}
\label{sec:analysis}
In this section, we conduct extensive studies from different perspectives to better understand our \textbf{CSA} method and regularization data.
Since the most important merit of our CSA method is simplicity and effectiveness in improving the robustness against multiple types of attack, we first evaluate its defense capability of \textbf{CSA} with a recently proposed defense method for the GEC task \citep{wan2020improving} in Section \ref{sec:defence_com}. 
Then, we conduct experiments to study the effects of self-augmenting and hard samples in Section \ref{ana:c2b}, followed by hyper-parameter analysis in Section \ref{ana:effect_CSA}. 
Afterward, we conduct a case study of regularization data, elaborate the influence brought by two data components $X_{unl}$ and $X_{unc}$ decomposed from regularization data, and try different variations of our \textbf{CSA} method in Section \ref{ana:regular}.
Finally, we analyze the reason behind the slight improvement of \textit{Seq2Edits} framework in Section \ref{sec: generalization}.
These studies are mainly taken on CoNLL-2014 dataset and the attack set is constructed by the Mapping \& Rules method unless there is a clear explanation, and all the experiments are launched on the Transformer.

\subsection{Large Langeage Models Robustness}
\label{sec:llm}
We evaluate the robustness of GPT-3.5~(text-davinci-003) using the original CoNLL-2014 testing data and its three types of attack sets. Given that the format and the exact wording of GPT-3.5's prompts can significantly impact the task performance, we conduct zero-shot and few-shot prompt setting proposed by previous work~\citep{coyne2023analysis}, which performs best in prompt engineering experiments.
The results are listed in Table~\ref{tab:preliminary_main}, indicate that GPT-3.5 suffers from a significant decrease in performance on the attack sets, e.g., from 34.8 to 28.4  on the Vector-based set in few-shot setting. Despite being trained on a vast amount of data, GPT-3.5 is still susceptible to a reduction in its robustness.
\begin{table*}[t]
\small
\centering
\begin{tabular}{l | ccc | ccc | ccc}
\toprule
\multirow{2}{*}{\bf Data} & \multicolumn{3}{c|}{\bf Transformer} & \multicolumn{3}{c|}{\bf Davinci-003 $\dagger$} & \multicolumn{3}{c}{\bf Davinci-003 $\ddagger$}  \\
\cmidrule{2-10}
& \textbf{Prec.} & \textbf{Rec.} & \textbf{F\_{0.5}~~}& \textbf{Prec.} & \textbf{Rec.} & \textbf{F\_{0.5}~~} & \textbf{Prec.} & \textbf{Rec.} & \textbf{F\_{0.5}~~} \\
\midrule
Origin & 68.9 & 43.9 & 61.8 &31.5  &52.3  &34.2 &32.1 & 52.3 & 34.8   \\
Mapping \& Rules &36.5 &37.8 &36.7 &27.6  &48.2  &28.1 &28.0 &48.0 &30.6    \\
Vector-Based &26.5 &36.0 &28.0 &27.6  &48.2  &30.2 &25.9 &46.6 &28.4    \\
Back-Translation &33.0 &48.9 &35.3 &26.4  &50.0  &29.1 &26.5 &49.7 &29.2    \\
\bottomrule
\end{tabular}
 \caption{Experimental results on CoNLL-2014 testing data and its correlated attack set, where $\dagger$ represents the zero-shot setting and $\ddagger$ represents the few-shot setting.}
\label{tab:preliminary_main}
\end{table*}

\subsection{Defense Capability Comparison}
\label{sec:defence_com}
previous adversarial training method~\citep{wan2020improving}, which add Mapping \& Rules attack samples into the training set.
Figure~\ref{fig:compara1} presents the evaluation results on four test sets under the four types mentioned above of adversarial attack.
Our CSA has better defense capability than the baseline model under three types of attack and achieves comparable results under the Mapping \& Rules attack.
In other words, our CSA can achieve competitive results with the previous defense method, which uses many well-crafted adversarial examples for training.
Our CSA achieves much better model robustness for other attacks without specifically designed adversarial training examples. 
These results demonstrate the effectiveness and generalization ability of our method.

\begin{table}[!tbp]
\centering
\small
\begin{tabular}{ c | c | c | c c c }
    \toprule
    \multirow{2}{*}{\textbf{Strategies}} &\multirow{2}{*}{\textbf{\#Cycles}} & \multirow{2}{*}{\textbf{\#Pairs}} & \multicolumn{3}{c}{\textbf{CoNLL-2014 ($M^{2}$)~}} \\
    \cmidrule(r){4-6}
     & & & \textbf{Prec.} & \textbf{Rec.} & \textbf{F\_{0.5}~~} \\
    \midrule
    \multirow{3}{*}{$x \mapsto y^{'}$} & 0 & 625,467 & 68.9 & 43.9 & 61.8 \\
    & 1 & 511,006 & 66.5 & 51.1 & 62.6 \\
    & 2 & 436,229 & 65.2 & 46.3 & 60.3 \\
    \midrule
    \multirow{3}{*}{$y^{'} \mapsto y$} & 0 & 625,467 & 68.9 & 43.9 & 61.8 \\
    & 1 & 506,572 & 67.2 & 49.4 & 62.6 \\
    & 2 & 263,993 & 68.3 & 50.3 & 63.8 \\
    \midrule
    \multirow{3}{*}{$x_{h} \mapsto y$} & 0 & 625,467 & 68.9 & 43.9 & 61.8 \\
    & 1 & 24,213 & 68.0 & 50.9 & 63.7 \\
    & 2 & 466,295 & 68.6 & 50.1 & 63.9 \\
    \bottomrule
    \end{tabular}
    \caption{Results of two strategies for self-augmenting.
The first group refers to the strategy of self-distillation by using failed pairs ($x \mapsto y^{'}$) to re-train the model.
The second and third group corresponds to our strategy of using the model outputs to construct ($y^{'} \mapsto y$) pairs and using hard samples to construct ($x_{h} \mapsto y$).}
\label{tab: AB-training}
\end{table}
\begin{table*}[hbt!]
\centering

\small
\resizebox{\textwidth}{!}{
    \begin{tabular}{l | c c c | c c c | c c | c c | c c}
    \toprule[1pt]
    \multirow{2}{*}{\textbf{Model}} & \multicolumn{3}{c|}{\textbf{CoNLL (CLN)}} & \multicolumn{3}{c|}{\textbf{CoNLL (ATK)}} & \multicolumn{2}{c|}{\textbf{ATK 1}} & \multicolumn{2}{c|}{\textbf{ATK 2}} & \multicolumn{2}{c}{\textbf{ATK 3}}\\
    
    \cmidrule(r){2-4} \cmidrule(r){5-7} \cmidrule(r){8-9} \cmidrule(r){10-11} \cmidrule(r){12-13}
      & \textbf{Prec.~~} & \textbf{Rec.~~}& \textbf{F\_{0.5}} &  \textbf{Prec.~~} & \textbf{Rec.~~}& \textbf{F\_{0.5}} &  \textbf{TR} & \textbf{SR} & \textbf{TR} & \textbf{SR} & \textbf{TR} & \textbf{SR}  \\
    \midrule
    
    Transformer~ & 68.9 & 43.9 & 61.8 & 34.1 & 39.7 & 34.9 & 32.6 &32.6 &23.5 &13.3 &16.1 &4.1 \\
    \textbf{$y^{'} \mapsto y$} (CSA)& 69.5 & 49.5 & 64.3 & 37.7 & 45.5 & 38.9 & \bf37.7 &\bf37.7 & 26.7 & 15.9 & 18.7 & 5.4 \\
    \textbf{$x_{h} \mapsto y$} (CSA)& 69.0 & 50.1 & 64.2 & 38.1 & 45.6 & 39.2 &  35.6 &	35.6 & \bf28.2 &\bf17.6 & 18.2 & 5.4  \\
    \textbf{$x_{h} \mapsto y$} (DIR)& 69.4 & 50.8 & \bf64.6 & 38.6 & 46.1 & \bf39.7 & 35.9 & 35.9 & 27.2 & 16.1 & \bf18.3 & \bf5.8\\
    \midrule

    BERT-fuse~ & 69.2 & 45.6 & 62.6 & 33.5 & 38.2 & 34.1 & 33.6 &33.6 &25.0 &15.0 &15.1 &3.7  \\
    \textbf{$y^{'} \mapsto y$} (CSA)& 69.4 & 49.8 & \bf64.4 & 37.9 & 45.5 & \bf39.0 & \bf36.0 &\bf36.0 & 26.6 & 15.4 & 17.9 & 4.9   \\
    \textbf{$x_{h} \mapsto y$} (CSA)& 67.5 & 49.4 & 62.9  & 37.4  & 45.5  &	38.7 & 35.1 & 35.1 & 26.5 & 15.4 & 17.8 & 4.9  \\
    \textbf{$x_{h} \mapsto y$} (DIR)& 67.2 & 50.2 &	62.9  & 37.7 & 45.9 & 38.9& 35.2 & 35.2 & \bf26.9 & \bf15.7 & \bf18.3 & \bf5.1  \\
    \midrule

    BART~ & 69.3 & 45.0 & 62.6 & 34.5 & 38.8 & 35.0 & 38.2 &38.2 &27.2  &16.6  &18.1 &5.3 \\
    \textbf{$y^{'} \mapsto y$} (CSA)& 70.5 & 46.7 & \bf64.0 & 39.1 & 46.1 & \bf40.1 & \bf38.5 &\bf38.5 &\bf28.6  &\bf18.1 & 18.6 &5.3  \\
    \textbf{$x_{h} \mapsto y$} (CSA)& 66.3 & 45.7 &	60.8 & 36.7 & 43.4 & 37.7 & 38.2 & 38.2 &27.0 &16.9 & 17.8 & 5.2   \\
    \textbf{$x_{h} \mapsto y$} (DIR)& 65.1 & 47.4 & 60.6 & 37.6 & 45.1 & 37.5 & 39.2 & 39.2 & 27.8 & 17.7 & \bf19.2 & \bf5.9 \\
    \midrule
    
    LM-critic~ & 64.4 & 35.6 & 55.5 & 34.5 & 35.9 & 34.5 & 34.5 & 34.5& 23.5 &12.9 &15.8 &4.6  \\
    \textbf{$y^{'} \mapsto y$} (CSA)& 65.7 & 47.4 & \bf61.0 & 41.1 & 46.1 & \bf41.8 & 34.9 & \bf34.9 &24.4 &14.4 &16.4 &\bf5.0 \\
    \textbf{$x_{h} \mapsto y$} (CSA)& 64.8 & 36.4 &	56.1& 31.7& 35.3 &32.1 &34.5 &34.5 &26.0 &\bf15.8 & 16.1 & 4.2    \\
    \textbf{$x_{h} \mapsto y$} (DIR)& 58.5 & 35.8 & 51.9 & 31.0 & 35.1 & 31.5 & 34.3 & 34.3 & 24.4 & 14.4 & 16.1 & 5.0 \\
    \midrule
			
    BERT~ & 72.1 & 42.0 & \bf63.0 & 35.6 & 42.9 & 37.4 & \bf33.4 &\bf33.4 &23.2 &13.4 &15.9 &4.1 \\
    \textbf{$y^{'} \mapsto y$} (CSA)& 70.0 & 44.3 & 62.3 & 37.3 & 41.8 & \bf38.3 & 33.3 &33.3 &\bf23.3 &\bf13.7 &\bf16.7 &\bf4.5 \\
    \textbf{$x_{h} \mapsto y$} (CSA)&  68.7 &	42.8  &	61.3   & 35.2 &	40.4 & 	35.9   &  31.7 	&	31.7 &22.8 	&	13.3 &15.8 	&	4.2  \\
    \textbf{$x_{h} \mapsto y$} (DIR)& 68.0 & 43.6 & 61.2 & 35.6 & 41.2 & 36.3 & 32.6 & 32.6 & 23.4 & 13.5 & 16.2 & 4.4  \\
    \midrule
    
    RoBERTa~ & 68.7 & 47.2 & \bf62.9 & 38.2 & 44.1 & 39.0 & 36.9 &36.9 & 26.5 & 16.1 & 16.9 & 4.9 \\
    \textbf{$y^{'} \mapsto y$} (CSA)& 68.0 & 46.9 & 62.4 & 38.7 & 44.5 & \bf39.5 & \bf36.9 &\bf36.9 &25.3 &15.1 &16.6 &4.9 \\
    \textbf{$x_{h} \mapsto y$} (CSA)& 66.3 & 47.7 & 61.5 & 38.5 & 45.0 & 39.4 & 37.1 & 37.1 & \bf26.9 & \bf16.3 & 17.9 & 6.0  \\
    \textbf{$x_{h} \mapsto y$} (DIR)& 66.4 & 47.5 & 61.5 & 38.5 & 44.8 & 39.4 & 36.5 & 36.5 & 26.8 & 15.9 & \bf18.1 & \bf6.0 \\
    \midrule

    XLNet~ & 77.5 & 40.1 & \bf65.3 & 38.9 & 46.6 & 40.1 & 35.5 &35.5 &25.8 &15.4 &\bf17.7 &\bf5.1  \\
    \textbf{$y^{'} \mapsto y$} (CSA)& 75.3 & 41.6 & 64.8 & 39.0 & 46.6 & \bf40.1 & \bf36.3 &\bf36.3 & \bf26.3 & \bf15.7 &18.1 &5.0  \\
    \textbf{$x_{h} \mapsto y$} (CSA)& 66.7 & 48.6 &	62.1 & 38.1 & 45.0 & 39.1& 35.7& 35.7 & 25.4 & 15.4 & 17.7	&4.9  \\
    \textbf{$x_{h} \mapsto y$} (DIR)& 67.8 & 49.1 & 63.0 & 38.6 & 45.7 & 39.6 & 35.6 & 35.6 & 25.9 & 15.6 & 18.1 & 5.0  \\

    \bottomrule
\end{tabular}}
\caption{Comparison among different augmenting methods.
For each group, the first three settings were the same as in the previous experiment, while the last one which is denoted by \textbf{DIR} is conducting the \textbf{Stage \uppercase\expandafter{\romannumeral1}} of the CSA method directly without filtering the regularization samples.
The bold fonts are the optimal performance of each comparison. }
\label{tab:main_gec_com} 
\end{table*}

\subsection{The Effect of Self-Augmenting and Hard Samples}
\label{ana:c2b}
As mentioned before, there are different strategies during the self-augmenting process.
One is to follow the self-distillation method which combines each incorrect output $y^{'}$ generated by GEC model with the corresponding source data $x$ from the original training datasets to post-train the model.
The other strategy is to utilize  $D_{self}$ or $D_{hard}$ as new training set.
To illustrate the advancement of our CSA method, we compare these training strategies by reporting the results on the original testing data and the number of augmenting pairs per cycle in table \ref{tab: AB-training}. 
Afterward, to figure out the effect of the regularization data, we utilize augmenting pairs $D_{Aug}$ to retrain the model directly by applying the \textbf{Stage \uppercase\expandafter{\romannumeral1}} of the CSA method and report the results in Table \ref{tab:main_gec_com}. 

\subsubsection{Effect of Self-Augmenting}
We find that the performance of the baseline model can be improved in the first training cycle but will decrease after the second cycle under the self-distillation setting.
As for our introduced strategy in the self-augmenting process, the model performance rises continuously after two training cycles, with fewer training pairs.
This is because the GEC model is sensitive to the golden data, and the incorrect targets will make the model think that the syntactically incorrect parts are ``correct''.
This also confirms the above statement that the goal of self-distillation method is contrary to the GEC task.

\subsubsection{Effect of Regularization data}
We can conclude that the over-fitting problem of the GEC model hardly comes up when simply post-training the model with hard samples since the quantity of data is sufficient.
Specifically, for the Transformer model, we can observe that the hard samples can simultaneously improve the performance on the original testing sets and attack sets. Three out of five experiments (\textbf{CoNLL(CLN)}, \textbf{CoNLL(ATK)}, \textbf{ATK 3}) indicate the improvement brought by directly utilizing hard samples for training can even surpass the CSA method.
However, in terms of the overall results, the effects brought by regularization data are more prominent.
In short, on the one hand, we can infer that the success of our CSA method derives from the hard samples to some extent. On the other hand, the effects of regularization are beyond hard samples.

\subsection{Hyper-Parameter Analysis}
\label{ana:effect_CSA}
In the aforementioned experiments, we ignore analysing whether the improvement comes from the small high-quality data $\mathcal{D}_{tune}$ or the regularization data $\mathcal{D}_{Reg}$, and the influence of the hyper-parameter $\epsilon$ and $\mathcal{P}$ remains to be discovered. 
As a result, we train the model by simply utilizing the original training data or regularization data as well as compare the performance under the different settings of $\epsilon$ in Section \ref{sec:abla_eps}.
Meanwhile, we strictly control the training steps and the data selection strategy to study the influence of hyper-parameter $\mathcal{\bm P}$ in Section \ref{sec:abla_pat}.
In the end, we gradually reduce the proportion of the regularization data by controlling the reserving rate $\mathcal{\bm r}$ during the training step and compare the performance of the GEC model in the original testing set and attack test set simultaneously in Section \ref{sec:abla_res}.

\begin{table}[!tbp]
\centering
\small
\begin{tabular}{l | c c c c c c c }
    \toprule
    \textbf{\#Cycles} & \textbf{0} & \textbf{1} & \textbf{2} & \textbf{3} & \textbf{4} & \textbf{5} \\
    \midrule
    Ori Set & 61.8 & 56.2 & 57.0 & 56.5 & 57.1 & 55.2 \\
    \textbf{+ Ours} & 61.8 & 62.5 & 62.6 & 62.7 & 62.9 & 62.7 \\
    \midrule
    Attack Set & 36.7 &37.4  &37.0  &36.6  &37.0  &37.1  \\
    \textbf{+ Ours} & 36.7 &42.2  &41.3  &41.6  &41.8  &42.2 \\
    \bottomrule
\end{tabular}
\caption{Comparison between re-training the GEC model on $\mathcal{D}\cup \mathcal{D}_{tune}$ with the same epochs as our CSA, where Ori Set means the original testing set.}
\label{tab: epoch} 
\end{table}

\begin{table}[hbt!]

\centering
\small

\begin{tabular}{c | c c c | c c c}
    \toprule
    \multirow{2}{*}{$\mathcal{\bm P}$} & \multicolumn{3}{c|}{\textbf{CoNLL-2014~}} & \multicolumn{3}{c}{\textbf{CoNLL-2014 $(ATK)$~}} \\
    \cmidrule(r){2-7} 
    & \textbf{Prec.} & \textbf{Rec.} & \textbf{F\_{0.5}~~} & \textbf{Prec.} & \textbf{Rec.} & \textbf{F\_{0.5}~~} \\
    \midrule
    - & 67.9 & 44.1 & 61.3 & 34.1 & 39.7 & 34.9 \\
    2 & 67.2 & 49.4 & 62.6 & 40.6 & 44.3 & 41.3 \\
    3 & 68.1 & 48.5 & 63.2 & 40.3 & 43.6 & 40.9 \\
    4 & 68.2 & 48.9 & 63.2 & 40.6 & 43.7 & 41.2 \\
    5 & 68.6 & 48.6 & 63.4 & 40.0 & 43.4 & 40.7 \\ 
    6 & 66.2 & 48.9 & 61.8 & 40.2 & 43.2 & 40.8 \\
    \bottomrule
\end{tabular}
 \caption{The influence of $\mathcal{P}$ to model performance. ``\textbf{-}'' denotes baseline, and $(ATK)$ refers to the attack set.}
\label{tab: stage} 
\end{table}

\begin{table}[hbt!]
\centering
\small
\begin{tabular}{c | c c c | c c c }
    \toprule
    \multirow{2}{*}{\shortstack{\textbf{Reserving} \\ \textbf{Rates (\%)}}} & \multicolumn{3}{c|}{\textbf{CoNLL-2014 $(M^{2})$~}} & \multicolumn{3}{c}{\textbf{CoNLL-2014 $(ATK)$~}} \\
    \cmidrule(r){2-7} 
    & \textbf{P} & \textbf{R} & \textbf{F\_{0.5}~~} & \textbf{P} & \textbf{R} & \textbf{F\_{0.5}~~} \\
    \midrule
    0 \% & 68.6 & 48.6 & 63.4 & 40.0 & 43.4 & 40.7 \\
    25 \% & 68.5 & 48.6 & 63.3 & 40.1 & 43.3 & 40.7 \\
    50 \% & 68.3 & 48.8 & 63.2 & 40.3 & 43.7 & 40.9 \\
    75 \% & 68.3 & 48.5 & 63.1 & 40.5 & 43.8 & 41.1 \\
    100 \% & 67.5 & 49.4 & 62.9 & 40.9 & 44.5 & 41.6 \\
    \bottomrule
\end{tabular}
\caption{The influence of regularization data amount to model performance and defence capability.}
\label{tab:reg} 
\end{table}

\subsubsection{The Influence of Threshold $\epsilon$}
\label{sec:abla_eps}
Table~\ref{tab: epoch} presents the results of the model with different training cycles, correlating with the hyper-parameters of $\epsilon$. 
With the increase of training cycles, we can observe that our CSA method can surpass the baseline on the attack set in each cycle~(except for Cycle 5) and significantly improve the original test set, i.e., around 5 points improvement in Cycle 1.
To explore whether the improvement in cycle training comes from the introduced small dataset $\mathcal{D}_{tune}$, we train the baseline model on $\mathcal{D}\cup \mathcal{D}_{tune}$ with the same training epochs as our CSA method.
The dramatic decrease of the performance along with the cycle training proves that the improvement of robustness and performance on the original testing set is not simply brought by $\mathcal{D}_{tune}$ in the cycle training process.

\subsubsection{The Influence of Patience $\mathcal{P}$}
\label{sec:abla_pat}
Table~\ref{tab: stage} shows the performance of the model along with the different settings of patience $\mathcal{P}$.
With the increase of $\mathcal{P}$, the model can achieve better performance on the original testing set in five cycles, and the robustness is unstable but much better than the baseline.
It can be seen that $\mathcal{P}$=2 is sufficient to achieve competitive performance on the original testing set and best performance on the attack set, which is also used as the standard-setting in our implementations.

\subsubsection{The Influence of Reserving Rate}
\label{sec:abla_res}
We launch a preliminary experiment to show the relationship between the quantity of regularization data and the model's performance.
Table~\ref{tab:reg} presents the experimental results of removing different proportions of regularization data in the last training cycle.
It can be seen that more regularization data can improve the model robustness but suffer from the decreased performance on the original testing data.
This seemingly incompatible phenomenon validates previous work \citep{li2021searching} that there is a relative balance between the performance on the original training set and robustness of the model, and also reminds us of the possibility of utilizing fewer augmented data to achieve better results.

\subsection{The Influence of Regularization Data}
\label{ana:regular}
In this section, We'll take a thorough analysis of the impact of the two components $X_{unl}$ and $X_{unc}$ in the regularization data $D_{Reg}$.
We select some cases of regularization examples and make a case study in Section \ref{sec:case_study}.
To solve the problem of why regularization data can improve the performance on both attack set and original testing set mentioned in Section \ref{sec:effi_re}, we utilize $X_{unl}$ and $X_{unc}$ to fine-tune a plug-and-play GEC pre-trained model (transformer-big) to probe the latent capacity brought by the two data components in Section \ref{sec:com_2component}. 
Since there are some unexpected phenomena of the \textit{Seq2Edits} models, i.e., performing badly after post-training with regularization data, we will analyze such phenomena and propose a set of paradigms for different models about how to utilize regularization data efficiently in Section \ref{sec:ana_anomalies} and \ref{sec:para2util}, respectively.

\subsubsection{Case Study}
\label{sec:case_study}
\begin{table}[t]
\centering
\begin{tabular}{l l}
    \toprule
    \multicolumn{2}{l}{\textbf{Examples of Regularization Data}} \\
    \midrule

    \textit{Poor}: & I hope you 'll \textcolor{red}{attend} my suggestions . \\  
    \textit{Reg 1,2,3,4}:  & I hope you 'll \textcolor{olive}{follow} my suggestions . \\
    \textit{Good}:   & I hope you 'll \textcolor{blue}{act on} my suggestions . \\
    \midrule

    \textit{Poor}:   & So \textcolor{red}{It} was very \textcolor{red}{derty}. \\
    \textit{Reg 1,2,3,4}:  & So \textcolor{olive}{it} was very  \textcolor{olive}{dirty}. \\
    \textit{Good}:   &  So  \textcolor{blue}{It} was very  \textcolor{blue}{dirty}. \\ 
    \midrule
    
    \textit{Poor}: & I was very \textcolor{red}{frethend}, but I knew \textcolor{red}{,} that I should do something . \\
    \textit{Reg 1,2,3,4}:  & I was very \textcolor{olive}{free}, but I knew that I should do something . \\
    \textit{Good}: & I was very \textcolor{blue}{frightened}, but I knew that I had to do something . \\ 
    \midrule
    
    \textit{Poor}:   &  But I \textcolor{red}{do n't} stop \textcolor{red}{to walk} .\\
    \textit{Reg 1}:  &  But I \textcolor{olive}{did n't} stop \textcolor{olive}{walking} .\\
    \textit{Reg 2,3,4}:  &  But I \textcolor{olive}{do n't} stop \textcolor{olive}{walking} .\\
    \textit{Good}:   &  But I \textcolor{blue}{did n't} stop \textcolor{blue}{walking} .   \\ 
    \midrule
    
    \textit{Poor}:   &  Today, \textcolor{red}{It 's fist} time to entry \textcolor{red}{my message} .\\
    \textit{Reg 1}:  &  Today is \textcolor{olive}{my} first time to \textcolor{olive}{write} a message .\\
    \textit{Reg 2,3,4}:  &  Today, \textcolor{olive}{It 's my} first time to \textcolor{olive}{write} a message .\\ 
    \textit{Good}:   &  Today \textcolor{blue}{is the first} time for me to \textcolor{blue}{enter} a message .\\
    \midrule
    
    \textit{Poor}:   &  I thank you \textcolor{red}{about} my normal day .\\
    \textit{Reg 1}:  &  Thank you for my normal day .\\
    \textit{Reg 2,3}:  &  I thank you \textcolor{olive}{for} my normal day .\\ 
    \textit{Reg 4}:  &  I \textcolor{olive}{am thankful for} my normal day .\\
    \textit{Good}:   &  I thank you \textcolor{blue}{for} my normal day .   \\
    \midrule
    
    \textit{Poor}:   &  Hiroshima Carp lost the game, but it was very exciting game .\\
    \textit{Reg 1}:  &  Hiroshima Carp lost the game, but it was \textcolor{olive}{a} very exciting game .\\
    \textit{Reg 2,3}:  &  \textcolor{olive}{The} Hiroshima Carp lost the game, but it was \textcolor{olive}{a} very exciting game .\\ 
    \textit{Reg 4}:  &  Hiroshima Carp lost the game, but it was \textcolor{olive}{a} very exciting game .\\
    \textit{Good}:   &  Hiroshima Carp lose the game, but it was very exciting game .\\

    \bottomrule
\end{tabular}
\caption{Examples of regularization data, where $\textit{Poor}$, $\textit{Reg k}$ and $\textit{Good}$ represent for ungrammatical sentence, regularization data from $k$-th cycle and grammatical sentence respectively. We annotate the different positions of each sentence for three kinds of data by utilizing \textcolor{red}{red}, \textcolor{blue}{blue} and \textcolor{olive}{olive} color. We sample the sentences from four cycles of regularization data generated by \textit{Seq2Seq} model. To save space for the table, if there are multiple cycles of regularization with the same data, we will put them on the same row.}
\label{tab:case-study} 
\end{table}

As mentioned in Section \ref{sec:effi_re}, we decompose the regularization data into two components $X_{unl}$ and $X_{unc}$, and utilize those datasets to fine-tune the GEC model directly.
Similar to the process of building $D_{Reg}$, we take the intersection among different regularization data sets to extract $X_{unl}$, which can be formally written as $X_{unl}^{k} = \mathcal{D}_{Reg}^{1}\bigcap \mathcal{D}_{Reg}^{2} \dots \bigcap \mathcal{D}_{Reg}^{k}$ $(k \in \{2,3,4\})$.
As for $X_{unc}^{k}$ of the $k$-th cycle, we first combine all the regularization data of $k$ cycles and subtract the corresponding $X_{unl}^{k}$ set, which can be formally written as $X_{unc}^{k} = \mathcal{D}_{Reg}^{1}\bigcup \mathcal{D}_{Reg}^{2} \dots \bigcup \mathcal{D}_{Reg}^{k} - X_{unl}^{k}$ $(k \in \{2,3,4\})$.
Afterward, we select some representative cases from $X_{unl}$ and $X_{unc}$, and present them in Table \ref{tab:case-study}, where the first three cases are $X_{unl}$ cases, and the others are $X_{unc}$ cases.
We summarize five main types of regularization data as well as show some examples below:

\begin{itemize}

    \item[Type 1] The intermediate results are syntactic and semantic correct, while the editing positions are replaced by synonyms texts. 
    For example, in the first case, the regularization sentences have no syntactic error and these two phrases ``follow my suggestion'' and ``act on my suggestion'' have the same meaning. 
    
    \item[Type 2] The target sentence is mislabeled, i.e., in the second case which convert ``it'' to ``It''. 
    Such mislabeled samples impair the correction ability of the GEC model as they confuse the optimization goal, i.e., transferring $\textit{Poor} \rightarrow \textit{Good}$ to $\textit{Good} \rightarrow \textit{Poor}$. 
    
    \item[Type 3] The intermediate results are syntactic correct but semantic incorrect. 
    For example, in the third case, the goal is to correct the spelling of the word ``frightened'' but the regularization sentence change the meaning of original sentence from ``being frightened'' to ``being free''.  

    \item[Type 4] Because of the lack of prompt message, the intermediate results may have syntactic errors. 
    For example, in the forth case, the original sentence (\textit{Poor}) is in the simple present tense and the target sentence is in the simple past tense, but there is no obvious hint, which cause the changing of the generated sentence's tense. 
    
    
    \item[Type 5] For sentences which do not require modification, the intermediate results are excessively modified as shown in the last case.
\end{itemize}

From the above classification and explanation, we can summarize that the intermediate results are mostly syntactic correct but semantic incorrect, which impairs the performance.
From the perspective of data quality, the reasons behind this are the lack of enough context for \textit{Seq2Seq} models, and some mislabeled data interfere with optimization objectives.
From another aspect, we also think the shortcomings inherent in the Casual Language Model (CLM) count for this problem. 
Specifically, \textit{Seq2Seq} model suffers from the degeneration \citep{li2016diversity} and over-reasoning problems\footnote{The generating model tends to generate the generic, safe, universal text.}, and that is why it generates different variants of sentences even there is no need to modify the \textit{Poor} sentences.

\begin{figure}[!tbp]
  \centering
    \subfigure[Missing(R)]{\includegraphics[width=0.3\textwidth]{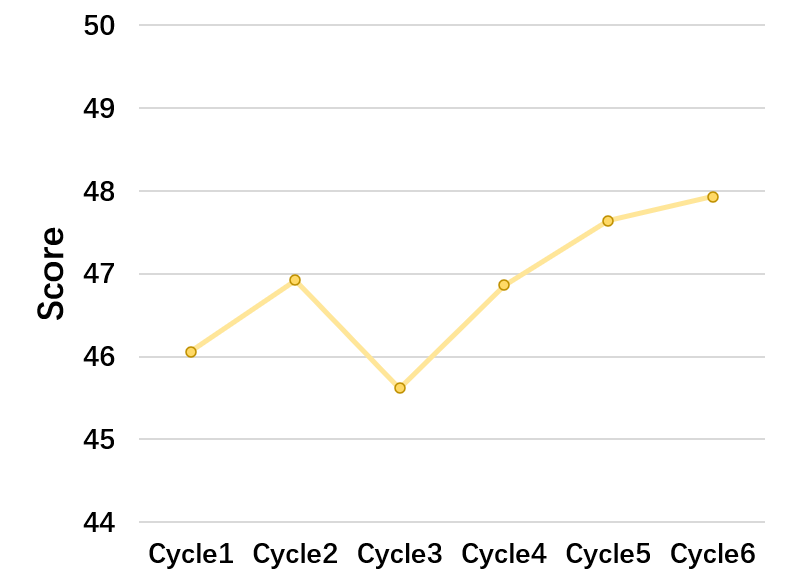}} 
    \hspace{0.01\linewidth}
	\subfigure[Replacement(R)]{\includegraphics[width=0.3\textwidth]{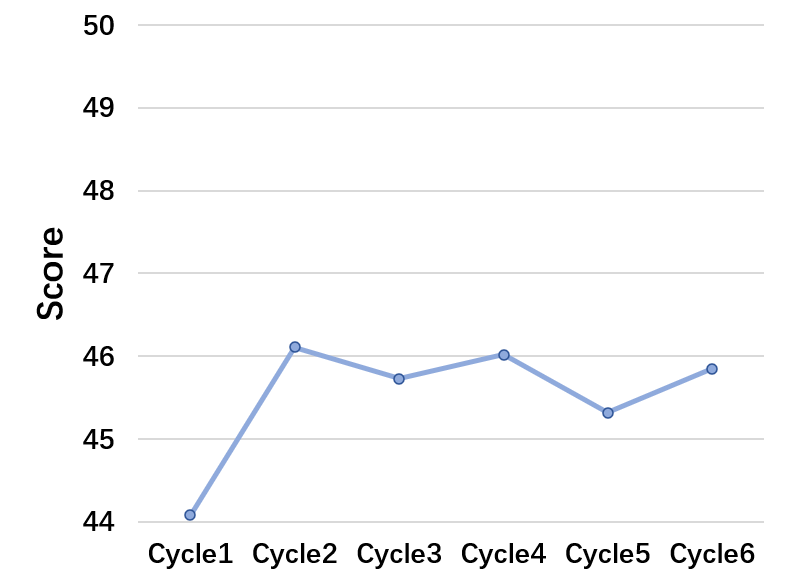}} 
	\hspace{0.01\linewidth}
	\subfigure[Unnecessary(R)]{\includegraphics[width=0.3\textwidth]{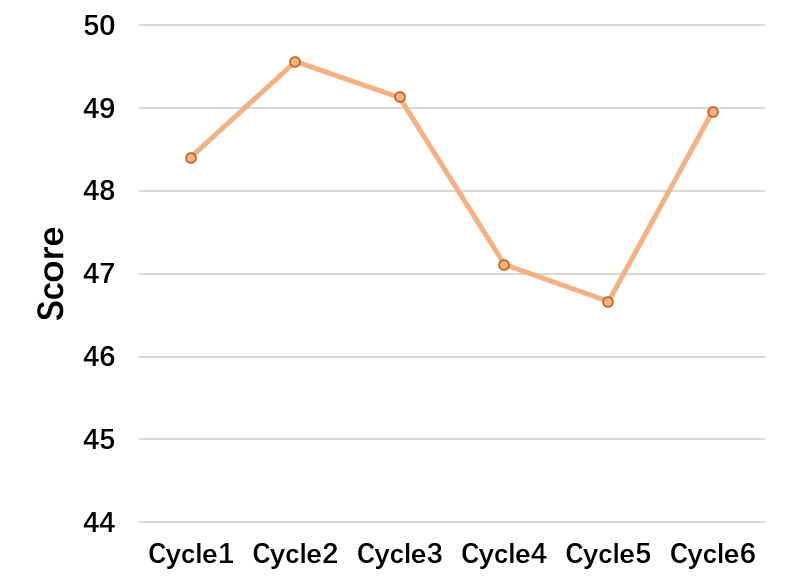}} 
	\vfill
    \subfigure[Missing(P)]{\includegraphics[width=0.3\textwidth]{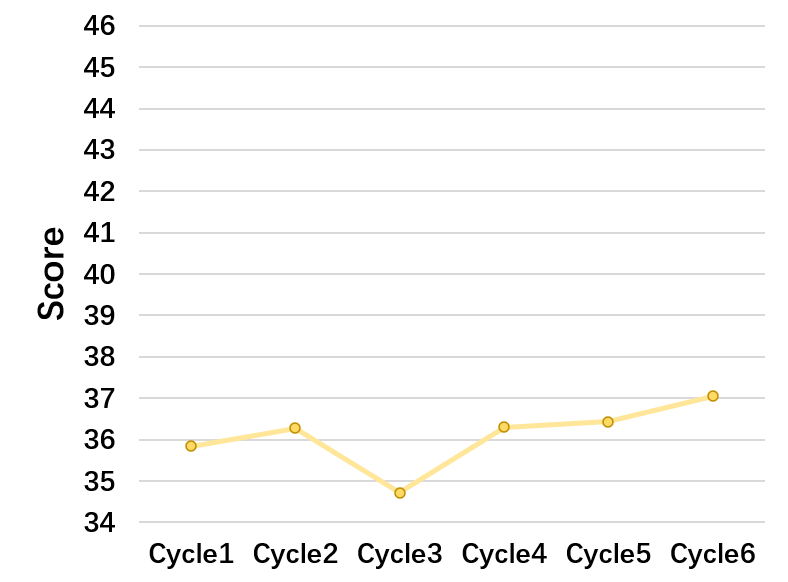}} 
    \hspace{0.01\linewidth}
	\subfigure[Replacement(P)]{\includegraphics[width=0.3\textwidth]{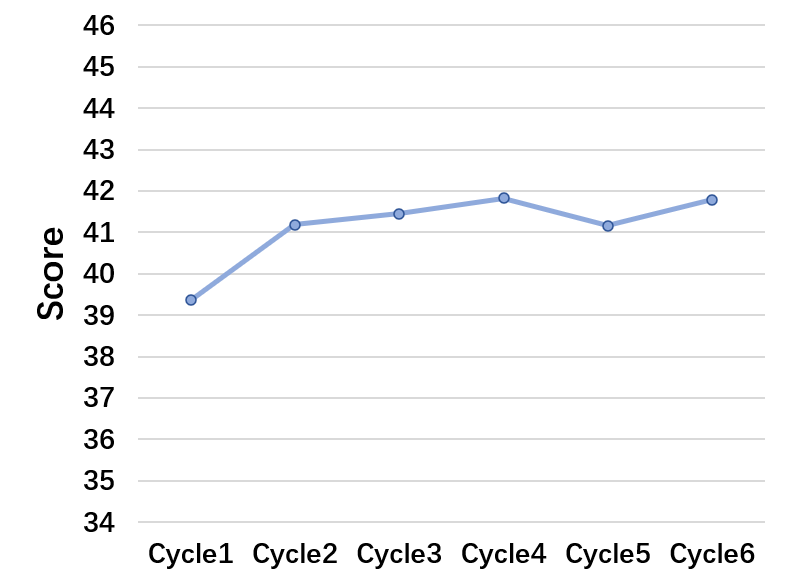}} 
	\hspace{0.01\linewidth}
	\subfigure[Unnecessary(P)]{\includegraphics[width=0.3\textwidth]{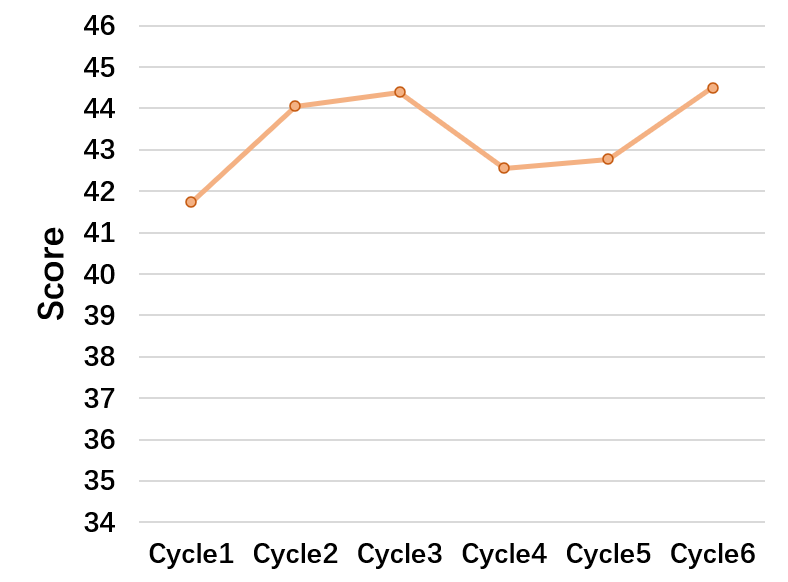}} 
	
    \caption{The trend of error type distribution along with the number of cycles. The above three figures describes the recall (R) of each error types while the bottom three figures describes the precision (P) of each error types. The error types are classified in terms of edit operation tier, i.e., whether tokens are missing (M), replaced (R) or unnecessary (U). }
	\label{fig:compara_seq}
\end{figure}

To verify this opinion, we collect the regularization data among different cycles and plot the recall and precision of editing operation in operation tier \citep{bryant2019automatic} in Figure \ref{fig:compara_seq}, where the above three sub-figures describe the recall and the bottom three describe the precision of editing operation, respectively.

As we can infer from the recall rate that the editing results (intermediate results) of \textit{Seq2Seq} model changes considerably with the increase of the number of cycles, especially for Unnecessary and Missing error types, which represents that the model tends to add or delete different tokens when generating the results.
Compared with recall for each error type, the precision tends to be stable, which means many editing operations are redundant.
In general, the editing ability is enhanced by the addition of regularization data, which counts for the improving performance of the GEC model.

\subsubsection{Comparison between $X_{unl}$ and $X_{unc}$}
\label{sec:com_2component}
\begin{table}[t]

\small
\centering
\begin{tabular}{l | c c| c c| c c}
\toprule[1pt]
\multirow{2}{*}{\textbf{Model}} & \multicolumn{2}{c|}{\textbf{ATK 1} ($\uparrow$)} & \multicolumn{2}{c|}{\textbf{ATK 2} ($\uparrow$)} & \multicolumn{2}{c}{\textbf{ATK 3} ($\uparrow$)} \\
\cmidrule(r){2-3} \cmidrule(r){4-5} \cmidrule(r){6-7} 
  &  \textbf{TR(\%)} & \textbf{SR(\%)} & \textbf{TR(\%)} & \textbf{SR(\%)}  & \textbf{TR(\%)} & \textbf{SR(\%)} \\
\midrule
Transformer (pre-train) &\bf35.4 &\bf35.4 &\bf27.5 &\bf16.1 &\bf19.2 &\bf6.1 \\
Transformer (benchmark) & 32.6 &32.6 &23.5 &13.3 &16.1 &4.1 \\
\midrule
+$X_{unl}$ (2 Cycles) & 34.4 & 34.4 &26.8 &15.7 &17.4 &4.4 \\
+$X_{unl}$ (3 Cycles) &34.0 &34.0 &\bf27.2 &\bf16.1 &17.2 &4.4 \\
+$X_{unl}$ (4 Cycles) &\bf33.4 &\bf33.4 &26.7 &15.3 &\bf17.2 &\bf4.6 \\
\midrule
+$X_{unl}$ (2 Cycles / reverse) &34.5 &34.5 &\bf27.7 &\bf17.2 &17.8 &4.9 \\
+$X_{unl}$ (3 Cycles / reverse) &34.4 &34.4 &27.4 &16.9  &17.1 &4.0 \\
+$X_{unl}$ (4 Cycles / reverse) &\bf34.6 &\bf34.6 &27.1 &16.5 &\bf18.1 &\bf5.1 \\
\midrule
+$X_{unc}$ (2 Cycles / seed 1) &34.9 &34.9 &27.0 &16.2 &18.2 &5.0 \\
+$X_{unc}$ (2 Cycles / seed 2) &35.1 &35.1 &27.4 &16.5 &18.9 &5.4 \\
+$X_{unc}$ (2 Cycles / seed 3) &34.6 &34.6 &27.5 &16.8 &18.0 &5.0 \\
+$X_{unc}$ (2 Cycles / seed 4) &35.1 &35.1 &27.5 &16.7 &18.4 &5.3 \\
+$X_{unc}$ (2 Cycles / seed 5) &34.0 &34.0 &27.2 &16.3 &18.2 &5.3 \\
+$X_{unc}$ (2 Cycles / Avg. ) &\bf34.7 &\bf34.7 &\bf27.3 &\bf16.5 &\bf18.3 &\bf5.2 \\
\midrule
+$X_{unc}$ (3 Cycles / seed 1) &35.0 &35.0 &27.3 &16.7 &18.4 &5.4 \\
+$X_{unc}$ (3 Cycles / seed 2) &35.1 &35.1 &26.8 &16.0 &18.7 &5.5 \\
+$X_{unc}$ (3 Cycles / seed 3) &34.7 &34.7 &27.5 &16.9 &18.0 &5.0 \\
+$X_{unc}$ (3 Cycles / seed 4) &34.7 &34.7 &27.2 &16.3 &18.2 &5.0 \\
+$X_{unc}$ (3 Cycles / seed 5) &34.7 &34.7 &27.8 &17.1 &18.8 &5.5 \\
+$X_{unc}$ (3 Cycles / Avg. ) &\bf34.8 &\bf34.8 &\bf27.3 &\bf16.6 &\bf18.4 &\bf5.3 \\
\midrule
+$X_{unc}$ (4 Cycles / seed 1) &35.0 &35.0 &27.6 &17.3 &18.7 &5.4 \\
+$X_{unc}$ (4 Cycles / seed 2) &35.1 &35.1 &26.8 &16.1 &18.2 &5.3 \\
+$X_{unc}$ (4 Cycles / seed 3) &34.0 &34.0 &27.3 &16.8 &17.7 &4.8 \\
+$X_{unc}$ (4 Cycles / seed 4) &33.9 &33.9 &27.3 &16.8 &18.4 &5.2 \\
+$X_{unc}$ (4 Cycles / seed 5) &34.4 &34.4 &26.4 &15.9 &18.2 &5.1 \\
+$X_{unc}$ (4 Cycles / Avg. ) &\bf34.5 &\bf34.5 &\bf27.1 &\bf16.6 &\bf18.2 &\bf5.2 \\
\bottomrule
\end{tabular}
\caption{Comparison of the recovery rate among the regularization data variants. \textbf{ATK $i$} ($i \in \{1,2,3\}$) represents for the evaluation sets with fixed number of attack positions per sentence, i.e., $i$ represents for the number of attack positions. We report the \textit{TR} and \textit{SR} score for each model, and annotate the cycle time $k$ of filtered regularization data. The bold fonts indicate the optimal performance of each comparison or the average score of five seeds.}
\label{tab:abla_recov} 
\end{table}

\begin{table}[hbt!]
\caption{Comparison of model performance among regularization data variants. \textbf{CLN} and \textbf{ATK} represents for original test data and standard attack set of CoNLL-2014 corpus respectively. We report the \textbf{F\_0.5} score and the corresponding \textbf{\#IPS} value. We regard the F\_0.5 of Transformer (pre-train) model as $score^{i}$ for calculating \textbf{\#IPS}. The bold fonts indicate the optimal performance of each comparison or the average score of five. It is worth noting that we keep two decimal places in the actual calculation, but we only keep one decimal place in this report.}
\label{tab:recov} 
    \begin{tabular}{l|c|cc|cc}
    \toprule[1pt]
    \multirow{2}{*}{\textbf{Model}} & \multirow{2}{*}{\textbf{\#Pairs}} & \multicolumn{2}{c|}{\textbf{CoNLL-2014 (CLN)}} & \multicolumn{2}{c}{\textbf{CoNLL-2014 (ATK)}} \\
     \cmidrule(r){3-4} \cmidrule(r){5-6} 
      & & \textbf{F\_0.5} ($\uparrow$) & \textbf{\#IPS} ($\downarrow$) & \textbf{F\_0.5} ($\uparrow$) & \textbf{\#IPS} ($\downarrow$) \\
    \midrule
    Transformer (pre-train) & 9,000,000 & 50.1 & - & 36.0 & - \\
    Transformer (benchmark) & 659,775 & 61.8 & 5,658.4  & 34.9 & -59,979.5\\
    \midrule
    +$X_{unl}$ (2 Cycles) &298,301 & 62.7 &2,375.0 &38.9 & 10,198.3 \\
    +$X_{unl}$ (3 Cycles) &271,499 & 62.6 &2,179.0 &38.8 & 9,696.4 \\
    +$X_{unl}$ (4 Cycles) &\bf259,780 & \bf62.8 &\bf2,052.0 &\bf38.8 &\bf9,195.8  \\
    \midrule
    +$X_{unl}$ (2 Cycles / reverse) &298,301 &62.5 &2,413.4 &38.6 &11,584.5 \\
    +$X_{unl}$ (3 Cycles / reverse) &271,499 &\bf62.9 &\bf2,127.7 &38.7 &10,245.3 \\
    +$X_{unl}$ (4 Cycles / reverse) &\bf259,780 &62.7 &2,068.3 &\bf38.7 &\bf9,711.4 \\
    \midrule
    +$X_{unc}$ (2 Cycles / seed 1) &298,301 &62.4 &2,433.1 &38.8 &10,847.3 \\
    +$X_{unc}$ (2 Cycles / seed 2) &298,301 &62.8 &2,356.2 &38.9 &10,286.2 \\
    +$X_{unc}$ (2 Cycles / seed 3) &298,301 &62.7 &2,375.0 &38.8 &10,653.6 \\
    +$X_{unc}$ (2 Cycles / seed 4) &298,301 &62.6 &2,394.1 &38.7 &10,946.8 \\
    +$X_{unc}$ (2 Cycles / seed 5) &298,301 &62.6 &2,394.1 &38.7 &11,151.4 \\
    +$X_{unc}$ (2 Cycles / Avg. ) &\bf298,301 &\bf62.6 &\bf2,390.5 &\bf38.8 &\bf10,777.1 \\
    \midrule
    +$X_{unc}$ (3 Cycles / seed 1) &271,499 &62.4 &2,214.5 &38.5 &10,752.4 \\
    +$X_{unc}$ (3 Cycles / seed 2) &271,499 &62.8 &2,144.5 &38.8 &9,783.7 \\
    +$X_{unc}$ (3 Cycles / seed 3) &271,499 &61.9 &2,308.7 &38.5 &10,860.0 \\
    +$X_{unc}$ (3 Cycles / seed 4) &271,499 &62.7 &2,161.6 &38.7 &10,245.3 \\
    +$X_{unc}$ (3 Cycles / seed 5) &271,499 &62.8 &2,144.5 &38.8 &9,872.7 \\
    +$X_{unc}$ (3 Cycles / Avg. ) &\bf271,499 &\bf62.5 &\bf2,144.8 &\bf38.6 &\bf10,302.8 \\
    \midrule
    +$X_{unc}$ (4 Cycles / seed 1) &259,780 &62.6 &2,084.9 &38.7 &9,533.2 \\
    +$X_{unc}$ (4 Cycles / seed 2) &259,780 &63.0 &2,020.1 &38.7 &9,621.5 \\
    +$X_{unc}$ (4 Cycles / seed 3) &259,780 &62.9 &2,035.9 &38.8 &9,631.4 \\
    +$X_{unc}$ (4 Cycles / seed 4) &259,780 &62.3 &2,136.3 &38.6 &9,991.5 \\
    +$X_{unc}$ (4 Cycles / seed 5) &259,780 &62.3 &2,136.3 &38.5 &10,603.3 \\
    +$X_{unc}$ (4 Cycles / Avg. ) &\bf259,780 &\bf62.6 &\bf2,082.7 &\bf38.7 &\bf9,822.2 \\
    \bottomrule
    \end{tabular}
\end{table}

To figure out the effect of two data components $X_{unl}$ and $X_{unc}$, we utilize those data sets respectively to fine-tune the plug-and-play GEC pre-trained model \citep{kiyono2019empirical}\footnote{\url{https://gec-pseudo-data.s3-ap-northeast-1.amazonaws.com/ldc_giga.pret.checkpoint_last.pt}} and compare the model's robustness and performance on the original testing set.
We divide the experiments into two groups: one is conducted on \textbf{ATK $i$} ($i \in \{1,2,3\}$) to compare the robustness, and the other one is conducted on original testing set and standard CoNLL-2014 attack set.
Considering that the $X_{unl}^{k}$ set always contains more data than the $X_{unc}^{k}$ set, we count the number of pairs in the $X_{unl}^{k}$ set and randomly sample the same number of sentences from the $X_{unc}^{k}$ set.
In order to reduce the randomness of sampling, we perform the above sampling operation five times with different random seeds and report the average result.

To calibrate the data quality, we also propose an evaluation metric \textbf{\#IPS} (\textbf{I}mprovement \textbf{P}er \textbf{S}core) which can be formally written as: $\text{\#IPS} = \frac{\#N_{sen}}{(score^{t} - score^{i}) / \gamma }$, where $\#N_{sen}$ represents for the number of sentences, $score^{t} - score^{i}$ represents for the difference value of scores, and $\gamma$ represents for the zoom factor.
We set $\gamma$ as 0.1 and utilize the \#IPS score to calibrate the quality of GEC data.

The results are shown in Table \ref{tab:abla_recov} and \ref{tab:recov}.
We report the performance of plug-and-play GEC model in the first group of each table.
We can see that the pre-trained GEC model has surprising strong robustness as it has the highest \textit{Recovery Rate} on \textbf{ATK 1} and \textbf{ATK 2} sets. 
Although the performance of fine-tuned GEC model on the original testing data is much higher than the pre-trained one, it has weak resistance to the attack data and even gets a negative value of \#IPS score.

\paragraph{The Influence of $X_{unl}$}
The results of $X_{unl}$ are shown in the second group and third group of each table.
With the increase of $k$ (Cycles), the quality of $X_{unl}$ becomes higher since the model achieve a relatively high \textit{Recovery Rate} and the \#IPS score is the lowest in the forth cycle compared with other datasets, i.e., we can utilize fewer data to achieve a comparable performance.
Moreover, as mentioned above, the type of errors in $X_{unl}$ is generally semantic but not syntactic, so we reverse the data pairs and use these data to fine-tune the pre-trained GEC model. The results are shown in the third group of the table. 
We surprisingly find that using reversed $X_{unl}$ can also improve the model robustness and performance on the original testing data and has the same trend as the original $X_{unl}$ with the increase of $k$.
One explanation for this phenomenon is that most of the source sentence (\textit{Poor}) in $X_{unl}$ set is grammatical in the syntactic level, and the reversed pairs can also make up for mislabeling problem as mentioned in Section \ref{sec:case_study}.

\paragraph{The Influence of $X_{unc}$}
The results of $X_{unc}$ are shown from the fourth to the fifth group of each table.
Each group shows the results of five seeds and one average at the bottom.
We can infer that the $X_{unc}$ is the main component for improving the model robustness as it achieves better \textit{Recovery Rate} scores than $X_{unl}$ in \textbf{ATK} sets.
Moreover, although the F\_0.5 score on standard CoNLL-2014 (ATK) set of $X_{unc}^{k}$ and corresponding $X_{unl}^{k}$ is similar, \textit{Recovery Rate} of $X_{unc}^{k}$ on \textbf{ATK $i$} set are higher than the F\_0.5 of $X_{unl}^{k}$, which means the GEC model trained with $X_{unc}^{k}$ is more sensitive to the robustness of local attacks.
In other words, to improve the robustness of the local attaks, more quantity of data is required as the \#IPS score of $X_{unc}^{k}$ on CoNLL-2014 (ATK) set in Table \ref{tab:recov} is much higher than the \#IPS of the corresponding $X_{unl}^{k}$.
As for the model performance on the original testing data, utilizing $X_{unc}$ does improve it and \#IPS score for each $X_{unc}^{k}$ is close to the \#IPS of the corresponding $X_{unl}^{k}$ as shown in Table \ref{tab:recov}, i.e., 2,82.7 for $X_{unc}^{k}$ and 2,052.0 for $X_{unl}^{k}$.

\subsubsection{Analysis of Anomalies on \textit{Seq2Edit} models}
\label{sec: generalization}
\begin{table}[t]
\small
\centering
\begin{tabular}{l l}
    \toprule
    \multicolumn{2}{l}{\textbf{Examples of Regularization Data}} \\
    \midrule
    
    \textit{Poor}: & I was very \textcolor{red}{frethend}, but I knew \textcolor{red}{,} that I should do something . \\
    \textit{BERT}:  & I was very \textcolor{orange}{frightened}, but I knew that I should do something .  \\
    \textit{Transformer}:  & I was very \textcolor{brown}{free}, but I knew that I should do something . \\
    \textit{Good}: & I was very \textcolor{blue}{frightened}, but I knew that I had to do something . \\ 
    \midrule 

    
    \textit{Poor}:   & It is a free  \textcolor{red}{soft wear} which is \textcolor{red}{serviced} by Google . \\
    \textit{BERT}:  & It is a free  \textcolor{orange}{software} which is \textcolor{red}{provided} by Google .  \\
    \textit{Transformer}:  & It is a free  \textcolor{brown}{software} which is \textcolor{red}{serviced} by Google . \\
    \textit{Good}:   & It is a free  \textcolor{blue}{software provided }by Google . \\ 
    \midrule

    \textit{Poor}:   &  I thank you \textcolor{red}{about} my normal day .\\
    \textit{BERT}:  & I thank you \textcolor{orange}{for} my normal day . \\
    \textit{Transformer}:  &  I \textcolor{brown}{am thankful for} my normal day .\\
    \textit{Good}:   &  I thank you \textcolor{blue}{for} my normal day .   \\
    \midrule
    
    
    
    \textit{Poor}:   &  I have \textcolor{red}{every day, holding} a pillow in bed .\\
    \textit{BERT}:  & I have \textcolor{orange}{every day, holding} a pillow in bed .  \\
    \textit{Transformer}:  &  I have a pillow \textcolor{brown}{every day} in bed .\\
    \textit{Good}:   &  I have \textcolor{blue}{the whole day to hold} a pillow in \textcolor{blue}{my} bed .\\

    \bottomrule
\end{tabular}
\caption{Comparison of regularization data between \textit{Seq2Edit} model and \textit{Seq2Seq} model, where $\textit{Poor}$ and $\textit{Good}$ represent for ungrammatical sentence and grammatical sentence respectively.
We use BERT and Transformer as the representatives of the \textit{Seq2Edit} and \textit{Seq2Seq} model architecture.
We annotate the different positions by utilizing \textcolor{red}{red} and \textcolor{blue}{blue} colors for \textit{Poor} and \textit{Good} sentences. 
As for regularization data belong to different model architectures, we utilize \textcolor{orange}{orange} and \textcolor{brown}{brown} colors to annotate the differences. }
\label{tab:case-study2} 
\end{table}

In order to figure out the slight improvement and unstable performance of the \textit{Seq2Edit} model in the previous experiments, we firstly compare the generated intermediate data between the \textit{Seq2Edit} model and \textit{Seq2Seq} model by utilizing Transformer and BERT as representatives.
We selected some representative cases from each intermediate data after two cycles of our CSA method and presented them in Table \ref{tab:case-study2}.
Combined with the six main types of regularization data generated by \textit{Seq2Seq} model (Section \ref{sec:case_study}), We can find some characteristics of the results generated by \textit{Seq2Edit} model:
\begin{itemize}
    \item The \textit{Seq2Edit} model tends to modify more for syntactic errors than for the semantics of the \textit{Poor} sentences. For example, in the first case, the \textit{Seq2Seq} model changes the original semantics while \textit{Seq2Edit} just corrects the wrong spelling of 'frightened'. 
    \item Compared with \textit{Seq2Seq} model, \textit{Seq2Edit} model seems to be less context-dependent when correcting syntax errors. For example, in the second case, both 'serviced' and 'provided' can be used in this sentence if there is no context, while \textit{Seq2Edit} can choose the latter one which matches the \textit{Good} sentences.
    \item The over-reasoning problem is mitigated in the \textit{Seq2Edit} model, and editing positions are fewer than \textit{Seq2Seq} model generally. In the third case, the outcome by \textit{Seq2Edit} model only edits one position while the \textit{Seq2Seq} generates another piece of content, so is the fourth case.
\end{itemize}

From the above case study, we can infer that \textit{Seq2Edit} tends to keep the structure and semantics of the original sentence.
Although this editing-based method is highly accurate, controllable, and interpretable, it to some extent goes against our CSA method since the regularization data generated by \textit{Seq2Edit} model is not as informative as the \textit{Seq2Seq} model.
Moreover, since the set of tags is generally not too large for \textit{Seq2Edit} model, it works better on closed sets, i.e., the official test datasets, but the complexities and long texts on open sets are challenging to deal with.
In other words, the limitation of tags, on the one hand, can improve the accuracy of editing operation on what the model has seen. On the other hand, it hurts the improvement in robustness since some co-occurrence tags never appeal in the dictionary, and the addition of regularization cannot change the distribution of tags.
Even worse, the released checkpoints of such model (\textit{Seq2Edit}) have already been meticulously trained on existing data, and any further post-training may hurt their performance.

To verify the above conjecture, we investigate the correction results for each error
type. 
Specifically, We use ERRANT\citep{felice2016automatic,bryant2017automatic} to measure the precision and recall of the model for each error type.
We compare the original plug-and-play GEC models with their enhanced versions, which are reinforced with our CSA method.
The results are shown in Table \ref{abla:ss_se}.
\begin{table}[hbt!]
\centering
\small
\resizebox{\textwidth}{!}{
\begin{tabular}{l | c c  c c | c c  c c }
    \toprule
    \multirow{2}{*}{\textbf{Error Type}} & \multicolumn{2}{c}{\textbf{Transformer}} & \multicolumn{2}{c|}{\textbf{+ CSA (2 cycles)}} & \multicolumn{2}{c}{\textbf{BERT}} & \multicolumn{2}{c}{\textbf{+ CSA (2 cycles)}} \\
    \cmidrule(r){2-3} \cmidrule(r){4-5} \cmidrule(r){6-7} \cmidrule(r){8-9}  
    & \textbf{Prec.} & \textbf{Rec.} & \textbf{Prec.} & \textbf{Rec.} & \textbf{Prec.} & \textbf{Rec.} & \textbf{Prec.} & \textbf{Rec.} \\
    \midrule
    \textbf{Missing} & 50.3 & 30.2 &\bf55.8 &\bf51.4 & \bf57.1 & \bf50.2 &55.4 &47.5 \\
    \textbf{Replacing} &53.9 &35.5 &\bf53.9 &\bf40.3 & 51.9 & 33.2 & \bf 54.2 & \bf 35.5 \\
    \textbf{Unnecessary} &44.7 &40.1 &\bf48.4 &\bf44.0 &\bf49.6 &\bf38.9 &45.4 &36.3 \\
    \midrule
    \textbf{PUNCT} &61.5 &25.5 &\bf61.7 &\bf56.5 & \bf60.3 & \bf55.6 &58.8 &51.6 \\
    \textbf{OTHER} &31.1 &15.6 &\bf30.9 &\bf17.5 &\bf29.7 &\bf12.0 &24.1 &11.5 \\
    \textbf{DET} &52.8 &46.6 &\bf55.8 &\bf53.1 &\bf57.2 &\bf53.1 &56.1 &48.7 \\
    \textbf{PREP} &52.8 &38.9 &\bf52.3 &\bf45.4 &\bf52.3 &\bf41.6 &51.7 &38.0 \\
    \textbf{VERB:TENSE} &49.8 &33.0 &\bf52.9 &\bf40.4 &\bf51.9 &\bf37.0 &50.7 &31.9 \\
    \bottomrule
\end{tabular}}
\caption{Comparison of the \textit{Seq2Seq} model and \textit{Seq2Edit} model on most
error types including all the top-5 frequent types of errors in W\&I-Dev evaluation set.
We report the precision and recall for each error type.
The first group represents the operation tier error types, and the second group shows the top-5 frequent types of errors.
The bold fonts indicate the optimal performance of each comparison.} 
\label{abla:ss_se}
\end{table}

As we can see from the Table, the original \textit{Seq2Edit} model does much better than the original \textit{Seq2Seq} model in the missing and unnecessary error types which are corresponding to insert and delete operations, and the editing-based operation also exceeds the generating-based operation on the four out of the top-5 frequent types of errors.
However, after the CSA method, We can also surprisingly find the considerable improvement of \textit{Seq2Seq} model on the top-5 frequent types of errors, and it exceeds \textit{Seq2Edit} model on all the error types.
We can also find that the performance of \textit{Seq2Edit} model gets worse with the addition of the regularization data.


\label{sec:ana_anomalies}

\subsubsection{Paradigms for Utilizing Regularization Data}
\label{sec:para2util}
From the above experiments and analysis, we find different effects of $X_{unc}$ set and $X_{unl}$ set on improving the performance, and we also compare the influence of regularization data between the \textit{Seq2Seq} model and the \textit{Seq2Edit} model.

In this section, we summarize the use of regularization data for different model architectures.
For \textit{Seq2Seq} model, the $X_{unc}$ set in regularization data contains much information which can make the model more robust and generalized since such model tends to generate diverse outcomes due to the decoding strategies, while the $X_{unl}$ set serves to compensate for the model's natural shortcomings because this data component force the model to reduces the diversity and reinforce the distributions which are not fully learned.

For \textit{Seq2Edit} model, lack of information in regularization data and the limitation of tags result in the side effects of regularization data on the performance and robustness.
However, there is no need to worry about this situation, as the original \textit{Seq2Edit} models work well on the closed sets of evaluation data due to the powerful encoder of PLMs.


\section{Conclusion}

Recent works have revealed that \textit{Seq2Seq} GEC models (even with data augmentation) are vulnerable to adversarial examples~\cite{wan2020improving}.
Nevertheless, those previous works mainly rested on simplex attack sets and tedious defense methods, i.e., training with adversarial samples, and the scope of the evaluation is narrow.
In this paper, we further explore the robustness of the GEC models by thoroughly evaluating various types of adversarial attacks and implementing an effective cycle self-augmenting method to improve model robustness.
Specifically, with our method, the model can be improved by 1.0 point (F\_0.5) on the original testing set and 3.9 points (F\_0.5) on the attack set by utilizing only about $39.4\%$ of the original data without requiring well-crafted adversarial examples at scale for a specific type of adversarial attack.
Experimental results on \textbf{seven} strong baselines, \textbf{four} benchmark test sets, \textbf{five} types of adversarial attacks, and \textbf{two} newly introduced evaluation metrics confirm the effectiveness of our proposed method, which can generalize well to various GEC models with only a few more training epochs as the extra cost.
Moreover, we also indicate that the reason behind the improvement is attributed to two data components in the regularization data, where one data component contributes to improving the correction ability while the other component focuses on improving the robustness.
In the future, we will explore how to improve the performance of the \textit{Seq2Edit} model on the open attack sets and some more efficient ways to use regularization data.

\section*{Acknowledge}
We would like to thank the efforts of anonymous reviewers for improving this paper.



\bibliography{main}
\bibliographystyle{iclr2024_conference}

\clearpage
\appendix
\section{Variants of Attack Results}
We post the results on four different attack sets in Table \ref{tab:attack_map}$\sim$\ref{tab:attack_ant} respectively.

\begin{table}[h]
\centering
\small
\resizebox{\textwidth}{!}{
    \begin{tabular}{l |c c c| c  c c| c c c| c}
    \toprule[1pt]
    \multirow{2}{*}{\textbf{Model}} & \multicolumn{3}{c|}{\textbf{BEA~} (ERRANT)} & \multicolumn{3}{c|}{\textbf{CoNLL~} ($M^{2}$)} & \multicolumn{3}{c|}{\textbf{FCE~} ($M^{2}$)} & \textbf{JFLEG~~}\\
    \cmidrule(r){2-4} \cmidrule(r){5-7} \cmidrule(r){8-10} \cmidrule(r){11-11}
      & \textbf{Prec.} & \textbf{Rec.} & \textbf{F\_{0.5}~~} & \textbf{Prec.} & \textbf{Rec.} & \textbf{F\_{0.5}~~} & \textbf{Prec.} & \textbf{Rec.} & \textbf{F\_{0.5}~~} & \textbf{GLEU~~} \\
    \midrule
    
    Transformer~ &8.4 &37.3 &9.9 &36.5 &37.8 &36.7 &19.7 &27.1 &20.8 &37.7  \\
    \textbf{$y^{'} \mapsto y$} (4 cycles) &11.2 &44.4 &13.2 &41.2 &45.8 &\bf42.0 &22.8 &32.5 &\bf24.3 &\bf38.5 \\
    \textbf{$x_{h} \mapsto y$} (3 Cycles)&11.0  &44.6  &13.0 &41.8  &45.4  &42.5  &23.4  &33.3  &24.9  &39.3 \\
    \textbf {$\nabla_{ATK}$} (2 Cycles) &9.9 &42.8 &11.9 &37.6 &38.1 &37.8 &21.0 &27.4 &21.2 &37.6  \\
    \midrule
    
    Bert-fuse~ &7.9 &36.0 &9.3 &36.7 &37.5 &36.9 &19.9 &27.5 &21.1 &37.4 \\
    \textbf{$y^{'} \mapsto y$} (3 cycles) &11.5 &45.4 &\bf13.3 &41.5 &45.1 &\bf42.2 &23.9 &34.3 &\bf25.4 &\bf38.7 \\
    \textbf{$x_{h} \mapsto y$} (2 Cycles)&11.2  &45.2  &13.2 &40.4  &44.6  &41.2  &24.3  &35.1  &25.9  &39.1 \\
    \textbf {$\nabla_{ATK}$} (1 Cycles) &12.3 &45.9 &14.3 &42.1 &45.9 &42.9 &23.5 &34.0 &25.0 &40.1  \\
    \midrule
    
    BART~ &8.1 &34.8 &9.6 &36.7 &37.4 &36.8 &18.7 &24.9 &19.7 &35.8 \\
    \textbf{$y^{'} \mapsto y$} (2 cycles) &12.2 &46.6 &\bf14.3 &42.3 &45.5 &42.9 &22.9 &32.8 &\bf24.4 &\bf37.7 \\
    \textbf{$x_{h} \mapsto y$} (1 Cycle)&11.1 &44.3 &13.1 &39.9 &42.9 &40.8 &21.8 &30.9 &23.1 &37.3 \\
    \textbf {$\nabla_{ATK}$} (2 Cycles) &10.6 &41.7 &12.4 &37.9 &42.7 &39.0 &20.2 &28.1 &21.2 &36.6  \\
    \midrule
    
    LM-Critic~ &6.4 &26.7 &7.6 &33.1 &30.8 &32.6 &14.0 &17.1 &14.5 &34.4 \\
    \textbf{$y^{'} \mapsto y$} (2 cycles) &11.4 &42.9 &13.4 &40.8 &43.8 &41.4 &21.9 &29.8 &\bf23.1 &\bf38.6 \\
    \textbf{$x_{h} \mapsto y$} (1 Cycle)&7.8  &33.6  &9.2 &33.5  &31.9  &33.1  &17.2  &22.5  &18.1  &35.2 \\
    \textbf {$\nabla_{ATK}$} (1 Cycles) &9.9 &38.7 &11.6 &38.1 &36.2 &37.6 &20.2 &29.5 &21.5 &36.3  \\
    \midrule
    
    BERT~ &10.8 &40.9 &12.7 &35.6 &41.6 &39.9 &23.9 &34.0 &25.4 &39.0 \\
    \textbf{$y^{'} \mapsto y$} (1 cycle)&11.4 &42.5 &\bf13.4 &40.5 &43.2 &\bf40.9 &24.5 &34.7 &26.0 &\bf39.1 \\
    \textbf{$x_{h} \mapsto y$} (1 Cycle)&9.7  &38.0  &11.4 &37.7  &38.4  &37.8  &22.8  &31.6  &24.2  &38.7 \\
    \textbf {$\nabla_{ATK}$} (1 Cycles) &11.2 &41.5 &13.2 &39.8 &42.7 &40.4 &21.7 &29.8 &22.8 &38.7  \\
    \midrule
    
    RoBERTa~ &11.8 &43.8 &13.8 &41.3 &42.6 &41.6 &24.5 &34.2 &26.0 &40.0 \\
    \textbf{$y^{'} \mapsto y$} (1 cycle) &12.2 &44.6 &\bf14.3 &42.2 &43.2 &\bf42.2 &24.6 &34.6 &\bf26.1 &\bf40.3 \\
    \textbf{$x_{h} \mapsto y$} (1 Cycle)&12.1  &44.5  &14.2 &41.9  &43.7  &42.3  &24.8  &34.4  &26.3  &40.1 \\
    \textbf {$\nabla_{ATK}$} (3 Cycles) &12.0 &43.9 &14.1 &40.1 &40.8 &42.0 &22.9 &31.7 &24.2 &39.5  \\
    \midrule
    
    XLNet~ &13.2 &46.8 &15.4 &42.1 &45.3 &42.7 &27.4 &39.3 &29.1 &40.7 \\
    \textbf{$y^{'} \mapsto y$} (2 cycles) &13.3 &47.4 &15.5 &42.1 &45.9 &42.8 &27.4 &39.4 &\bf29.2 &\bf40.9 \\
    \textbf{$x_{h} \mapsto y$} (1 Cycle)&12.5  &45.1  &14.6 &41.4  &43.9  &41.9  &27.1  &38.3  &28.8  &40.6 \\
    \textbf {$\nabla_{ATK}$} (1 Cycles) &12.7 &45.8 &14.8 &42.0 &43.8 &42.0 &24.5 &24.6 &24.5 &39.8  \\

    \bottomrule
    \end{tabular}
}
\caption{Evaluation results on \textit{Mapping \& Rules} attack sets. 
We generate the attack sets by using \textit{Mapping \& Rules} method as mentioned in Section \ref{sec:sub_op}.
We report the performance of two variants of regularization data for each model and the corresponding best cycle times, where $\mapsto$ represents the direction of data flow.
The bold fonts indicate the optimal performance of each comparison.}
\label{tab:attack_map}
\end{table}

\begin{table}[hbt!]
\centering
\small
\resizebox{\textwidth}{!}{
    \begin{tabular}{l |c c c| c  c c| c c c| c}
    \toprule[1pt]
    \multirow{2}{*}{\textbf{Model}} & \multicolumn{3}{c|}{\textbf{BEA~} (ERRANT)} & \multicolumn{3}{c|}{\textbf{CoNLL~} ($M^{2}$)} & \multicolumn{3}{c|}{\textbf{FCE~} ($M^{2}$)} & \textbf{JELEG~~}\\
    \cmidrule(r){2-4} \cmidrule(r){5-7} \cmidrule(r){8-10} \cmidrule(r){11-11}
      & \textbf{Prec.} & \textbf{Rec.} & \textbf{F\_{0.5}~~} & \textbf{Prec.} & \textbf{Rec.} & \textbf{F\_{0.5}~~} & \textbf{Prec.} & \textbf{Rec.} & \textbf{F\_{0.5}~~} & \textbf{GLEU~~} \\
    \midrule
    
    Transformer~ &15.9 &44.3 &18.3 &24.1 &33.1 &25.5 &26.3 &31.0 &27.1 &43.8  \\
    \textbf{$y^{'} \mapsto y$} (4 cycles) &19.4 &50.8 &22.1 &29.6 &40.7 &31.3 &28.7 &35.4 &29.8 &\bf45.0 \\
    \textbf{$x_{h} \mapsto y$} (3 Cycles)&19.2  &50.8  &21.9 &29.8  &40.9  &31.5  &29.5  &36.1  &30.6  &46.0 \\
    \textbf {$\nabla_{ATK}$} (2 Cycles) &17.5 &47.4 &20.1 &28.9 &33.3 &29.8 &25.3 &30.2 &26.2 &43.2  \\
    \midrule
    
    Bert-fuse~ &14.8 &41.4 &17.0 &22.7 &31.0 &24.0 &26.2 &29.8 &26.9 &44.0 \\
    \textbf{$y^{'} \mapsto y$} (3 cycles) &19.4 &51.2 &\bf22.2 &29.8 &40.1 &31.4 &30.0 &36.8 &31.2 &\bf45.8 \\
    \textbf{$x_{h} \mapsto y$} (2 Cycles)&19.1  &50.9  &21.9 &29.7  &41.1  &31.5  &30.9  &38.1  &32.1  &46.3 \\
    \textbf {$\nabla_{ATK}$} (1 Cycles) &20.1 &51.8 &22.8 &30.5 &42.0 &32.4 &29.6 &37.1 &30.7 &45.6  \\
    \midrule
    
    BART~ &15.1 &39.4 &17.3 &23.1 &31.2 &24.4 &25.1 &27.0 &25.5 &41.8 \\
    \textbf{$y^{'} \mapsto y$} (2 cycles) &20.5 &50.5 &\bf23.3 &30.9 &41.8 &\bf32.6 &28.8 &34.5 &\bf29.8 &\bf44.5 \\
    \textbf{$x_{h} \mapsto y$} (1 Cycle)&17.6 &45.5 &20.0 &26.6 &36.8 &28.2 &27.8 &32.0 &28.5 &43.6 \\
    \textbf {$\nabla_{ATK}$} (2 Cycles) &18.4 &45.5 &20.8 &27.9 &37.2 &29.3 &26.7 &27.2 &26.9 &42.8  \\
    \midrule
    
    LM-Critic~ &11.7 &31.4 &13.4 &33.0 &31.3 &32.7 &17.8 &18.4 &17.9 &37.7 \\
    \textbf{$y^{'} \mapsto y$} (2 cycles) &20.0 &49.0 &22.7 &40.7 &43.9 &\bf41.3 &28.1 &32.7 &\bf29.0 &\bf44.3 \\
    \textbf{$x_{h} \mapsto y$} (1 Cycle)&14.0  &36.9  &16.0 &21.4  &28.5  &22.5  &22.8  &24.2  &23.0  &39.8 \\
    \textbf {$\nabla_{ATK}$} (1 Cycles)&19.2 &20.1 &19.4 &31.1 &26.5 &30.0 &26.1 &25.1 &26.0 &41.8  \\
    \midrule
    
    BERT~ &18.2 &46.7 &20.8 &27.3 &37.3 &28.8 &29.9 &36.6 &31.1 &43.6 \\
    \textbf{$y^{'} \mapsto y$} (1 cycle) &18.6 &47.8 &\bf21.2 &28.9 &29.4 &29.0 &30.5 &37.6 &\bf31.7 &\bf43.7 \\
    \textbf{$x_{h} \mapsto y$} (1 Cycle)&16.6  &43.5  &18.9 &25.2  &34.0  &26.6  &28.5  &33.8  &29.4  &43.3 \\
    \textbf {$\nabla_{ATK}$} (1 Cycles) &18.7 &47.9 &21.3 &27.1 &37.2 &28.7 &27.1 &28.0 &27.5 &43.3  \\
    \midrule
    
    RoBERTa~ &19.8 &49.3 &22.5 &28.9 &38.8 &30.5 &30.7 &37.3 &31.8 &44.6 \\
    \textbf{$y^{'} \mapsto y$} (1 cycle) &19.9 &49.4 &\bf22.6 &29.9 &39.4 &31.4 &31.1 &38.2 &32.3 &\bf44.8 \\
    \textbf{$x_{h} \mapsto y$} (1 Cycle)&19.6  &49.0  &22.3 &29.5  &39.5  &31.0  &30.9  &38.0  &32.1  &44.6 \\
    \textbf {$\nabla_{ATK}$}  (3 Cycles) &19.7 &49.2 &22.4 &29.6 &39.6 &31.1 &28.7 &29.0 &28.8 &44.3  \\
    \midrule
    
    XLNet~ &20.8 &51.1 &23.6 &30.7 &42.0 &32.5 &34.0 &43.2 &35.5 &45.7 \\
    \textbf{$y^{'} \mapsto y$} (2 cycles) &20.8 &50.9 &\bf23.6 &30.8 &41.8 &32.5 &34.1 &43.2 &\bf35.6 &\bf45.8 \\
    \textbf{$x_{h} \mapsto y$} (1 Cycle)&19.3  &48.6  &21.9 &29.3  &40.0  &30.9  &33.5  &41.9  &34.9  &45.5 \\
    \textbf {$\nabla_{ATK}$} (1 Cycles) &20.4 &49.1 &23.3 &30.4 &40.8 &31.8 &29.3 &30.9 &29.4 &44.4  \\
    
    \bottomrule

    \end{tabular}
}
\caption{Evaluation results on \textit{Synonyms Substitutions} attack sets. 
We generate the attack sets by using \textit{Synonyms Substitutions} method as as mentioned in Section \ref{sec:sub_op}.
We report the performance of two variants of regularization data for each model and the corresponding best cycle times, where $\mapsto$ represents the direction of data flow.
The bold fonts indicate the optimal performance of each comparison.}
\label{tab:attack_syn} 
\end{table}

\begin{table}[hbt!]
\centering
\small
\resizebox{\textwidth}{!}{
    \begin{tabular}{l |c c c| c  c c| c c c| c}
    \toprule[1pt]
    \multirow{2}{*}{\textbf{Model}} & \multicolumn{3}{c|}{\textbf{BEA~} (ERRANT)} & \multicolumn{3}{c|}{\textbf{CoNLL~} ($M^{2}$)} & \multicolumn{3}{c|}{\textbf{FCE~} ($M^{2}$)} & \textbf{JFLEG~~}\\
    \cmidrule(r){2-4} \cmidrule(r){5-7} \cmidrule(r){8-10} \cmidrule(r){11-11}
      & \textbf{Prec.} & \textbf{Rec.} & \textbf{F\_{0.5}~~} & \textbf{Prec.} & \textbf{Rec.} & \textbf{F\_{0.5}~~} & \textbf{Prec.} & \textbf{Rec.} & \textbf{F\_{0.5}~~} & \textbf{GLEU~~} \\
    \midrule
    
    Transformer~ &24.3 &57.1 &27.4 &33.0 &48.9 &35.3 &31.3 &42.7 &33.1 &46.3 \\
    \textbf{$y^{'} \mapsto y$} (4 cycles) &25.2 &58.0 &\bf28.4 &33.5 &49.8 &35.8 &31.9 &42.9 &\bf33.6 &\bf46.7 \\
    \textbf{$x_{h} \mapsto y$} (3 Cycles)&25.2  &58.7  &28.4 &33.9  &50.7  &36.3  &32.3  &44.0  &34.1  &47.1 \\
    \textbf {$\nabla_{ATK}$} (2 Cycles) &24.5 &57.3 &27.8 &32.96 &48.9 &35.2 &30.9 &42.1 &32.5 &46.1  \\
    \midrule
    
    Bert-fuse~ &24.2 &56.6 &27.3 &32.6 &48.2 &34.9 &31.8 &43.1 &33.6 &46.6  \\
    \textbf{$y^{'} \mapsto y$} (3 cycles) &25.3 &58.6 &\bf28.6 &33.7 &50.3  &\bf36.1 &32.5 &44.5  &\bf34.4 &\bf47.1 \\
    \textbf{$x_{h} \mapsto y$} (2 Cycles)&25.1  &58.7  &28.3  &33.8  &50.8  &36.2  &32.6  &44.8  &34.5  &46.9\\
    \textbf {$\nabla_{ATK}$} (1 Cycles) &25.0 &58.9 &28.3 &34.0 &50.8 &36.3 &32.2 &44.0 &34.0 &47.2  \\
    \midrule
    
    BART~&24.2 &57.2 &27.4 &33.2 &48.9 &35.5 &31.2 &42.0 &32.9 &46.3  \\
    \textbf{$y^{'} \mapsto y$} (2 cycles) &25.3 &59.5 &28.6 &34.4 &51.1 &\bf36.6 &31.9 &43.7 &\bf33.7 &\bf46.7 \\
    \textbf{$x_{h} \mapsto y$} (1 Cycle)&24.9 &59.2 &28.2 &34.1 &51.0 &36.5 &32.3 &44.5 &34.2 &46.6 \\
    \textbf {$\nabla_{ATK}$} (2 Cycles) &24.7 &58.5 &27.9 &33.5 &49.7 &35.7 &31.5 &42.7 &33.1 &46.4  \\
    \midrule
    
    LM-Critic~ &23.9 &56.7 &27.0 &32.9 &49.1 &35.3 &30.2 &41.7 &32.0 &46.0 \\
    \textbf{$y^{'} \mapsto y$} (2 cycles)  &24.7 &59.0 &27.9 &34.3 &50.5 &\bf36.5 &32.3 &44.3 &34.2 &\bf47.0 \\
    \textbf{$x_{h} \mapsto y$} (1 Cycle)&23.9  &56.3  &27.0 &32.4  &47.8  &34.6  &31.2  &41.9  &32.9  &46.2 \\
    \textbf {$\nabla_{ATK}$} (1 Cycles) &24.3 &57.7 &27.5 &33.4 &50.1 &35.8 &31.4 &42.5 &33.1 &46.3  \\
    \midrule
    
    BERT~ &24.7 &58.8 &28.0 &33.5 &50.2 &35.9 &32.5 &45.0 &34.5 & 46.7 \\
    \textbf{$y^{'} \mapsto y$} (1 cycle) &24.9 &59.1 &\bf28.3 &33.8 &50.6 &\bf36.2 &32.8 &45.5 &\bf34.7 &\bf46.9 \\
    \textbf{$x_{h} \mapsto y$} (1 Cycle)&24.5  &57.8  &27.6 &33.2  &49.5  &35.6  &32.2  &44.4  &34.1  &46.7 \\
    \textbf {$\nabla_{ATK}$} (1 Cycles) &25.1 &59.7 &28.4 &34.0 &50.1 &36.4 &31.7 &44.3 &33.7 &46.8  \\
    \midrule
    
    RoBERTa~ &25.4 &59.3 &28.7 &34.4 &51.1 &36.8 &32.7 &45.1 &34.6 &46.5 \\
    \textbf{$y^{'} \mapsto y$} (1 cycle) &25.6 &59.5 &\bf28.9 &34.5 &51.4 &\bf37.0 &33.1 &45.1 &\bf34.9 &\bf46.8 \\
    \textbf{$x_{h} \mapsto y$} (1 Cycle)&25.4  &59.3  &28.6 &34.5  &51.5  &36.9  &32.6  &45.2  &34.6  &46.5 \\
    \textbf {$\nabla_{ATK}$} (3 Cycles) &25.5 &59.4 &28.8 &34.8 &51.7 &37.2 &32.2 &45.0 &34.1 &46.8  \\
    \midrule
    
    XLNet~ &25.8 &60.0 &29.1 &34.4 &51.8 &36.9 &33.8 &47.5 &35.8 &47.5 \\
    \textbf{$y^{'} \mapsto y$} (2 cycles) &25.8 &61.0 &\bf29.2 &34.7 &51.9 &\bf37.1 &34.2 &47.4 &\bf36.3 &\bf47.7 \\
    \textbf{$x_{h} \mapsto y$} (1 Cycle)&25.4  &59.5  &28.7 &34.2  &51.3  &36.6  &33.8  &47.4  &35.9  &47.5 \\
    \textbf {$\nabla_{ATK}$} (1 Cycles) &32.2 &45.1 &29.0 &34.8 &51.8 &37.1 &32.5 &44.4 &34.3 &47.1  \\
    
    \bottomrule

    \end{tabular}
}
\caption{Evaluation results on \textit{Back-Translation} attack sets. We generate the attack sets by using \textit{Back-Translation} method as as mentioned in Section \ref{sec:con}.
We report the performance of two variants of regularization data for each model and the corresponding best cycle times, where $\mapsto$ represents the direction of data flow.
The bold fonts indicate the optimal performance of each comparison.}
\label{tab:attack_back} 
\end{table}

\begin{table}[hbt!]
\centering
\small
\resizebox{\textwidth}{!}{
    \begin{tabular}{l |c c c| c  c c| c c c| c}
    \toprule[1pt]
    \multirow{2}{*}{\textbf{Model}} & \multicolumn{3}{c|}{\textbf{BEA~} (ERRANT)} & \multicolumn{3}{c|}{\textbf{CoNLL~} ($M^{2}$)} & \multicolumn{3}{c|}{\textbf{FCE~} ($M^{2}$)} & \textbf{JFLEG~~}\\
    \cmidrule(r){2-4} \cmidrule(r){5-7} \cmidrule(r){8-10} \cmidrule(r){11-11}
      & \textbf{Prec.} & \textbf{Rec.} & \textbf{F\_{0.5}~~} & \textbf{Prec.} & \textbf{Rec.} & \textbf{F\_{0.5}~~} & \textbf{Prec.} & \textbf{Rec.} & \textbf{F\_{0.5}~~} & \textbf{GLEU~~} \\
    \midrule
    Transformer~ &35.3 &53.4 &37.8 &42.9 &39.0 &42.0 &41.5 &36.0 &40.3 &53.7  \\
    \textbf{$y^{'} \mapsto y$} (4 Cycles)&39.8 &59.8 &42.7 &46.1 &45.6 &\bf46.0 &46.4 &44.2 &45.9 &\bf55.5 \\
    \textbf{$x_{h} \mapsto y$} (3 Cycles)&39.8  &60.2  &42.7 &47.3  &46.4  &47.1  &44.6  &41.6  &45.5  &56.4 \\
    \textbf {$\nabla_{ATK}$} (1 Cycles) &39.6 &59.7  &42.5  &40.3 &38.4  &39.9  &43.2 &41.2 &42.6  &52.4  \\
    \midrule
    
    Bert-fuse~ &34.6 &50.3 &36.9 &41.8 &36.2 &40.5 &46.1 &37.5 &44.1 &53.5 \\
    \textbf{$y^{'} \mapsto y$} (3 Cycles)&39.4 &59.5 &\bf42.3 &46.5 &45.9 &46.4 &48.3 &44.1 &47.4 &\bf56.3 \\
    \textbf{$x_{h} \mapsto y$} (2 Cycles)&39.9  &59.7  &42.7 &45.8  &45.3  &45.7  &47.8  &44.4  &47.1  &56.3 \\
    \textbf {$\nabla_{ATK}$} (1 Cycles) &40.6 &62.7  &43.7  &47.7 &45.6  &47.1  &46.9 &42.7  &45.9  &56.4  \\
    \midrule
    
    BART~ &36.0 &47.3 &37.8 &44.9 &37.6 &43.2 &45.2 &32.0 &41.7 &51.2 \\
    \textbf{$y^{'} \mapsto y$} (1 Cycle)&42.0 &59.0 &44.6 &48.8 &45.7 &48.1 &44.9 &38.8 &43.5 &\bf54.2 \\
    \textbf{$x_{h} \mapsto y$} (1 Cycle)&39.8  &56.6  &42.3 &46.0  &42.8  &45.3  &45.1  &37.9  &43.5  &53.0 \\
    \textbf {$\nabla_{ATK}$} (2 Cycles) &41.3 &57.3  &43.8  &46.8 &43.7  &46.2  &45.4 &32.3  &41.8  &52.7  \\
    \midrule
    
    LM-Critic~ &32.3 &41.3 &33.8 &38.9 &32.3 &37.3 &32.2 &21.7 &29.4 &46.4 \\
    \textbf{$y^{'} \mapsto y$} (2 Cycles)&42.0 &57.6 &44.4 &48.7 &45.8 &\bf48.0 &45.4 &38.2 &43.8 &\bf54.5 \\
    \textbf{$x_{h} \mapsto y$} (1 Cycle)&36.5  &46.3  &38.1 &39.5  &33.1  &38.0  &41.6  &28.9  &38.2  &50.5 \\
    \textbf {$\nabla_{ATK}$} (1 Cycles) &37.6 &48.7 &39.3 &42.5 &37.9 &41.6 &42.1 &33.8 &40.1 &51.8  \\
    \midrule
    
    BERT~ &38.8 &54.5 &41.1 &45.9 &42.3 &45.1 &46.6 &41.9 &45.6 &53.3 \\
    \textbf{$y^{'} \mapsto y$} (1 Cycle)&38.8 &55.9 &41.3 &46.4 &44.2 &45.9 &46.8 &42.7 &45.9 &\bf53.5 \\
    \textbf{$x_{h} \mapsto y$} (1 Cycle)&37.2  &51.9  &39.5 &44.5  &39.7  &43.5  &46.4  &40.0  &45.0  &52.9 \\
    \textbf {$\nabla_{ATK}$} (1 Cycles) &40.1 &54.1 &42.2 &46.1 &42.1 &45.2 &40.3 &38.7 &40.1 &53.5  \\
    \midrule
    
    RoBERTa~ &42.1 &57.0 &44.4 &48.1 &43.8 &47.2 &47.8 &42.8 &46.7 &53.9 \\
    \textbf{$y^{'} \mapsto y$} (1 Cycle)&42.2 &57.4 &44.6 &48.3 &44.3 &47.5 &48.3 &43.4 &47.2 &\bf54.2 \\
    \textbf{$x_{h} \mapsto y$} (1 Cycle)&41.5  &57.5  &43.9 &48.0  &45.1  &47.4  &47.5  &43.3  &46.6  &54.5 \\
    \textbf {$\nabla_{ATK}$} (3 Cycles) &42.8 &57.3 &45.1 &48.2 &46.1 &48.1 &45.1 &42.2 &44.4 &54.4  \\
    \midrule

    XLNet~ &43.0 &60.3 &45.6 &48.3 &47.1 &48.1 &50.6 &49.5 &50.4 &55.4 \\
    \textbf{$y^{'} \mapsto y$} (2 Cycles)&43.3 &60.0 &45.8 &48.4 &46.8 &48.1 &50.8 &49.5 &\bf50.5 &\bf55.5 \\
    \textbf{$x_{h} \mapsto y$} (1 Cycle)&42.6  &58.9  &45.1 &47.5  &44.7  &46.9  &50.6  &48.3  &50.1  &55.4 \\
    \textbf {$\nabla_{ATK}$} (1 Cycles) &43.3 &60.1 &45.9 &48.4 &46.9 &48.1 &43.5 &46.2 &44.1 &54.6  \\
    
    \bottomrule

    \end{tabular}
}
\caption{Evaluation results on \textit{Antonym Substitutions} attack sets. 
We generate the attack sets by using \textit{Antonym Substitutions} method as as mentioned in Section \ref{sec:sub_op}.
We report the performance of two variants of regularization data for each model and the corresponding best cycle times, where $\mapsto$ represents the direction of data flow.
The bold fonts indicate the optimal performance of each comparison.}
\label{tab:attack_ant} 
\end{table}

\clearpage
\section{Template for attack samples}
In this section, we provide some examples of the attack samples.
\begin{table}[t]
\centering
\small
\begin{tabular}{l l}
    \toprule
    \multicolumn{2}{l}{\textbf{Examples of Attacked Data}} \\
    \midrule

   
    
    
    \textit{source}:   &  Genetic risk  \textcolor{orange}{refers more}  to your chance of inheriting a disorder or disease .\\
    \textit{backtrans}:  &  Genetic risk \textcolor{teal}{means more}  to your chance of inheriting a disorder or disease .\\
    \textit{golden}:  & Genetic risk \textcolor{red}{refers} to your chance of inheriting a disorder or disease . \\
    \midrule
    
    \textit{source}:   &  Genetic \textcolor{orange}{risk does carry its} consequences.\\
    \textit{backtrans}:  &  Genetic \textcolor{teal}{risks do carry} consequences.\\
    \textit{golden}:  & Genetic \textcolor{red}{ does carry} consequences. \\
    \midrule
    
    \textit{source}:   &  For example , \textcolor{orange}{within a} family , ...\\
\bottomrule
\end{tabular}
\caption{Examples of mapping \& rules and back-translation attack samples.}
\label{tab:case-study} 
\end{table}

\begin{table}[hbt!]
\centering
\small
\begin{tabular}{l l}
    \toprule
    \multicolumn{2}{l}{\textbf{Examples of Attacked Data}} \\
    \midrule

    \textit{source}: &A \textcolor{orange}{good} law should be clear and easy to \textcolor{orange}{enforce} . \\  
    \textit{antonym}:  & A \textcolor{teal}{evil} law should be ill defined and easy to \textcolor{teal}{execute} . \\
    \textit{golden}:  & A \textcolor{red}{good} law should be clear and easy to \textcolor{red}{enforce} . \\
   
    \midrule
    \textit{source}:   & They can \textcolor{orange}{build} a daily habit \textcolor{orange}{to eat} less salt. \\
    \textit{antonym}:  & They can \textcolor{teal}{adopt} a daily habit \textcolor{teal}{of eating} less salt. \\
    \textit{golden}:  &  they can \textcolor{red}{adopt} a daily habit \textcolor{red}{of eating} less salt.\\
    \midrule
    
    \textit{source}: & For every \textcolor{orange}{action} there \textcolor{orange}{is} always pros and cons . \\
\bottomrule
\end{tabular}
\caption{Examples of antonym and synonym substitution attack samples.}
\label{tab:case-study} 
\end{table}

\clearpage
\section{Results with Different Seeds}
we report the results of $F_{0.5}$ score and GLEU score with five different seeds as well as provide the average and standard deviation of model performance.

\begin{table}[hbt!]
\centering
\small
\resizebox{\textwidth}{!}{
    \begin{tabular}{l |c c c c c| c c c c c| c c c c c| c c c c c }
    \toprule[1pt]
    \multirow{2}{*}{\textbf{Model}} & \multicolumn{5}{c|}{\textbf{BEA~} (ERRANT)} & \multicolumn{5}{c|}{\textbf{CoNLL~} ($M^{2}$)} & \multicolumn{5}{c|}{\textbf{FCE~} ($M^{2}$)} & \multicolumn{5}{c}{\textbf{JFLEG~~}}  \\
    \cline{2-6} \cline{7-11} \cline{12-16} \cline{17-21}
    & \textbf{\#1} & \textbf{\#2} & \textbf{\#3} & \textbf{\#4} & \textbf{\#5} & \textbf{\#1} & \textbf{\#2} & \textbf{\#3} & \textbf{\#4} & \textbf{\#5} & \textbf{\#1} & \textbf{\#2} & \textbf{\#3} & \textbf{\#4} & \textbf{\#5} & \textbf{\#1} & \textbf{\#2} & \textbf{\#3} & \textbf{\#4} & \textbf{\#5} \\
    \toprule
    \multicolumn{21}{c}{\bf Transformer} \\
    \hline
    \multirow{3}{*}{\bf $y^{'} \mapsto y$}  &13.18 &13.28 &13.05 &13.16 &13.33 &42.10 &41.92 &41.89 &41.98 &42.11 &24.33 &24.41 &24.26 &24.21 &24.29 &38.45 &38.40 &38.65 &38.44 &38.56   \\
    \cline{2-21}
     & \multicolumn{2}{c}{\bf Avg.} & \multicolumn{2}{c}{\bf Std.} & \bf F & \multicolumn{2}{c}{\bf Avg.} & \multicolumn{2}{c}{\bf Std.} & \bf F & \multicolumn{2}{c}{\bf Avg.} & \multicolumn{2}{c}{\bf Std.} & \bf F & \multicolumn{2}{c}{\bf Avg.} & \multicolumn{2}{c}{\bf Std.} & \bf F \\
    \cline{2-21}
    & \multicolumn{2}{c}{13.20} & \multicolumn{2}{c}{0.10} & - & \multicolumn{2}{c}{42.00} & \multicolumn{2}{c}{0.09} & 0.03 & \multicolumn{2}{c}{24.30} & \multicolumn{2}{c}{0.07} & 0.01 & \multicolumn{2}{c}{38.50} & \multicolumn{2}{c}{0.09} & 0.04 \\
    \hline
    \multirow{3}{*}{\bf $\nabla_{ATK}$}  &11.82 &12.03 &11.64 &11.85 &11.92 &37.67 &37.85 &37.82 &37.65 &37.80 &21.17 &21.18 &21.33 &21.10 &20.97 &37.56 &37.77 &37.65 &37.61 &37.47    \\
    \cline{2-21}
     & \multicolumn{2}{c}{\bf Avg.} & \multicolumn{2}{c}{\bf Std.} & \bf F & \multicolumn{2}{c}{\bf Avg.} & \multicolumn{2}{c}{\bf Std.} & \bf F & \multicolumn{2}{c}{\bf Avg.} & \multicolumn{2}{c}{\bf Std.} & \bf F & \multicolumn{2}{c}{\bf Avg.} & \multicolumn{2}{c}{\bf Std.} & \bf F \\
\bottomrule
    \end{tabular}
}
\caption{Evaluation results on \textit{Mapping $\&$ Rules} attack sets with five different seeds.
We report the performance of two variants of regularization data and one gradient-based adversarial attack for each model, where $\mapsto$ represents the direction of data flow.
In the end of each group, we also report the average and standard deviation as well as provide the P-Score~(\textbf{F}) of each model.}
\label{tab:attack_mapping_seed} 
\end{table}
\begin{table}[hbt!]
\centering
\small
\resizebox{\textwidth}{!}{
    \begin{tabular}{l |c c c c c| c c c c c| c c c c c| c c c c c }
    \toprule[1pt]
    \multirow{2}{*}{\textbf{Model}} & \multicolumn{5}{c|}{\textbf{BEA~} (ERRANT)} & \multicolumn{5}{c|}{\textbf{CoNLL~} ($M^{2}$)} & \multicolumn{5}{c|}{\textbf{FCE~} ($M^{2}$)} & \multicolumn{5}{c}{\textbf{JFLEG~~}}  \\
    \cline{2-6} \cline{7-11} \cline{12-16} \cline{17-21}
    & \textbf{\#1} & \textbf{\#2} & \textbf{\#3} & \textbf{\#4} & \textbf{\#5} & \textbf{\#1} & \textbf{\#2} & \textbf{\#3} & \textbf{\#4} & \textbf{\#5} & \textbf{\#1} & \textbf{\#2} & \textbf{\#3} & \textbf{\#4} & \textbf{\#5} & \textbf{\#1} & \textbf{\#2} & \textbf{\#3} & \textbf{\#4} & \textbf{\#5} \\
      \toprule
    \multicolumn{21}{c}{\bf Transformer} \\
    \hline
    \multirow{3}{*}{\bf $y^{'} \mapsto y$}  &22.11 &22.28 &22.04 &22.29 &21.78 &31.28 &31.12 &31.23 &31.24 &31.64 &29.85 &29.49 &29.80 &29.80 &30.07 &45.20 &45.03 &44.80 &45.12 &44.84   \\
    \cline{2-21}
     & \multicolumn{2}{c}{\bf Avg.} & \multicolumn{2}{c}{\bf Std.} & \bf F & \multicolumn{2}{c}{\bf Avg.} & \multicolumn{2}{c}{\bf Std.} & \bf F & \multicolumn{2}{c}{\bf Avg.} & \multicolumn{2}{c}{\bf Std.} & \bf F & \multicolumn{2}{c}{\bf Avg.} & \multicolumn{2}{c}{\bf Std.} & \bf F \\
    \cline{2-21}
    & \multicolumn{2}{c}{22.10} & \multicolumn{2}{c}{0.19} & - & \multicolumn{2}{c}{31.30} & \multicolumn{2}{c}{0.18} & 0.04 & \multicolumn{2}{c}{29.80} & \multicolumn{2}{c}{0.18} & 0.05 & \multicolumn{2}{c}{45.00} & \multicolumn{2}{c}{0.18} & 0.04 \\
    \hline
    \multirow{3}{*}{\bf $\nabla_{ATK}$}  &20.05 &20.08 &20.21 &20.05 &20.16  &29.73 &29.88 &29.67 &29.84 &29.79  &26.17 &26.28 &26.09 &26.02 &26.39 &43.25 &43.15 &43.57 &43.31 &43.42   \\
    \cline{2-21}
     & \multicolumn{2}{c}{\bf Avg.} & \multicolumn{2}{c}{\bf Std.} & \bf F & \multicolumn{2}{c}{\bf Avg.} & \multicolumn{2}{c}{\bf Std.} & \bf F & \multicolumn{2}{c}{\bf Avg.} & \multicolumn{2}{c}{\bf Std.} & \bf F & \multicolumn{2}{c}{\bf Avg.} & \multicolumn{2}{c}{\bf Std.} & \bf F \\
\bottomrule
    \end{tabular}
}
\caption{Evaluation results on \textit{Synonym Substitutions} attack sets with five different seeds.
We report the performance of two variants of regularization data and one gradient-based adversarial attack for each model, where $\mapsto$ represents the direction of data flow.
In the end of each group, we also report the average and standard deviation as well as provide the P-Score~(\textbf{F}) of each model.}
\label{tab:attack_ant_seed} 
\end{table}
\begin{table}[hbt!]
\centering
\small
\resizebox{\textwidth}{!}{
    \begin{tabular}{l |c c c c c| c c c c c| c c c c c| c c c c c }
    \toprule[1pt]
    \multirow{2}{*}{\textbf{Model}} & \multicolumn{5}{c|}{\textbf{BEA~} (ERRANT)} & \multicolumn{5}{c|}{\textbf{CoNLL~} ($M^{2}$)} & \multicolumn{5}{c|}{\textbf{FCE~} ($M^{2}$)} & \multicolumn{5}{c}{\textbf{JFLEG~~}}  \\
    \cline{2-6} \cline{7-11} \cline{12-16} \cline{17-21}
    & \textbf{\#1} & \textbf{\#2} & \textbf{\#3} & \textbf{\#4} & \textbf{\#5} & \textbf{\#1} & \textbf{\#2} & \textbf{\#3} & \textbf{\#4} & \textbf{\#5} & \textbf{\#1} & \textbf{\#2} & \textbf{\#3} & \textbf{\#4} & \textbf{\#5} & \textbf{\#1} & \textbf{\#2} & \textbf{\#3} & \textbf{\#4} & \textbf{\#5} \\
        \toprule
    \multicolumn{21}{c}{\bf Transformer} \\
    \hline
    \multirow{3}{*}{\bf $y^{'} \mapsto y$}  &28.43 &28.27 &28.22 &28.61 &28.47  &35.81 &35.99 &35.75 &35.65 &35.81 &33.57 &33.37 &33.65 &33.74 &33.66 &46.71 &46.80 &46.62 &46.70 &46.67    \\
    \cline{2-21}
     & \multicolumn{2}{c}{\bf Avg.} & \multicolumn{2}{c}{\bf Std.} & \bf F & \multicolumn{2}{c}{\bf Avg.} & \multicolumn{2}{c}{\bf Std.} & \bf F & \multicolumn{2}{c}{\bf Avg.} & \multicolumn{2}{c}{\bf Std.} & \bf F & \multicolumn{2}{c}{\bf Avg.} & \multicolumn{2}{c}{\bf Std.} & \bf F \\
    \cline{2-21}
    & \multicolumn{2}{c}{28.40} & \multicolumn{2}{c}{0.14} & - & \multicolumn{2}{c}{35.80} & \multicolumn{2}{c}{0.11} & 0.04 & \multicolumn{2}{c}{33.60} & \multicolumn{2}{c}{0.13} & 0.01 & \multicolumn{2}{c}{46.70} & \multicolumn{2}{c}{0.06} & 0.02 \\
    \hline
    \multirow{3}{*}{\bf $\nabla_{ATK}$}  &27.78 &27.86 &27.59 &27.69 &27.92 &35.11 &35.29 &34.99 &35.17 &35.19 &32.47 &32.32 &32.70 &32.37 &32.39  &46.12 &46.34 &46.15 &46.08 &45.97   \\
    \cline{2-21}
     & \multicolumn{2}{c}{\bf Avg.} & \multicolumn{2}{c}{\bf Std.} & \bf F & \multicolumn{2}{c}{\bf Avg.} & \multicolumn{2}{c}{\bf Std.} & \bf F & \multicolumn{2}{c}{\bf Avg.} & \multicolumn{2}{c}{\bf Std.} & \bf F & \multicolumn{2}{c}{\bf Avg.} & \multicolumn{2}{c}{\bf Std.} & \bf F \\
\bottomrule
    \end{tabular}
}
\caption{Evaluation results on \textit{Back-Translation} attack sets with five different seeds.
We report the performance of two variants of regularization data and one gradient-based adversarial attack for each model, where $\mapsto$ represents the direction of data flow.
In the end of each group, we also report the average and standard deviation as well as provide the P-Score~(\textbf{F}) of each model.}
\label{tab:attack_ant_seed} 
\end{table}
\begin{table}[t]
\small
\centering
\resizebox{\textwidth}{!}{
    \begin{tabular}{l |c c c c c| c c c c c| c c c c c| c c c c c }
    \toprule[1pt]
    \multirow{2}{*}{\textbf{Model}} & \multicolumn{5}{c|}{\textbf{BEA~} (ERRANT)} & \multicolumn{5}{c|}{\textbf{CoNLL~} ($M^{2}$)} & \multicolumn{5}{c|}{\textbf{FCE~} ($M^{2}$)} & \multicolumn{5}{c}{\textbf{JFLEG~~}}  \\
    \cline{2-6} \cline{7-11} \cline{12-16} \cline{17-21}
    & \textbf{\#1} & \textbf{\#2} & \textbf{\#3} & \textbf{\#4} & \textbf{\#5} & \textbf{\#1} & \textbf{\#2} & \textbf{\#3} & \textbf{\#4} & \textbf{\#5} & \textbf{\#1} & \textbf{\#2} & \textbf{\#3} & \textbf{\#4} & \textbf{\#5} & \textbf{\#1} & \textbf{\#2} & \textbf{\#3} & \textbf{\#4} & \textbf{\#5} \\
      \toprule
    \multicolumn{21}{c}{\bf Transformer} \\
    \hline
    \multirow{3}{*}{\bf $y^{'} \mapsto y$}  &42.74 &42.73 &42.53 &42.94 &42.57 &45.97 &46.03 &45.82 &45.98 &46.20 &45.88 &45.77 &45.74 &45.97 &46.14  &55.52 &55.54 &55.35 &55.43 &55.66    \\
    \cline{2-21}
     & \multicolumn{2}{c}{\bf Avg.} & \multicolumn{2}{c}{\bf Std.} & \bf F & \multicolumn{2}{c}{\bf Avg.} & \multicolumn{2}{c}{\bf Std.} & \bf F & \multicolumn{2}{c}{\bf Avg.} & \multicolumn{2}{c}{\bf Std.} & \bf F & \multicolumn{2}{c}{\bf Avg.} & \multicolumn{2}{c}{\bf Std.} & \bf F \\
    \cline{2-21}
    & \multicolumn{2}{c}{42.70} & \multicolumn{2}{c}{0.15} & - & \multicolumn{2}{c}{46.00} & \multicolumn{2}{c}{0.12} & 0.03 & \multicolumn{2}{c}{45.90} & \multicolumn{2}{c}{0.15} & 0.05 & \multicolumn{2}{c}{55.50} & \multicolumn{2}{c}{0.11} & 0.05 \\
    \hline
    \multirow{3}{*}{\bf $\nabla_{ATK}$}  &42.53 &42.47 &42.65 &42.55 &42.31 &39.82 &39.89 &39.93 &39.79 &39.82 &42.69 &42.48 &42.41 &42.63 &42.78 &52.34 &52.53 &52.51 &52.42 &52.20   \\
    \cline{2-21}
     & \multicolumn{2}{c}{\bf Avg.} & \multicolumn{2}{c}{\bf Std.} & \bf F & \multicolumn{2}{c}{\bf Avg.} & \multicolumn{2}{c}{\bf Std.} & \bf F & \multicolumn{2}{c}{\bf Avg.} & \multicolumn{2}{c}{\bf Std.} & \bf F & \multicolumn{2}{c}{\bf Avg.} & \multicolumn{2}{c}{\bf Std.} & \bf F \\
\bottomrule
    \end{tabular}
}
\caption{Evaluation results on \textit{Antonym Substitutions} attack sets with five different seeds.
We report the performance of two variants of regularization data and one gradient-based adversarial attack for each model, where $\mapsto$ represents the direction of data flow.
In the end of each group, we also report the average and standard deviation as well as provide the P-Score~(\textbf{F}) of each model.}
\label{tab:attack_ant_seed} 
\end{table}

\clearpage
\section{Gradient-Based Defence Method}
\label{app:gdm}
Gradient-based adversarial attack~\cite{szegedy2013intriguing,yasunaga2017robust,wallace2019universal} is another popular continuous attack method, which considers continuous perturbations to inputs. 
Followed by the previous work~\cite{yasunaga2017robust}, we inject the noise into the embedding $E$ of each model.
Specifically, given the training pairs $(x, y)$, we can obtain an approximate worst-case perturbation $\eta$ of norm $\tau$:
\begin{equation}
    \left\{
    \begin{aligned}
        \eta &= \tau g/||g||_{2}, \\
        g &= \nabla_{E}\mathcal{L}(\theta;E, y),
    \end{aligned}
    \right.
\end{equation}
where $\tau$ is a hyper-parameter and $\theta$ represents for the parameters in the model. 
Thus, we can construct the adversarial examples by
\begin{equation}
    E_{adv} = E + \eta
\end{equation}
Then the total loss function can be written as:
\begin{equation}
\label{equ:gradient_loss}
    \mathcal{L} = (1 - \gamma)\mathcal{L}(\theta; E, y) + \gamma\mathcal{L}(\theta; E_{adv}, y)
\end{equation}
Specifically, we set $\gamma$ as 0.5 for all the baselines. 
Besides, we find that the settings of $\tau$ have a great impact on the final results.
Thus, we search for the optimal value of $\tau \in \{0.05, 0.1, 0.15, 0.2\}$ by conducting numerous experiments, and report the best settings of $\tau$ for each GEC models in Table~\ref{tab:tau_value}.

\begin{table}[h]
\centering
\small
\begin{tabular}{l|l}
    \toprule
     \bf Model & \bf Value  \\
     \midrule
     Transformer &  0.1 \\
     BERT-fuse & 0.01 \\
     BART & 0.1 \\
     LM-Critic & 0.01\\
     BERT & 0.5 \\
     RoBERTa & 0.5 \\
     XLNet & 0.5 \\
     \bottomrule
\end{tabular}
\caption{Settings of $\tau$ for different GEC models.}
\label{tab:tau_value}
\end{table}

\clearpage
\section{Chinese Grammatical Error Correction}
To explore whether the proposed CSA method can contribute to non-English GEC research, we conduct experiments on the Chinese GEC dataset.
Following the previous work~\cite{tang2021chinese}, we utilize the Chinese Gigaword dataset\footnote{\url{https://catalog.ldc.upenn.edu/LDC2009T27}} to pre-train the Transformer~\cite{vaswani2017attention} model and fine-tune the model with Lang-8 dataset\footnote{\url{http://tcci.ccf.org.cn/conference/2018/dldoc/trainingdata02.tar.gz}}.
As for constructing the Chinese attack set, we set the perturbation rate as 30\% with a random sampling strategy and utilize the Chinese synonyms tools\footnote{\url{https://github.com/425776024/nlpcda}} for replacing the words in the selecting positions.  
If there is no candidate, we delete the words directly. 
We follow the process of the CSA method to enhance the robustness of the GEC model and compare the results on the original NLPCC2018 testing set and the corresponding attack set.
The results are shown in Table~\ref{tab:cgec}. 
The results confirm the effectiveness and generalization capability of our proposed method. 

\begin{table}[htbp]
\centering
\small
\begin{tabular}{l | l | l l l}
\toprule
\bf Model &\bf Data & \bf $P$ & \bf $R$ & \bf $F_{0.5}$ \\
\midrule
Transformer-base & ori test & 41.47 & 22.62 & 35.55 \\ 
Transformer-CSA & ori test & \bf 43.26 & \bf 23.74 & \bf 37.15 \\                  
\midrule
Transformer-base & attack & 30.28 & 34.29 & 31.01 \\
Transformer-CSA & attack & \bf 34.19 &\bf  31.29 & \bf 33.57 \\
\bottomrule
\end{tabular}
\caption{Results of the Chinese GEC models. The bold fonts are the best performance of each comparison.}
\label{tab:cgec}
\end{table}

\end{document}